\documentclass[reqno,11pt]{amsart}
\usepackage{amsmath,amsthm,amsfonts,color,graphicx,bbm}
\graphicspath{ {figs/} }
\usepackage[latin1]{inputenc}
\usepackage[makeroom]{cancel}
\usepackage{pdfsync,subfigure,multirow,float}
\usepackage{filecontents,diagbox,bbold}
\definecolor{lightGray}{RGB}{235,235,235}
\definecolor{orange}{RGB}{255,128,0}
\definecolor{ucib}{RGB}{0,36,105}
\definecolor{mygreen}{RGB}{0,128,0}
\definecolor{lightBlue}{RGB}{102,153,204}

\newtheorem{thm}{Theorem}[section]
\newtheorem{lem}[thm]{Lemma}

\newtheorem{prop}[thm]{Proposition}

\newtheorem{rem}[thm]{Remark}

\newtheorem{deff}[thm]{Definition}

\DeclareMathAlphabet{\mathpzc}{OT1}{pzc}{m}{it}

\numberwithin{equation}{section}

\begin{document}
\bibliographystyle{plain}

\title[Data Sets Approximation and Classification]{Dimension
  Independent Data Sets Approximation and Applications to Classification}

\author{Patrick Guidotti}
\address{University of California, Irvine\\
Department of Mathematics\\
340 Rowland Hall\\
Irvine, CA 92697-3875\\ USA }
\email{gpatrick@math.uci.edu}

\begin{abstract}
We revisit the classical kernel method of approximation/interpolation
theory in a very specific context motivated by the desire to obtain a
robust procedure to approximate discrete data sets by (super)level
sets of functions that are merely continuous at the data set arguments
but are otherwise smooth. Special functions, called data signals, are
defined for any given data set and are used to succesfully solve
supervised classification problems in a robust way that depends
continuously on the data set.  The efficacy of the method is
illustrated with a series of low dimensional examples and by its
application to the standard benchmark high dimensional problem of
MNIST digit classification.
\end{abstract}

\subjclass[1991]{}

\maketitle

\section{Introduction}
The problem of finding a function that explains a given set of data is
a fundamental problem in mathematics and statistics. If the data are
assumed to be the discrete manifestation of a function defined on a
continuous (as opposed to discrete) domain of definition, the problem
can be viewed as an approximation problem where the data can be
leveraged to help identify a sensible approximation to the
function. Often one resorts the prior 
knowledge about the target function to reduce the set of candidates
from which to choose a good approximation, if not the best
approximation. Within this framework, an extensive mathematical
knowledge has been obtained over the past several decades along with a
variety of powerful tools (see \cite{W04}, in particular, for the
philosophical approach taken here). In the current world, where data
about almost anything you can imagine or wish for is available, one of
the most interesting and often challening problems consists in
extracting information, knowledge, and structure from high dimensional data. A
variety of commonly used approaches belong to a category referred to
as Machine Learning. Neural networks in all forms and shapes are
particularly widespread due to their success in 
dealing with a series of challenging problems. Another class is that
of Support-Vector Machines (SVM) \cite{VC64}, which were very popular
before being somewhat superseded by improved neural networks . While
they often use linear classification via hyperlanes, the so-called
kernel trick \cite{ABR64,BGV92} makes it possible for them
to capture nonlinear decision boundaries albeit at the cost of working
in a higher dimensional space to which the data is mapped but that
admits an efficient computation of scalar products (via a suitable
kernel). A connection between SVMs and neural networks was discovered
in \cite{PG90} and has since been investigated by many more authors. 
The method proposed here is philosophically in the category of SVMs but
distinguishes itself by directly working with the data at hand without
embedding into a higher dimensional feature space. We first review
a classical method of inexact interpolation that yields a continuous
function approximating a given data set. We do so by taking a PDE
perspective that reveals important features that are exploited in
order to obtain the announced stable method of classification even
when the distribution of available data is not uniform in space and,
possibly, noisy. In high dimension, even large data sets are often
sparse due to the so-called curse of dimensionality. Moreover, the
data is often supported on lower dimensional manifolds, where even
dense data make up a thin slice of the ambient space. In the latter
case, it is demonstrated in this paper that, while the data may not be
sufficient for the reliable identification of a well-defined global
approximant, it can still be used fruitfully (in a global or local
fashion) for data analysis purposes. This is mainly due to the fact
that the proposed method is capable of connecting the dots (data
points) into a manifold by capturing geometric features of the data. 

It is widely understood and accepted that a function interpolating
discrete data should be at least continuous so as to provide a certain
stability of prediction and resilience in the face of noise. In
learning problems this is sometimes expressed 
as local constancy, even if that does not require continuity. Unless
the data is known to stem from a very smooth underlying function, but,
typically, even in that case, it should also not be exceedingly smooth,
as this would lead to some blurring and reduce its ability to capture
sharp transitions. It would, moreover, be advantageous for the
interpolating function to depend at least continuously on the data set
it is constructed from. This would make it robust with respect to data
uncertainty and thus defensible, for example, against what in the
learning community would be referred 
to as adversarial attacks. These in fact use the very sensititve
dependence of the model parameters on the training set.

The starting point is a set of data consisting of point/value pairs
$$
 \mathbb{D}=\big\{ (x^i,y^i)\,\big |\, i=1,\dots, m\big\} 
 $$
 where $m\in \mathbb{N}$, $x^i\in \mathbb{R}^d$ and $y^i\in
 \mathbb{R}^n$, $i=1,\dots,m$, for $d,n\in \mathbb{N}$. In order to
 enforce minimal regularity on the interpolant function
$$
u:\mathbb{R}^d\to \mathbb{R}^n
$$
we take it from the space
$$
\operatorname{H}^{\frac{d+1}{2}}(\mathbb{R}^d,\mathbb{R}^n)=\big\{ u\in \mathcal{S}
'\,\big |\, [\xi\mapsto (1+|\xi|^2)^{\frac{d+1}{2}}\hat u(\xi)]\in
\operatorname{L}^2(\mathbb{R}^d,\mathbb{R}^n)\big\}.
$$
Thanks to the general Sobolev inequality it holds that
\begin{equation}\label{sob}
\operatorname{H}^{\frac{d+1}{2}}(\mathbb{R}^d,\mathbb{R}^n)\hookrightarrow
\operatorname{BUC}^{\frac{1}{2}}(\mathbb{R}^d,\mathbb{R}^n), 
\end{equation}
where the containing space is that of bounded and uniformly H\"older
continuous functions of exponent $\frac{1}{2}$. In order to obtain
stability we may sacrifice some interpolation accuracy by not
necessarily requiring the exact validity of
\begin{equation}\label{idata}
 u(x^i)=y^i\text{ for }i=1,\dots, m.
\end{equation}
Then, for approximate interpolation, $u$ is determined by minimization
of the energy functional given by
\begin{align}
 E_\alpha (u)&=\frac{\alpha}{2c_d}\int _{\mathbb{R}^d} \big |
 (1-\Delta)^{\frac{d+1}{4}}u(x)\big |^2\, dx+ \frac{1}{2}\sum_{i=1}^m
 \big | u(x^i)-y^i\big |^2\\
  &= \frac{\alpha}{2c_d}\| u\|_{\operatorname{H}^{\frac{d+1}{2}}}^2+ \frac{1}{2}\sum_{i=1}^m
 \big | u(x^i)-y^i\big |^2, \: u\in
         \operatorname{H}^{\frac{d+1}{2}}(\mathbb{R}^d,\mathbb{R}^n)
\end{align}
for $\alpha>0$ and where the normalizing constant $c_d$ will be explicitly given in the
next section. For exact interpolation $u$ is determined by 
minimization of $E_0=\| \cdot\|_{\operatorname{H}^{\frac{d+1}{2}}}^2$ with
constraints \eqref{idata} that shall be summarized as
$u(\mathbb{X})=\mathbb{Y}$. Formally, it is set
\begin{equation}\label{minpb}
  u_{\mathbb{D},\alpha }=\begin{cases}
    \operatorname{argmin}_{u\in\operatorname{H}^{\frac{d+1}{2}}(\mathbb{R}^d,\mathbb{R}^n)}
    E_\alpha(u),&\alpha>0,\\
    \operatorname{argmin}_{u\in\operatorname{H}^{\frac{d+1}{2}}(\mathbb{R}^d,\mathbb{R}^n),\:
    u(\mathbb{X})=\mathbb{Y}}E_0(u),&\alpha=0.\end{cases}
\end{equation}
While in many practical problems the dimension $d$ can
be very large and the constant $c_d$ astronomical, this approach
remains viable since the minimizers can be identified by
solving a well-posed $m\times m$ linear system of equations for any
$\alpha\geq 0$.
The rest of the paper is organized as follows. In the next section we
provide a detailed description of the method and obtain some of its basic
mathematical properties. In the following section, we discuss a
variety of numerical experiments that showcase the viability and
efficacy of the method.
\section{The Method}
In order to derive a concrete method it needs to be shown that
minimizers of $E_d$ can be computed efficiently. We first observe that
the functional has a unique minimizer no matter what the given data
set is.
\begin{thm}
  The functional $E_\alpha$ has a unique minimizer
  $u_{\mathbb{D},\alpha}$  for any given data set $\mathbb{D}$. 
\end{thm}
\begin{proof}
Take $\alpha>0$ first. Thanks to the embedding \eqref{sob} the
functional $E_\alpha :
\operatorname{H}^{\frac{d+1}{2}}(\mathbb{R}^d,\mathbb{R}^n)\to [0,\infty)$ is
continuous. It is clearly also strictly convex as a quadratic
functional since the first term is the square of a norm. Strongly lower
semi-continuous convex functionals on a Hilbert space are known to
be weakly lower-semicontinuous and so is therefore $E_\alpha$. It is also
coercive since closed bounded sets in
$\operatorname{H}^{\frac{d+1}{2}}(\mathbb{R}^d,\mathbb{R}^n)$ are relatively weakly
compact by the Banach-Alaoglu Theorem. A weakly lower-semicontinuous
and coercive functional possesses a minimum, which, by strict
convexity, is necessarily unique. The case $\alpha=0$ uses essentially
the same argument where the full function space is replaced by the
convex closed subset of functions satisfying $u(\mathbb{X})=\mathbb{Y}$.
\end{proof}
Now that a unique continuous interpolant $u_{\mathbb{D}}$ 
has been obtained for each given data set $\mathbb{D}$, it needs to be
determined in a usable form. The next step consists in deriving the
Euler-Lagrange equation for $u_{\mathbb{D}}$.
\begin{thm}
  The minimizer $u_{\mathbb{D},\alpha}$ for $\alpha>0$satisfies the
  equation (system)
  \begin{equation}\label{ela}
   \frac{\alpha}{c_d} (1-\Delta)^{\frac{d+1}{2}}u=\sum_{i=1}^m \bigl[
   y^i-u(x^i)\bigr] \delta_{x^i}
  \end{equation}
  in the weak sense, i.e. the equation holds in the space
  $\operatorname{H}^{-\frac{d+1}{2}}(\mathbb{R}^d,\mathbb{R}^n)$ or, equivalently
  it holds that
  \begin{equation}\label{elv}
  \frac{\alpha}{c_d} \int _{\mathbb{R}^d}\bigl[(1-\Delta)^{\frac{d+1}{4}}u\bigr](x)\cdot
  \bigl[(1-\Delta)^{\frac{d+1}{4}}v\bigr](x)\, dx=\sum_{i=1}^m \bigl[
   y^i-u(x^i)\bigr] \cdot v(x^i),
  \end{equation}
  for all
  $v\in\operatorname{H}^{\frac{d+1}{2}}(\mathbb{R}^d,\mathbb{R}^n)$. Here
  $\delta_x$ denotes the Dirac distribution supported at the point
  $x\in \mathbb{R}^d$. If $\alpha=0$ it holds that
  \begin{equation}\label{u0}
   (1-\Delta)^{\frac{d+1}{4}}u_{\mathbb{D},0}=\sum _{i=1}^m \lambda_i \delta _{x^i},
  \end{equation}
  in the weak sense for some $\lambda_i\in \mathbb{R}^n$ for $i=1,\dots,m$. 
\end{thm}
\begin{proof}
  Notice that, thanks to \eqref{sob}, it holds that $\delta _x\in
  \operatorname{H}^{-\frac{d+1}{2}}(\mathbb{R}^d,\mathbb{R})$ for each
  $x\in \mathbb{R}^d$. 
  Taking variations in direction of any function $ve_k$ for $k=1,\dots,n$, where $e_k$ is
  the $k$th basis element of $\mathbb{R}^n$, with arbitrary $v\in
  \operatorname{H}^{\frac{d+1}{2}}(\mathbb{R}^d)$ yields the equations
  \begin{multline*}
  0=\left.\frac{d}{ds}\right|_{s=0} E_\alpha (u_
  {\mathbb{D}}+sve_k)=\frac{\alpha}{c_d} \int
  _{\mathbb{R}^d}\bigl[(1-\Delta)^{\frac{d+1}{4}}u_{\mathbb{D},k}\bigr](x)
  \bigl[(1-\Delta)^{\frac{d+1}{4}}v\bigr](x)\, dx\\-\sum_{i=1}^m \bigl[
   y^i_k-u_{\mathbb{D},k} (x^i)\bigr] \cdot v(x^i)\text{ for } k=1,\dots, n,
 \end{multline*}
for $v\in \operatorname{H}^{\frac{d+1}{2}}(\mathbb{R}^d)$, and where
$u_ {\mathbb{D},k} =(u_ {\mathbb{D}})_k$ . The
identities amount to the validity of the system \eqref{elv}. The
equivalence of the latter with \eqref{ela} follows from
$$
 \sum_{i=1}^m \bigl[ y^i-u_ {\mathbb{D}}(x^i)\bigr] \cdot v(x^i)=\big \langle 
 \sum_{i=1}^m \bigl[ y^i-u_ {\mathbb{D}}(x^i)\bigr] \delta_{x^i},v
 \big\rangle ,\:
 v\in\operatorname{H}^{\frac{d+1}{2}}(\mathbb{R}^d,\mathbb{R}^n),
$$
and the fact that the (pseudo)differential operator
$(1-\Delta)^{\frac{d+1}{4}}$ is self-adjoint along with the validity
of
\begin{multline*}
 \int _{\mathbb{R}^d}\bigl[(1-\Delta)^{\frac{d+1}{4}}u\bigr](x)\cdot
  \bigl[(1-\Delta)^{\frac{d+1}{4}}v\bigr](x)\, dx\\=\big \langle (1-\Delta)^{\frac{d+1}{4}}u
 ,(1-\Delta)^{\frac{d+1}{4}}v\big \rangle =\big \langle (1-\Delta)^{\frac{d+1}{2}}u
 ,v\big \rangle\text{ for }u,v\in\operatorname{H}^{\frac{d+1}{2}}(\mathbb{R}^d,\mathbb{R}^n),
\end{multline*}
where the latter duality pairing is between the space
$\operatorname{H}^{\frac{d+1}{2}}(\mathbb{R}^d,\mathbb{R}^n)$ and its
dual
$\operatorname{H}^{-\frac{d+1}{2}}(\mathbb{R}^d,\mathbb{R}^n)$. In the
case when $\alpha=0$, one takes variations in direction of test
$\operatorname{C}^\infty$ functions with
$\varphi(\mathbb{X})=\mathbb{0}$ to obtain that 
$$
 (1-\Delta)^{\frac{d+1}{2}}u=0\in \mathbb{R}^d\setminus
 \mathbb{X},
$$
in the sense of distributions. This means that
$$
 \operatorname{supp}\bigl( (1-\Delta)^{\frac{d+1}{2}}u\bigr) \subset
 \mathbb{X}.
$$
It is know that compactly supported distributions are of finite order
and thus it must hold that 
$(1-\Delta)^{\frac{d+1}{2}}u$ is a finite linear combinations of
$\delta_{x^i}$, $i=1,\dots,m$, and derivatives of them. Notice, however, that
$\partial_l\delta _x \notin
\operatorname{H}^{-\frac{d+1}{2}}(\mathbb{R}^d)$ for $l=1,\dots,d$ and
so no derivatives can be present in the linear combination. This
yields the claim.
\end{proof}
\begin{lem}\label{c-infty}
Let $u_{\mathbb{D},\alpha}$ be the minimizer of $E_\alpha$ for the data set
$\mathbb{D}$. Then $u_{\mathbb{D},\alpha}\in \operatorname{C}^\infty(\mathbb{R}^d
\setminus \mathbb{X},\mathbb{R}^n)$, regardless of the regularization
parameter $\alpha\geq 0$. 
\end{lem}
\begin{proof}
This regularity will readily follow from the reprentation of the
solution discussed next.
\end{proof}
The minimization of $E_0$ or a variety of similar functionals has long
been recognized to provide an answer to the so-called universal
approximation property in the context of learning. It indeed
can be shown that
$$
 u_{\mathbb{D},0}\to f\text{ as } |\mathbb{X}|\to\infty,
$$
if the values $\mathbb{Y}=\big\{ y^i\,\big |\, i=1,\dots,m\big\} $ are
those $\mathbb{Y}=f(\mathbb{X})$ of a function $f$
belonging to a variety of (suitably chosen) functions spaces, such as
$\operatorname{C}_c(\mathbb{R}^d,\mathbb{R}^n)$ (compactly supported
continuous functions) or
$\operatorname{L}^p(\mathbb{R}^d,\mathbb{R}^n)$ for $p\in[1,\infty)$. The convergence
takes place in the space's natural topology as $\mathbb{X}$ becomes a
finer and finer, not necessarily regular, discrete grid that fills the
whole domain. The curse of dimensionality, however, limits the
applicabilty of this approximation procedure as the size of any finite
``filling'' grid is exponential in the ambient dimension.

A reason the above approach or, more in general, kernel based
approximation or interpolation has found widespread use, is its
ability to bridge the gap between finite and infinite dimensional
spaces. This property amounts in the case at hand in the possibility
of computing the continuous variable solution $u_{\mathbb{D},\alpha}$ by
solving finite dimensional systems.
\begin{thm}
 The minimizer $u_ {\mathbb{D},\alpha}$, $\alpha>0$, is completely
 determined by its values $u_{\mathbb{X},\alpha}=u_{\mathbb{D},\alpha}\big |_\mathbb{X}
 =\bigl( u_ {\mathbb{D},\alpha }(x^i)\bigr)_{i=1,\dots,m}$, on
 the set of arguments $\mathbb{X}=\big\{ x^i\,\big |\,
 i=1,\dots,m\big\}$. The latter can be determined by solving the
 well-posed linear system
\begin{equation}\label{theSys}
  \bigl(\alpha +M_\mathbb{D}\bigr) u_{\mathbb{X},\alpha }= M_\mathbb{D}
  \mathbb{Y},
\end{equation}
where the matrix $M_\mathbb{D} \in \mathbb{R}^{m\times m}$ is given by
$$
 M_{ij}=\exp\bigl(-2\pi |x^i-x^j|\bigr),\: i,j=1\dots, m.
$$
 It then holds that
 \begin{equation}\label{theInterpolant}
 u_ {\mathbb{D},\alpha }(x)=\frac{1}{\alpha}\sum_{j=1}^m \bigl[ y^j-
 u_{\mathbb{X},\alpha }^j\bigr]\exp\bigl(-2\pi |x-x^j|\bigr),
 \end{equation}
 for any $x\in \mathbb{R}^d$. For $\alpha=0$, it holds that
 $$
 \Lambda =M_\mathbb{D}^{-1}\mathbb{Y},\text{ i.e. that }
 \lambda_i=\sum_{j=1}^m  \bigl[M_\mathbb{D}^{-1}\bigr]_{ij}y^j
 $$
\end{thm}
\begin{proof}
The Fourier transform of the Laplace kernel is known. Indeed
\begin{equation}\label{ft}
 \mathcal{F}_d \bigl[ \exp(-2\pi \varepsilon |\cdot|)\bigr](\xi)
 =\frac{c_d\varepsilon}{\bigl(\varepsilon^2+|\xi|^2\bigr)^{\frac{d+1}{2}}},
\end{equation}
where $c_d=\Gamma(d+1)/\pi^\frac{d+1}{2}$. For the purposes of this
paper the parameter $\varepsilon$ is set to be 1. The right-hand side
of the above identity is the symbol of the pseudo-differential
operator $c_d(1-\Delta)^{-\frac{d+1}{2}}$. Now, equation \eqref{ela}
is equivalent to
\begin{equation}
\alpha\, u=\sum_{j=1}^m \bigl[
y^j-u(x^j)\bigr] c_d(1-\Delta)^{-\frac{d+1}{2}}\delta_{x^j}\notag
=\sum_{j=1}^m \bigl[ y^j-u(x^j)\bigr] \exp\bigl(-2\pi |\cdot-x^j|\bigr),
\end{equation}
where the second equality sign follows from \eqref{ft} and well-known
properties of the Fourier transform. It only remains to evaluate this
identity on the arguments $\mathbb{X}$ to obtain the finite linear
system. When $\alpha=0$, representation \eqref{u0} entails that
$$
 u_{\mathbb{D},0}=\sum_{j=1}^m \lambda_j \exp\bigl(-2\pi
 |\cdot-x^j|\bigr),
$$
and therefore that
$$
 \mathbb{Y}=u_{\mathbb{D},0}(\mathbb{X})=M_\mathbb{D}\Lambda,
 $$
as claimed.
\end{proof}
\begin{prop}
The matrix $M$ is invertible and it holds that
$$
 u_{\mathbb{X},\alpha}\to \mathbb{Y} \text{ as }\alpha\to 0.
$$
\end{prop}
\begin{proof}
The fact that $M$ is invertible is a consequence of the fact that
$\exp(-2\pi|\cdot-\cdot|)$ is a positive kernel as follows from its Fourier
transform and the well-known characterization of
positivity. Continuity of the inversion function inv
$$
\operatorname{inv}: \operatorname{GL}_m\to \operatorname{GL}_m,
M\mapsto M^{-1},
$$
then shows that
$$
 u_{\mathbb{X},\alpha}=(\alpha+M_\mathbb{D} )^{-1}M_\mathbb{D} \to
 M_\mathbb{D}^{-1}M_\mathbb{D} \mathbb{Y}=\mathbb{Y} \text{ as }
 \alpha\to 0.
$$
\end{proof}

\begin{rem}
The proof of Lemma \ref{c-infty} is now obvious since the explicit
representation of $u_{\mathbb{D},\alpha}$ reveals that singularities
are only found on the set $\mathbb{X}$.
\end{rem}

\begin{rem}
The convergence $u_{\mathbb{X},\alpha}\to u_{\mathbb{X},0}$ as
$\alpha\to 0$ implies convergence $u_{\mathbb{D},\alpha}$ to
$u_{\mathbb{D},0}$ in the topology of
$\operatorname{H}^{\frac{d+1}{2}}(\mathbb{R}^d,\mathbb{R}^n)$ and
hence, uniform convergence as well, thanks the embedding \eqref{sob}.
\end{rem}

\begin{rem}
Since the continuous minimization problem \eqref{minpb} has a unique
solution, the linear system \eqref{theSys} is assured to be solvable
and, in fact, well conditioned. It also follows that its
solution $u_{\mathbb{X}}$, as well as its extension $u_\mathbb{D}$ to $\mathbb{R}^d$,
depend continuously on the data $\mathbb{D}$ since the forcing term in
\eqref{ela} depends continuously on $\mathbb{D}$. The latter follows
from the linear dependence on $\mathbb{Y}$ and the fact that Dirac
distributions depend continuously on the location of their support in
the topology of
$\operatorname{H}^{-\frac{d+1}{2}}(\mathbb{R}^d)$, again
a known consequence of \eqref{sob}.
\end{rem}
\begin{rem}
Depending on the data set $\mathbb{D}$, the values $u_\mathbb{X}$
will be close or not so close to the prescribed values $\big\{ y^i\,\big
|\, i=1,\dots,m\big\}$. Thus, if $u_\mathbb{D}$ is considered an
interpolant of the data, it will not be exact, but only approximately
capture the data. In many applications, some of which are considered
in next section, this is a small price to be paid for the gain in
robustness that the approach guarantees.
\end{rem}
\begin{rem}
Using directly that
$\frac{1}{c_d}(1-\Delta)^{-\frac{d+1}{2}}u_\mathbb{D}
=\sum_{i=1}^m \lambda _i \delta_{x^i}$ for some $\lambda\in
\mathbb{R}^m$, it is possible to derive a system of equations for
$\lambda$ using that $u_\mathbb{D}=\sum_{i=1}^m \lambda _i \exp \bigl(
-2\pi |x-x^i|\bigr) $. Indeed it must hold
$$
\alpha \sum_{i=1}^m \lambda _i \delta_{x^i}=\sum_{i=1}^m  \bigl(
y^i-u_\mathbb{D}(x^i)\bigr) \delta_{x^i},
$$
which yields the system
$$
 \bigl(\alpha+M\bigr)\lambda =\mathbb{Y}. 
$$
This shows that $u_\mathbb{D}(x^i)=y^i-\alpha \lambda_i$ and, in
particular, reiterates the point about the convergence as $\alpha\to
0$. In practice, it is more convenient to work with this system in
order to compute $u_\mathbb{D}$.  
\end{rem}
While the parameter $\alpha\geq0$ plays an important role, it will be
dropped from the notation from now on. The understanding is that its
value can be inferred from the context and that, whatever its value is,
it is kept fixed. In this paper we are particularly focussed on the
case $n=1$ and on the
trivial value set $\mathbb{Y}=\mathbbm{1}$, where $y^i=1$ for
$i=1,\dots, m$.
\begin{deff}
If $\mathbb{Y}=\mathbbm{1}$, we say that $u_\mathbb{D}$ is the
(continuous) {\em signal} generated by the data $\mathbb{X}$. We
sometimes denote it by $u_\mathbb{X}$ or $u_{\mathbb{X},\mathbbm{1}}$.
\end{deff}
The signal is the inexact interpolation of the characteristic function
of the data set\footnote{We observe, in particular, that, if the data set discretizes a
set $S$ of measure 0, then then its characteristic function $\chi_S$ is the trivial
function and of not much use.}. It can be strong, if $u_\mathbb{X}(x^i)\simeq 1$,
$i=1,\dots,m$. This is the case, as was observed above, when
$\mathbb{X}$ is a fine and locally filling discretization of the ambient
space, such as when approximating a set of positive measure by a set of
discrete points. More often, however, the signal will be weak in the
sense that $u_\mathbb{X}(x^i)$ is significantly less than 1 for
$i=1,\dots,m$. In this paper we contend that 
the usefulness of the signal $u_\mathbb{X}$ does not only stem from
its approximation or interpolation properties, but also (and perhaps
mainly) from the fact that
most of its level sets are very smooth due to Proposition
\ref{c-infty} and, in fact, deliver smooth manifold approximations of
$\mathbb{X}$ that can effectively be employed as stable decision
boundaries in supervised classification problems. Superlevel sets of
the signal are reliable approximations with positive measure of any discrete
data set that prove robust against noise. They, in a sense, connect the
dots and capture the shape of the data. Thus the main philosophical
difference between the traditional view point, that considers the data
as the manifestation of a function that needs to be reconstructed, and
the view point taken in this paper is that here the data set itself is
approximated by the mostly smooth (super)level sets of a function that
may not even fit the data well at all. We are indeed more interested in the
level surfaces generated by the data's signal than we are in its values.
It will be shown that data signals, whether they are strong or weak,
can be succesfully exploited in this sense. The practical experiments
run in the next section will make use the following proposition.
\begin{prop}
Let $\mathbb{X}_0=\mathbb{X}_1\dot\cup\,
\mathbb{X}_2\dot\cup\cdots\dot\cup\,\mathbb{X}_N$
(the notation means that the union is disjoint) be a given data set
consisting of $N\in \mathbb{N}$ subsets (or classes). For
\begin{gather*}
\mathbb{D}_0 =\big\{ (x^i,1)\,\big |\, i=1,\dots,|\mathbb{X}_0|\big\},\\
\mathbb{D}_l=\big\{ (x,1)\,\big |\, x\in \mathbb{X}_l\big\}\cup \big\{
(x,0)\,\big |\, x\in\bigcup_{k\neq l}\mathbb{X}_k\big\},\: l=1,\dots, N,
\end{gather*}
it holds that
$$
u_{\mathbb{D}_0} = \sum_{l=1}^N u_{\mathbb{D}_l}.
$$
\end{prop}
\begin{proof}
The signal $u_{\mathbb{D}_0}$ and the signals
$u_{\mathbb{D}_1},\dots,u_{\mathbb{D}_N}$ relative to $\mathbb{X}_0$,
are solutions of the linear equations
$$
 A_d\, u +u\big |_{\mathbb{X}_0}\cdot
 \delta_{\mathbb{X}_0}=\mathbb{Y}_l\cdot \delta_{\mathbb{X}_0},\: l=0,\dots,N, 
$$
where $A_d= \frac{\alpha}{c_d} (1-\Delta)^{\frac{d+1}{2}}$ and where
$$
u\big |_{\mathbb{X}_0}\cdot\delta_{\mathbb{X}_0}=\sum_{i=1}^{|\mathbb{X}_0|}u(x^i)\delta_{x^i},\:
\mathbb{Y}_l\cdot \delta_{\mathbb{X}_0}=\sum_{i=1}^{|\mathbb{X}_0|}y^i_l\delta_{x^i}. 
$$
The claim therefore follows from the fact that
$$
 \sum_{l=1}^N\mathbb{Y}_l=\mathbb{Y}_0.
$$
\end{proof}
\begin{rem}
Notice that signals do not, however, behave additively, in the sense
that
$$
 u_{\mathbb{X}_1}+u_{\mathbb{X}_2}\neq u_{\mathbb{X}_1\cup \mathbb{X}_2},
$$
in general, even when $\mathbb{X}_1\cap \mathbb{X}_2=\emptyset$ 
\end{rem}
If one is given a labeled data set $\mathbb{X}_0$, where $N$ is the
number of labels and $\mathbb{X}_l$, $l=1,\dots,N$ are the subsets
consisting of the data corresponding to label $l$, i.e. $L(x)=l$ for
$x\in \mathbb{X}_l$, then one obtains a
classification algorithm by computing the relative signals
$u_{\mathbb{X}_l}$ for $l=1,\dots,N$ and assigning any unlabeled datum
$z\in \mathbb{R}^d$ to the class that exhibits the strongest relative
signal, that is,
\begin{equation}\label{algo}
 L(z)=\operatorname{argmax}_{l}u_{\mathbb{X}_l}(z).
\end{equation}
\begin{rem}
An important aspect of the proposed approach (and of kernel based
methods as well) is that the fundamental solution
\begin{equation}\label{fs}
G(x)=\exp\bigl( -2\pi|x|\bigr),\:x\in \mathbb{R}^d,
\end{equation}
of the (pseudo)differential operator
$\frac{1}{c_d}(1-\Delta)^\frac{d+1}{2}$
used to obtain the finite linear system is essentially dimension 
independent. It depends on it only through the Euclidean distance
function, which is a minimal ingredient that can hardly be
avoided.
It would of course be extremly difficult to work directly with the
(pseudo)differential operator or the energy functional in high
dimension. An application to the well-known MNIST classification
problem will be discussed in the next section using an approach based on
the signal generated by the training data. In that case one has that $d=784$.
\end{rem}
\begin{rem}\label{locality}
Just as for exact interpolation and due to the fact there are no
constraints like e.g. boundary conditions, the method is completely
local. This means that an approximant can be computed based on
a subset of the original data set that is confined or restricted to a
subregion of interest. This feature will be exploited in the MNIST
classification problem.
\end{rem}
\begin{rem}
It is well-known that the parameter $\alpha>0$ has a regularizing
effect that can be used to deal with noisy data when performing
interpolation. It turns out that it also helps smooth out the level
sets of $u_\mathbb{D}$. This is also illustrated in the next section.
\end{rem}
\begin{rem}
It is sometimes convenient to modify the decay rate of the
exponential ``basis'' functions, especially if the data undergoes some
initial normalization. This can be done without significant consequences other 
than a slight modification of the objective functional
or, equivalently, of the corresponding 
differential operator. Indeed, for $\gamma>0$ and using the well-known
scaling and translation properties of the Fourier transform
$\mathcal{F}_d$, it holds that
\begin{align*}
\Bigl( \frac{1}{c_d}(1-\gamma^2\Delta)^\frac{d+1}{2}\Bigr) u \bigl(
  \frac{\cdot_x-y}{\gamma}\bigr)&=\frac{1}{c_d}\mathcal{F}^{-1}_d \bigl(
  1+\gamma^2|\cdot_\xi|^2\bigr) ^{\frac{d+1}{2}}\mathcal{F}_d
  \Bigl(\tau_{y}\,\sigma _{1/\gamma}u\Bigr)\\&=
  \frac{1}{c_d}\mathcal{F}^{-1}_d \Bigl(\bigl( 1+|\gamma\cdot_\xi|^2\bigr) ^{\frac{d+1}{2}}
  \exp\bigl(-i\frac{y}{\gamma}\cdot \gamma\cdot_\xi\bigr)\gamma^n\,\widehat
  u(\gamma \cdot_\xi)\Bigr)\\&=
   \frac{1}{c_d}\mathcal{F}^{-1}_d \Bigl[ \gamma^n \sigma_\gamma
  \Bigl(  \exp\bigl(-i\frac{y}{\gamma}\cdot\cdot_\xi \bigr) \bigl(
  1+|\cdot_\xi|^2\bigr) ^{\frac{d+1}{2}}\widehat{u}\Bigr)\Bigr]\\&=
  \sigma_{1/\gamma}\Bigl( \frac{1}{c_d} (1-\Delta) ^{\frac{d+1}{2}}u
  \Bigr)\bigl( \cdot_x-\frac{y}{\gamma}\bigr)\\&= \Bigl( \frac{1}{c_d} (1-\Delta) ^{\frac{d+1}{2}}u
  \Bigr)\bigl(\frac{\cdot_x-y}{\gamma}\bigr).
\end{align*}
Here we use the notation $\widehat{u}=\mathcal{F}_d(u)$ for the Fourier transform of
the function $u$ as well as $\cdot _x$ and $\cdot_\xi$ as place holders for the
independent variables $x$ and $\xi$ in order to distinguish a function
from its values. Furthermore $\tau_y$ denotes tanslation, that is,
$\bigl(\tau_yu\bigr)(\cdot)=u(\cdot-y)$, and $\sigma_\gamma$ scaling,
or $\bigl(\sigma_\gamma u\bigr)(\cdot)=u(\gamma\cdot)$.
Replacing $u$ by $\exp \bigl( -2\pi|\cdot|\bigr)$ and
$y=x^i$ for any $i=1,\dots,m$ it is seen that
$$
 \frac{1}{c_d}(1-4\pi^2\Delta)^\frac{d+1}{2}\exp \bigl( -|\cdot
 -x_i|\bigr) =\delta_{x^i}.
$$
Thus the use of the modified exponentials merely corresponds to an
inconsequential modification of the Euler-Lagrange equation (or its generating
functional). 
\end{rem}
\section{Numerical Experiments}
In this section a series of experiments are performed to illustrate
the effectiveness of the method described in the previous
section. First we consider two dimensional problems to highlight
important aspects and in order to motivate and justify the use of the
method in a high dimensional context. The section then concludes with an
application to the classification of the MNIST data set.
\subsection{Stability of Signals' Level Sets}
Working in the context of approximating measurable functions, simple
functions play an important role as they are the building block of any
measurable function. While the approximation property is well-known we
present a few examples to illustrate the efficacy of the use of the
data signal's level sets for classification purposes. First we will
consider situations where the approximation is good, then examples when
it is rather poor. It will be shown that, in all cases, i.e.,
regardless of how good the approximation is, the signal's level sets
still provide very useful information. This observation is crucial
since it opens the door to applications to high dimensional data,
where it is inconceivable that the data arguments $\mathbb{X}$ provide
a fine grid of even a small portion of the ambient space.
\subsection{Characteristic Functions of Sets of Positive Measure}
\begin{figure}
  \includegraphics[scale=.35]{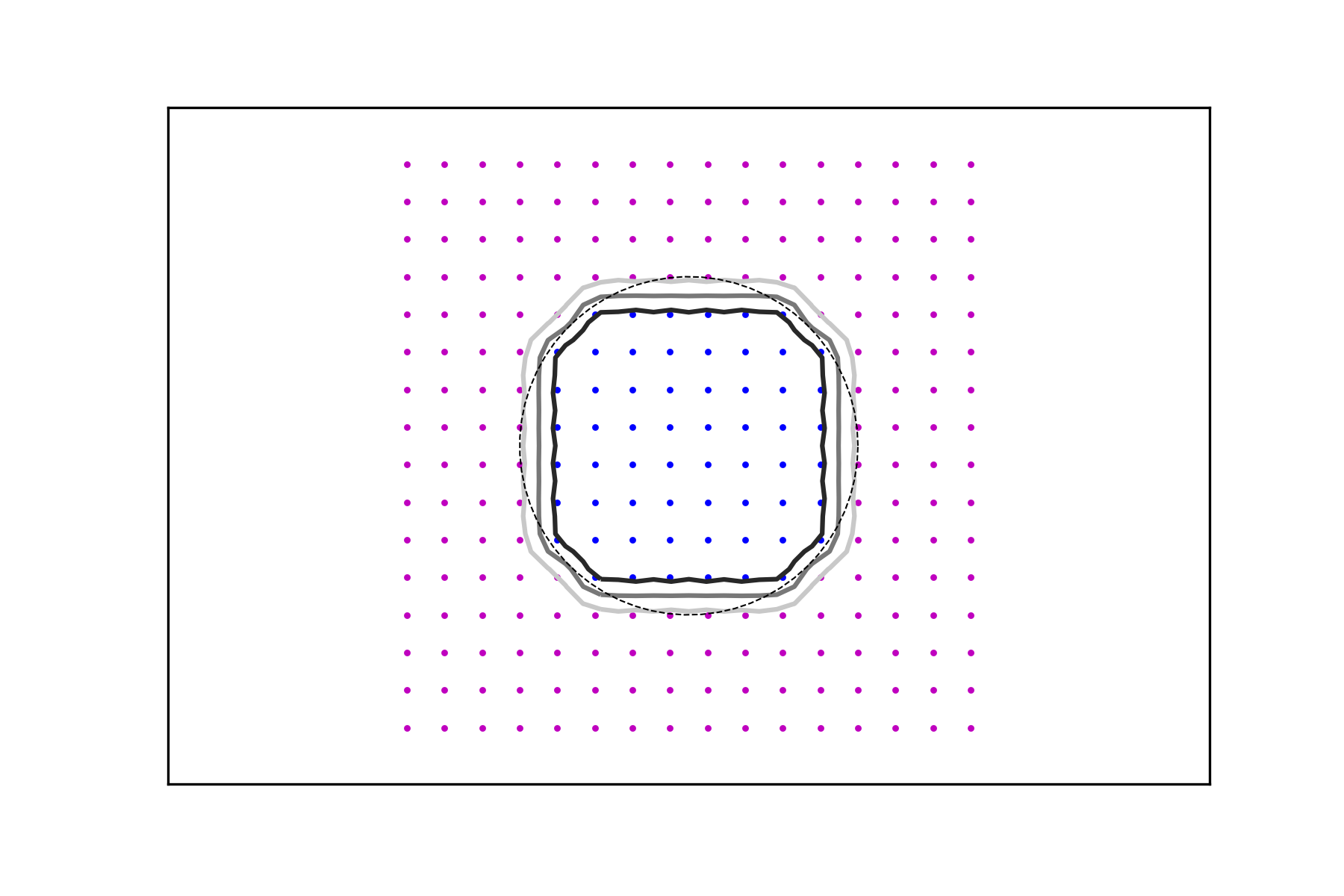}
  \includegraphics[scale=.35]{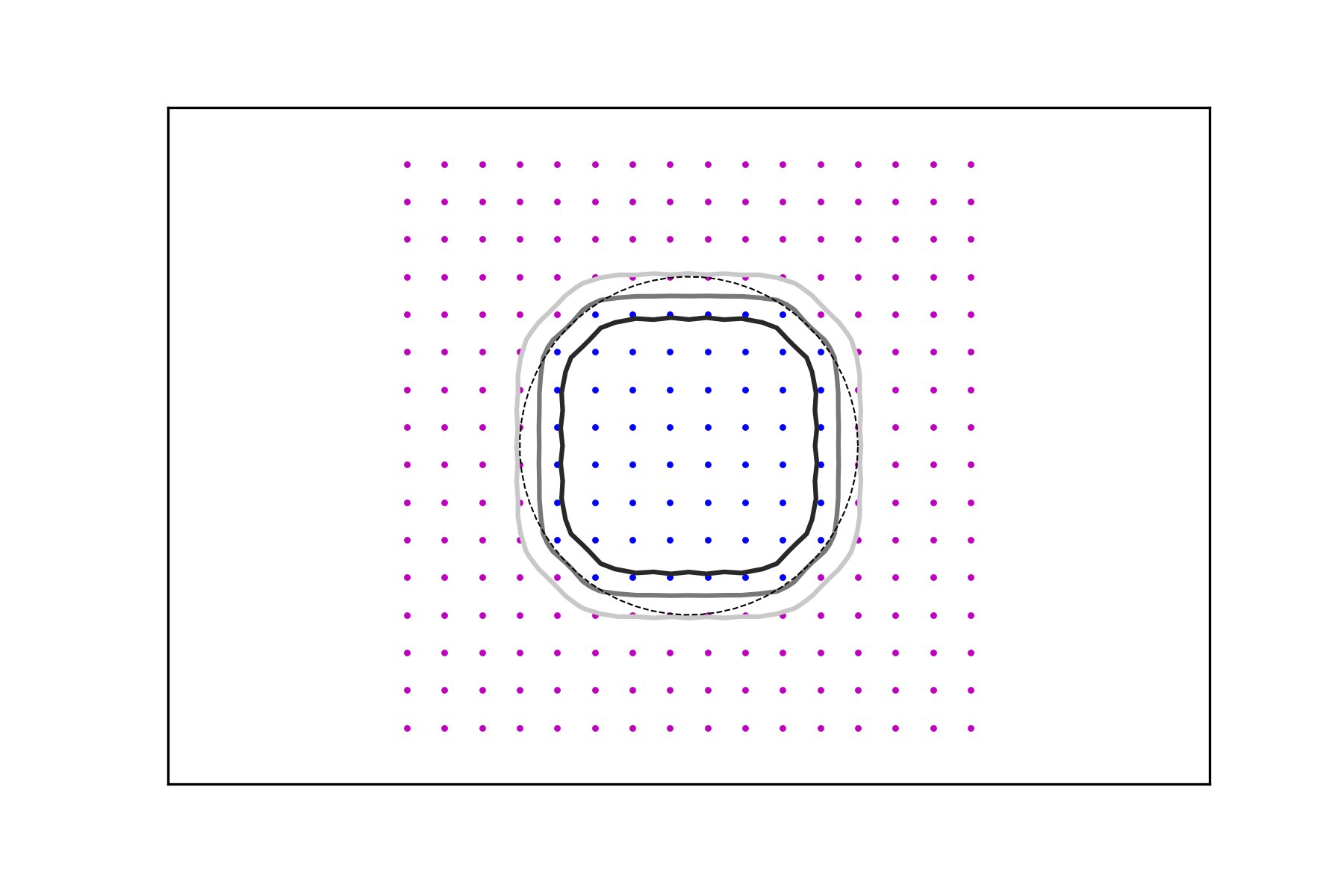}
  \includegraphics[scale=.35]{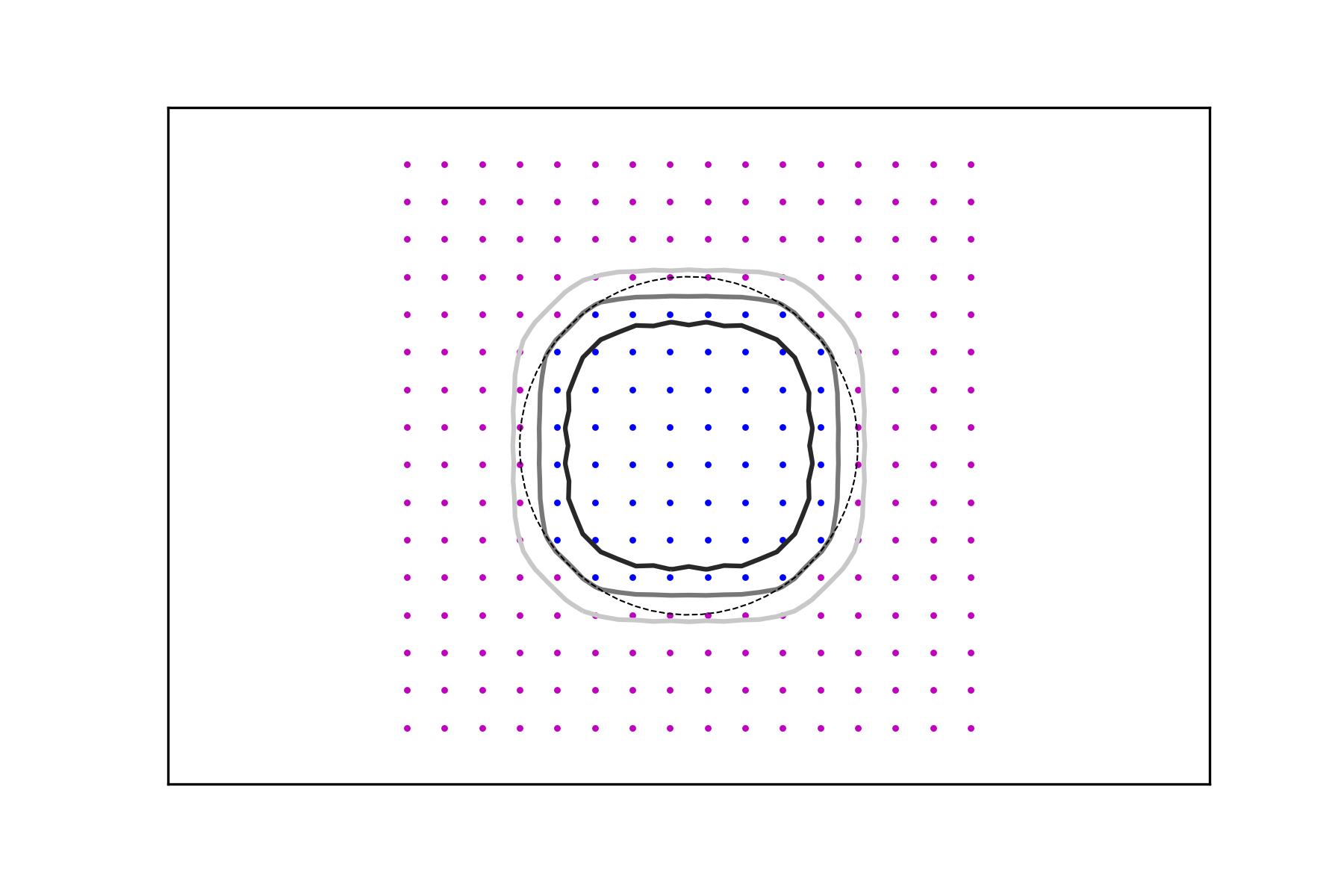}
  \includegraphics[scale=.35]{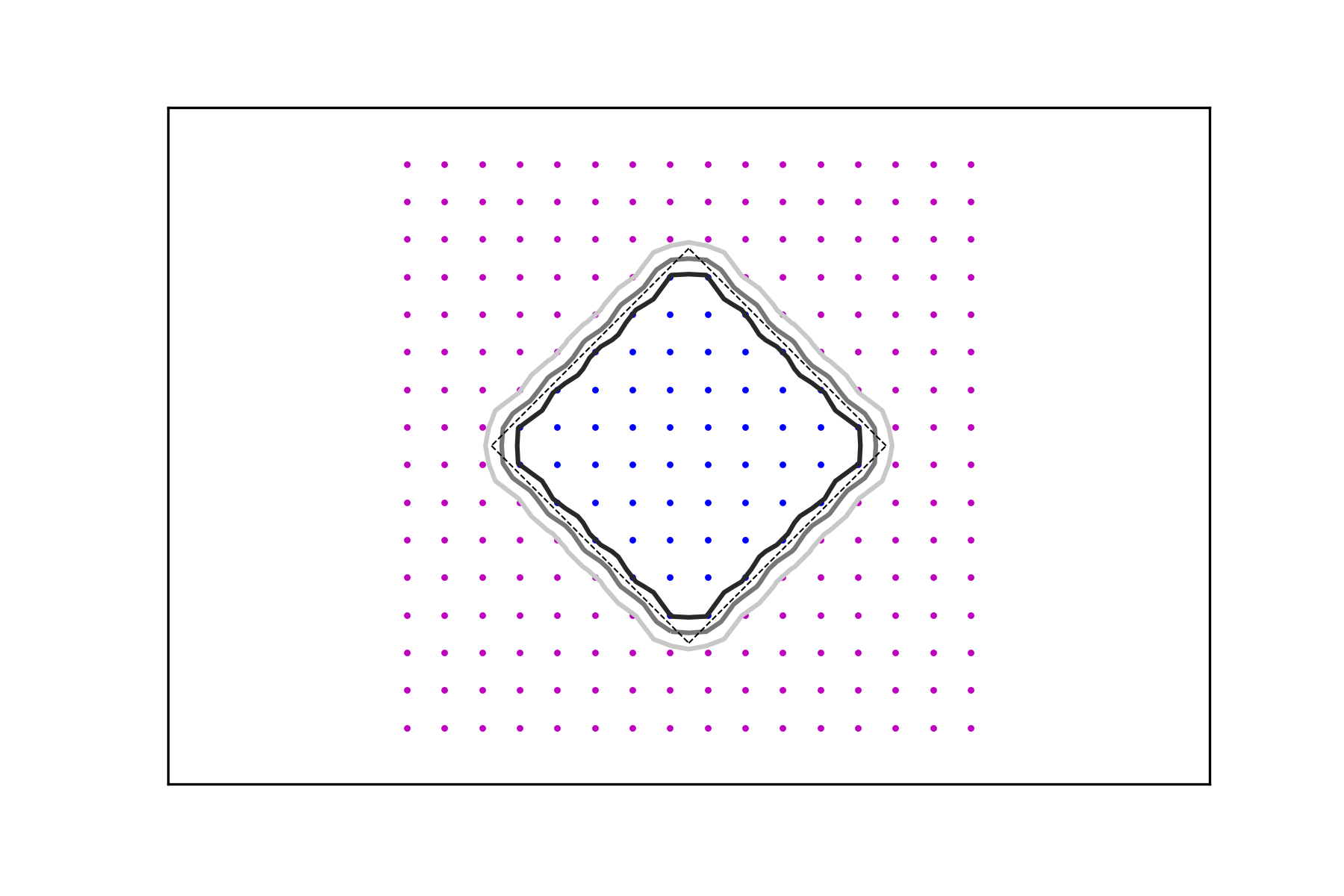}
  \includegraphics[scale=.35]{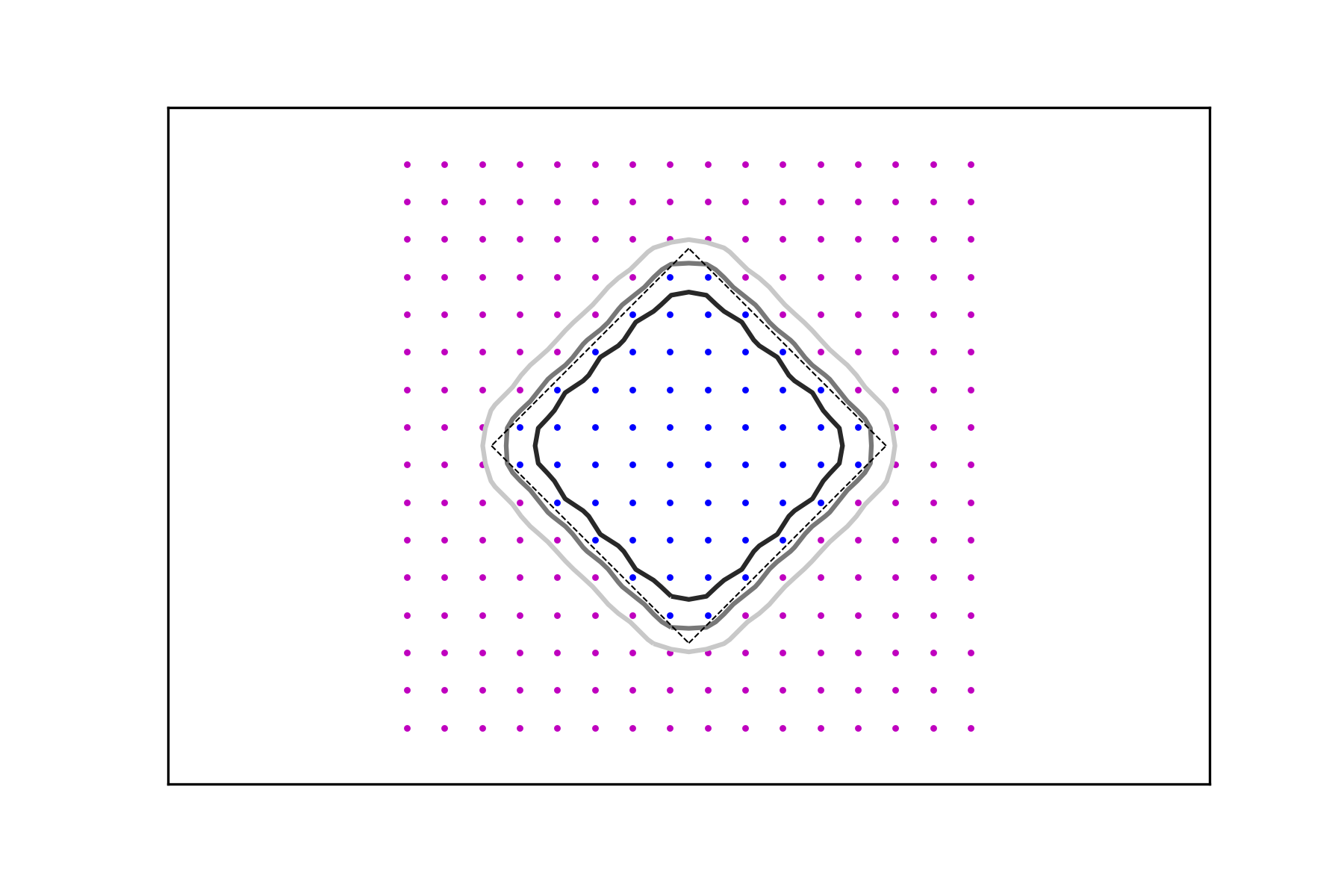}
  \includegraphics[scale=.35]{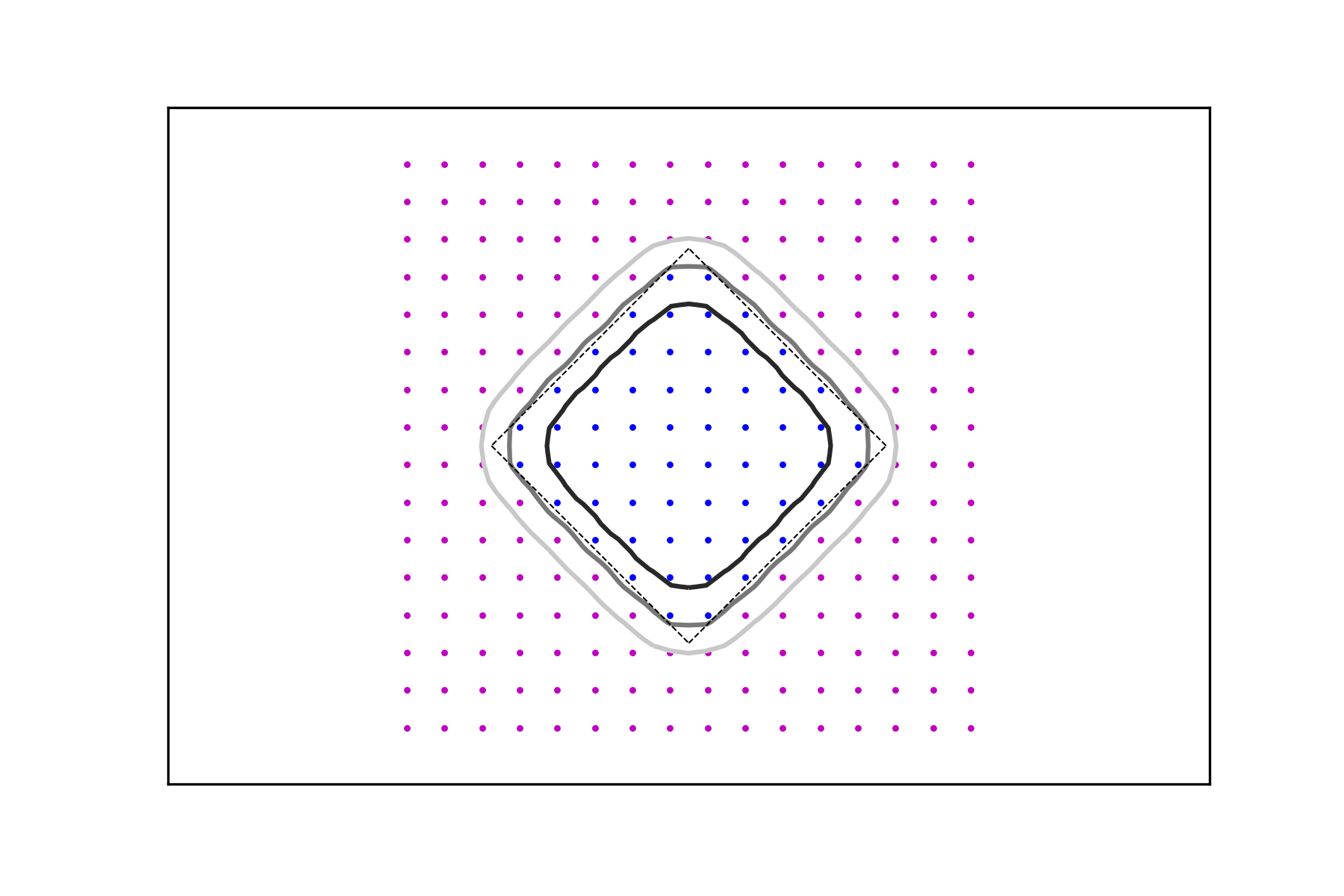}
    \includegraphics[scale=.35]{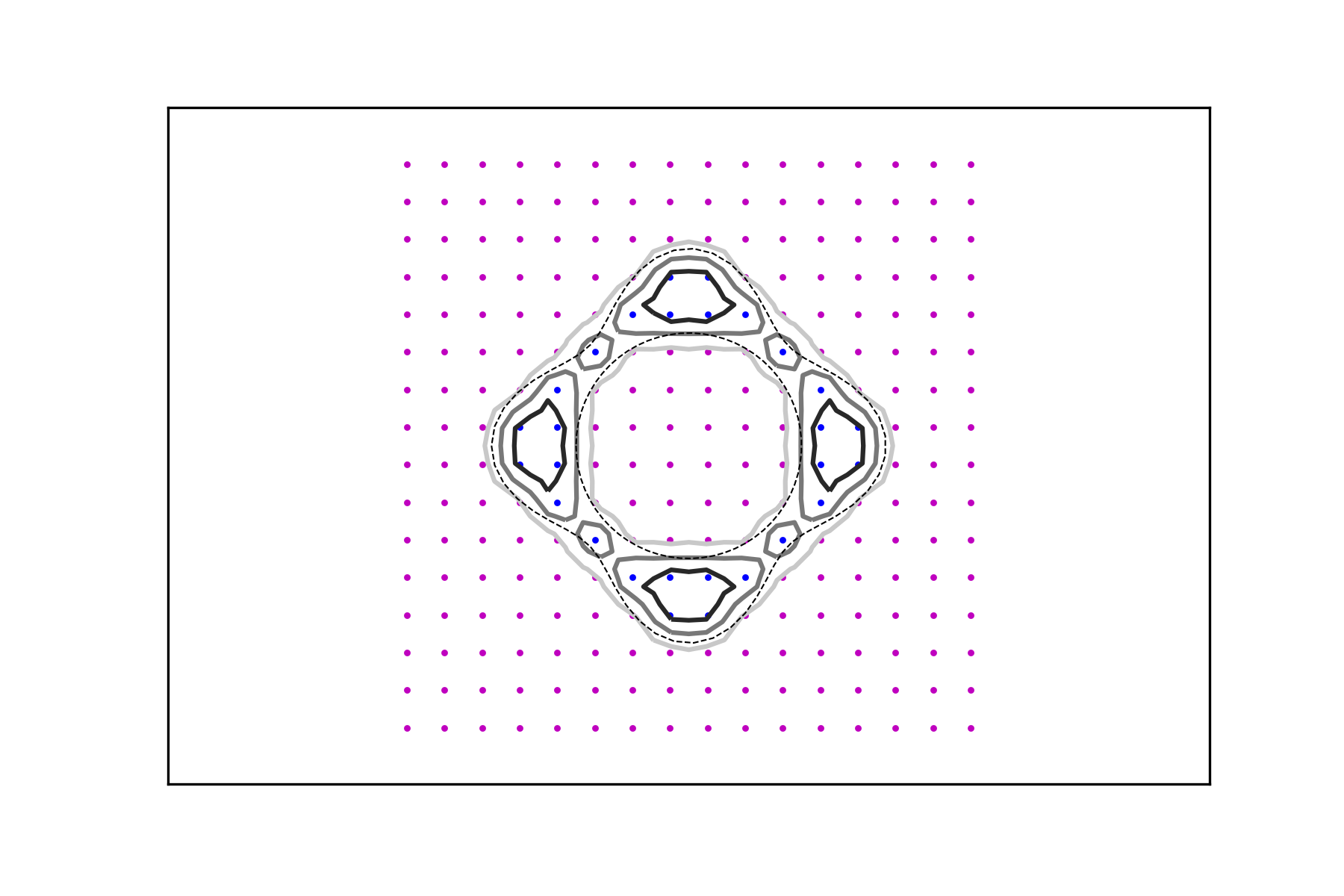}
  \includegraphics[scale=.35]{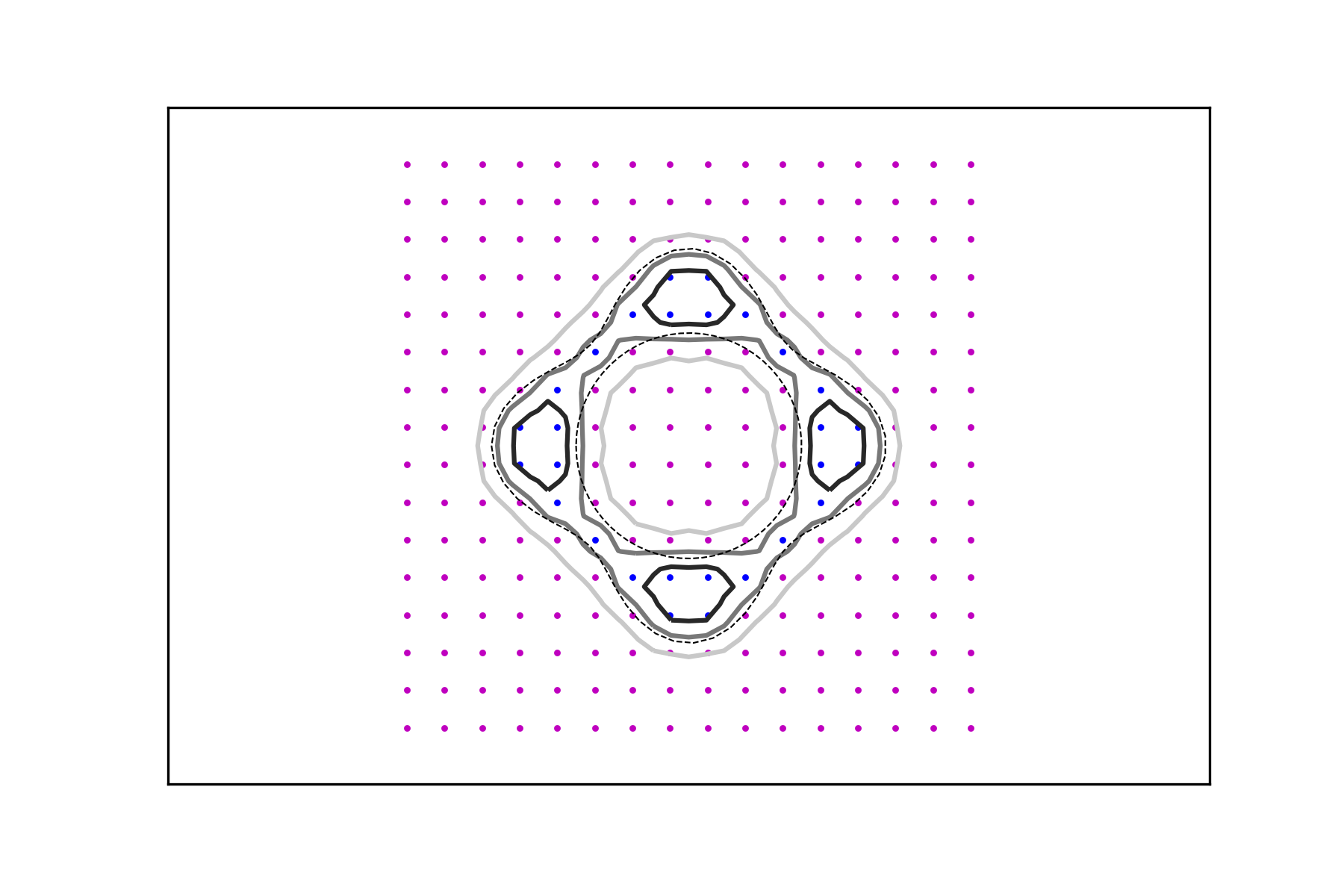}
  \includegraphics[scale=.35]{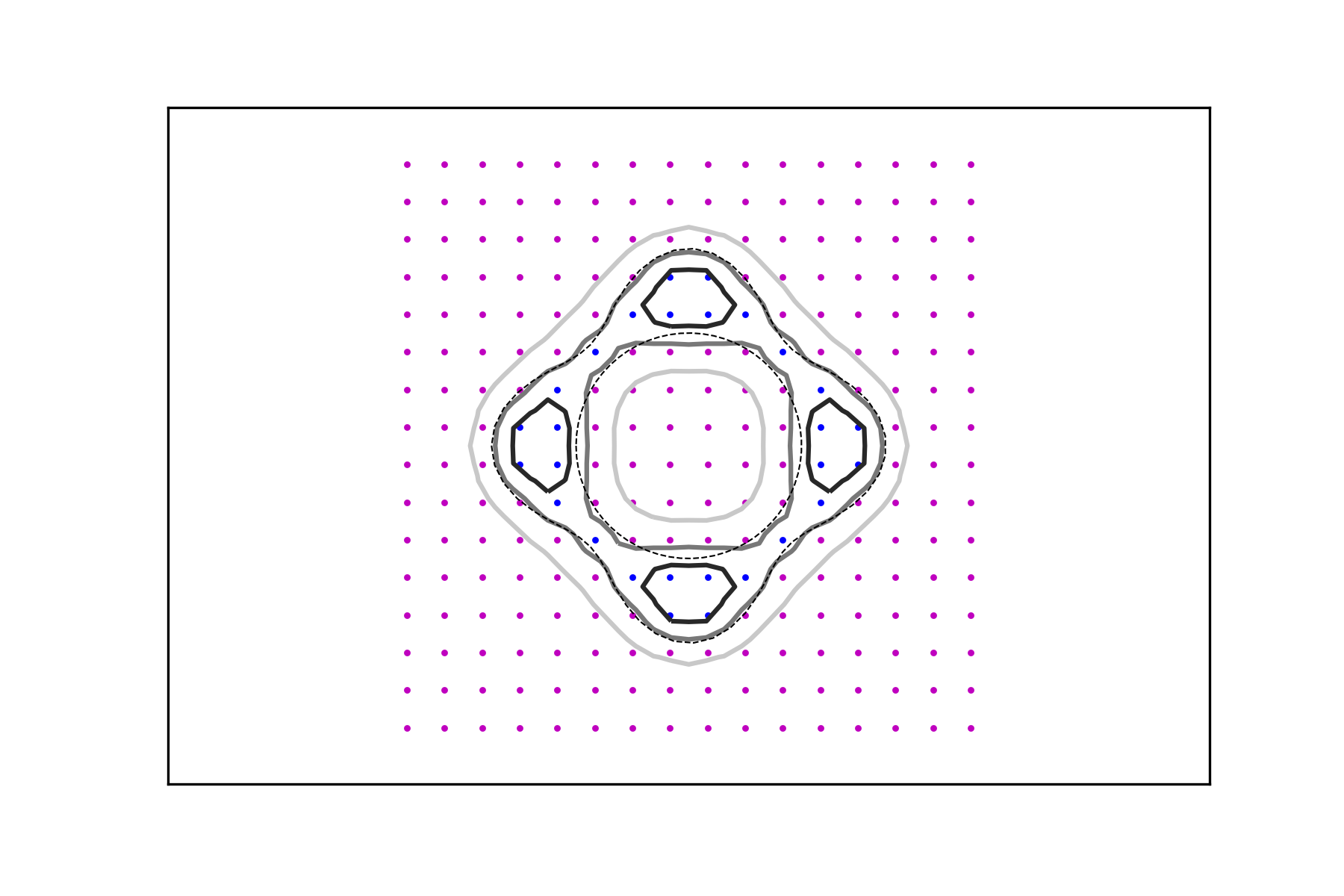}
  \caption{Level lines of the signal $u_{\mathbb{D}_{m,j}}$ for
    $m=16$, $j=1,2,3$, and $\alpha=0,1,1,2$. Depicted are the
    data set (blue dots correspond to the value 1 while magenta dots to
    the value 0) and three level lines corresponding to levels at
    20\%, 50\%, and 80\% of the signal's maximal value,
    respectively. Darker lines correspond to higher levels. The
    parameter $\alpha$ grows from left to right. The
    boundary of the set $S_j$ appears as a dashed black line. }
  \label{fig:noNoise}
\end{figure}

\begin{figure}
  \includegraphics[scale=.35]{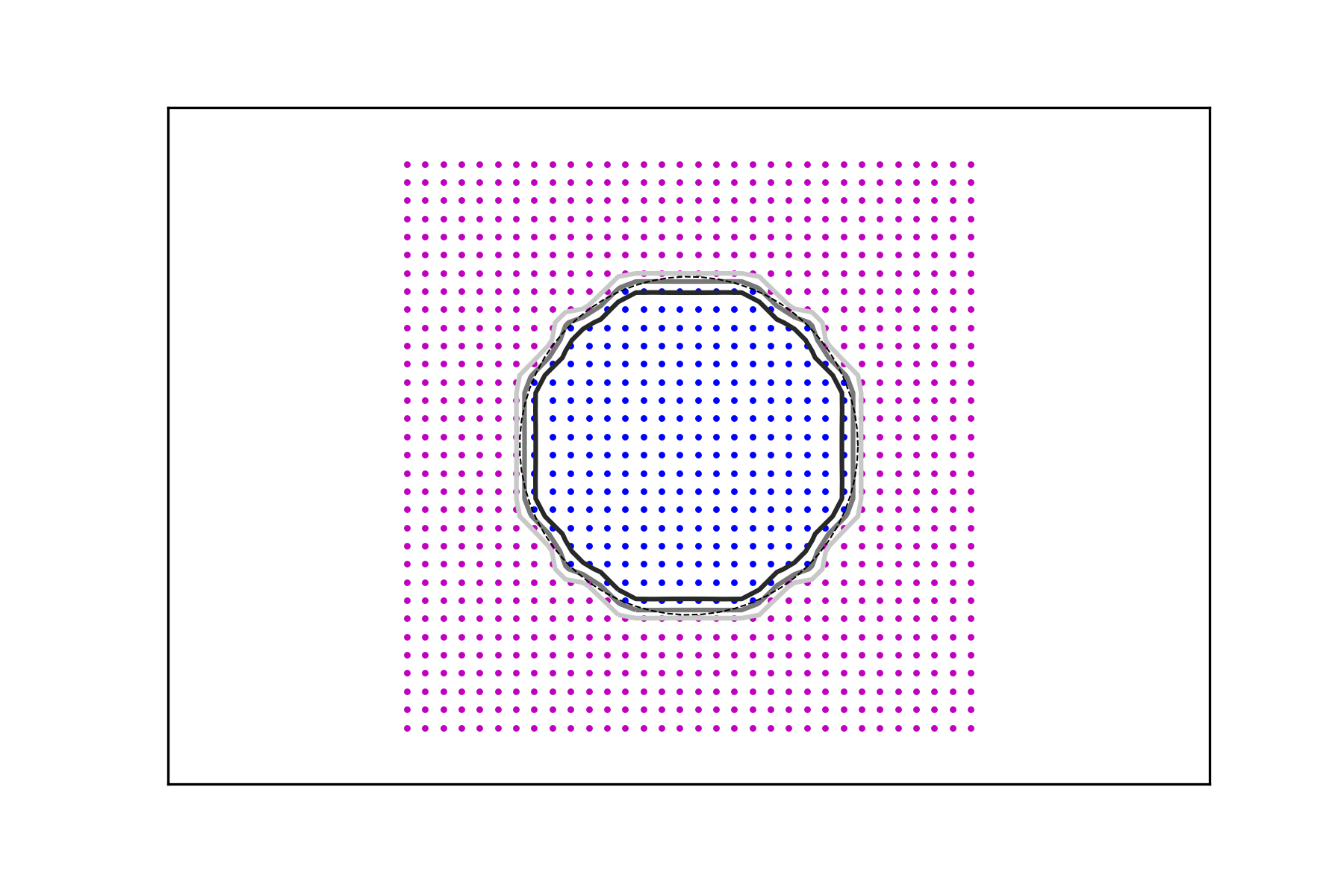}
  \includegraphics[scale=.35]{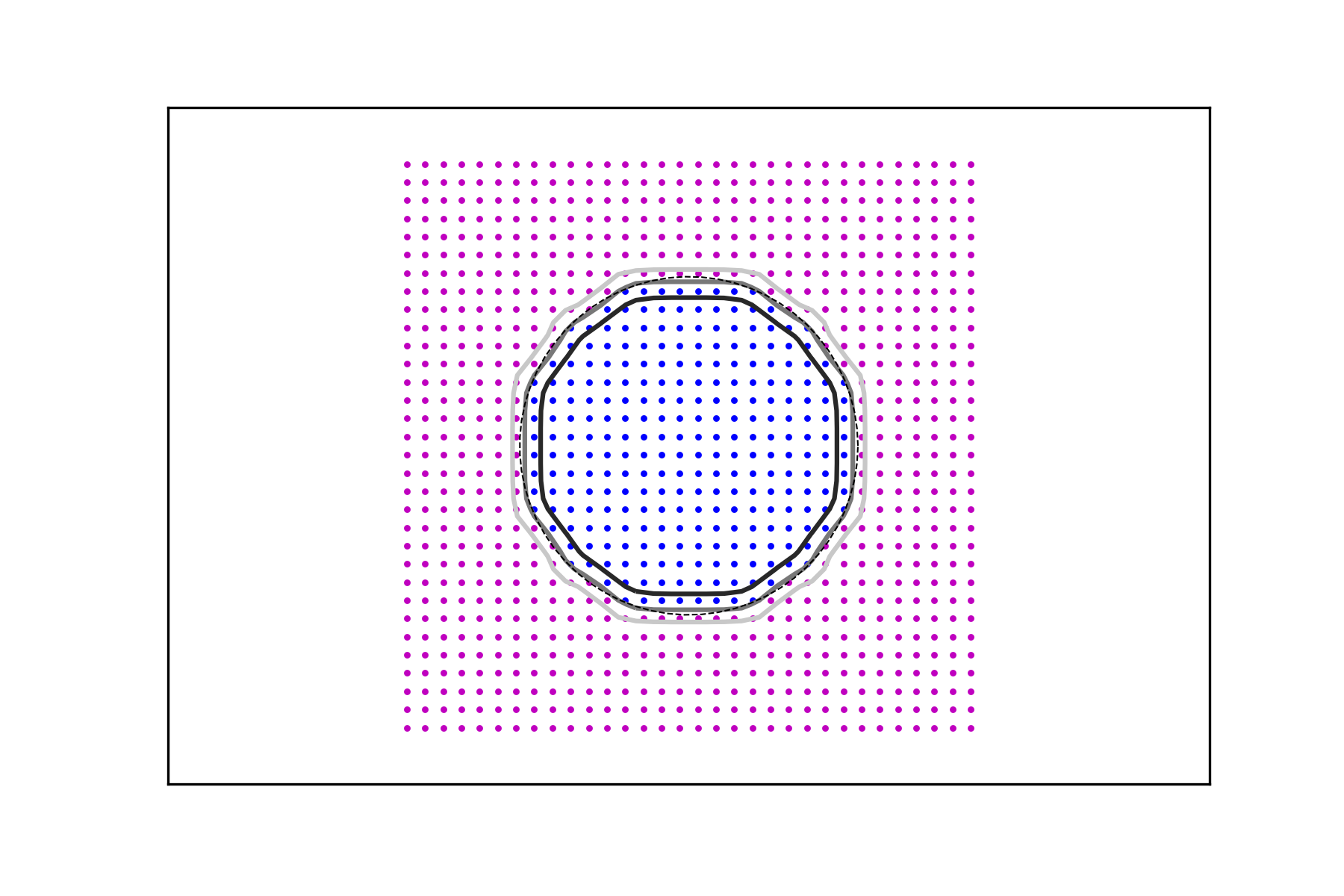}
  \includegraphics[scale=.35]{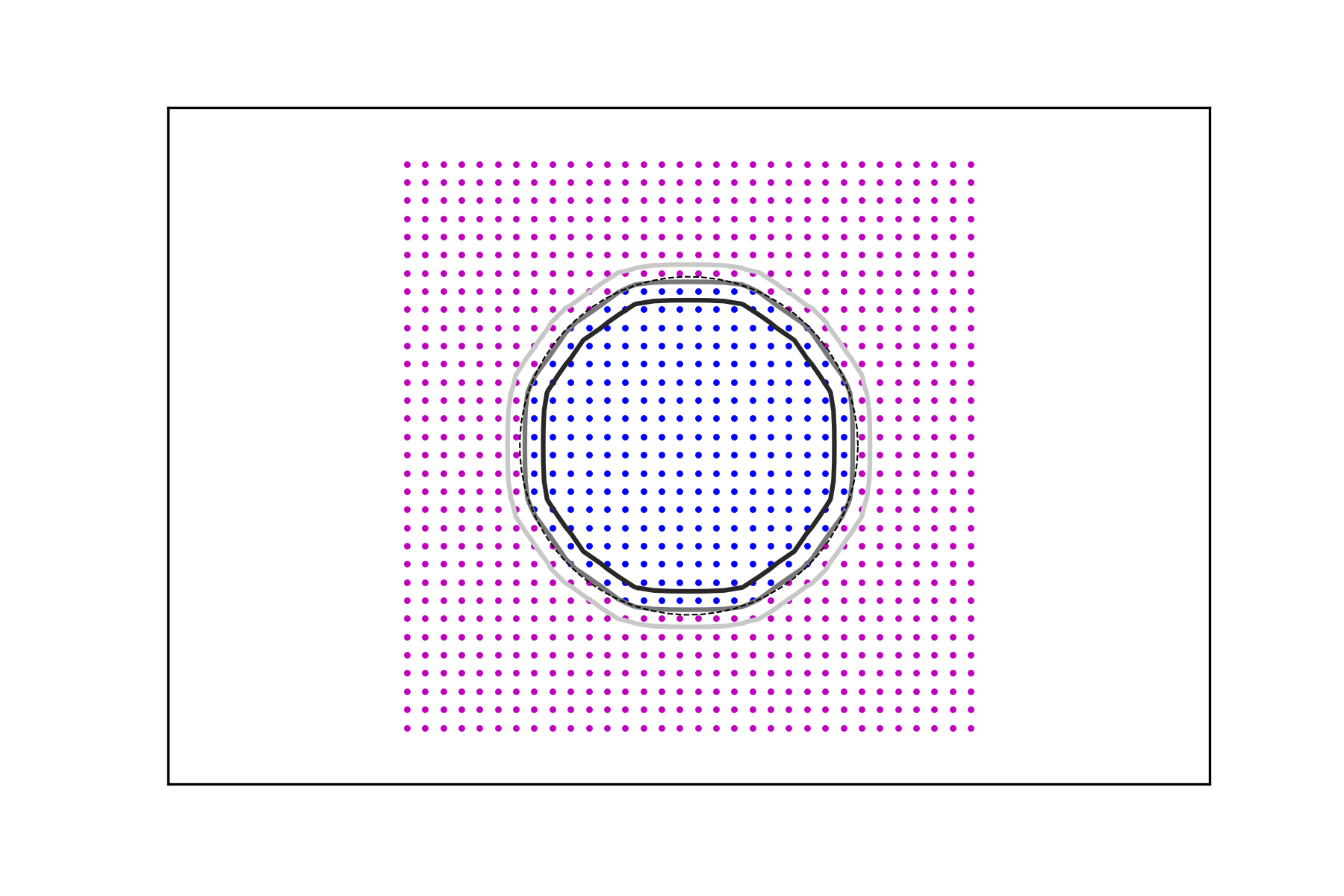}
  \includegraphics[scale=.35]{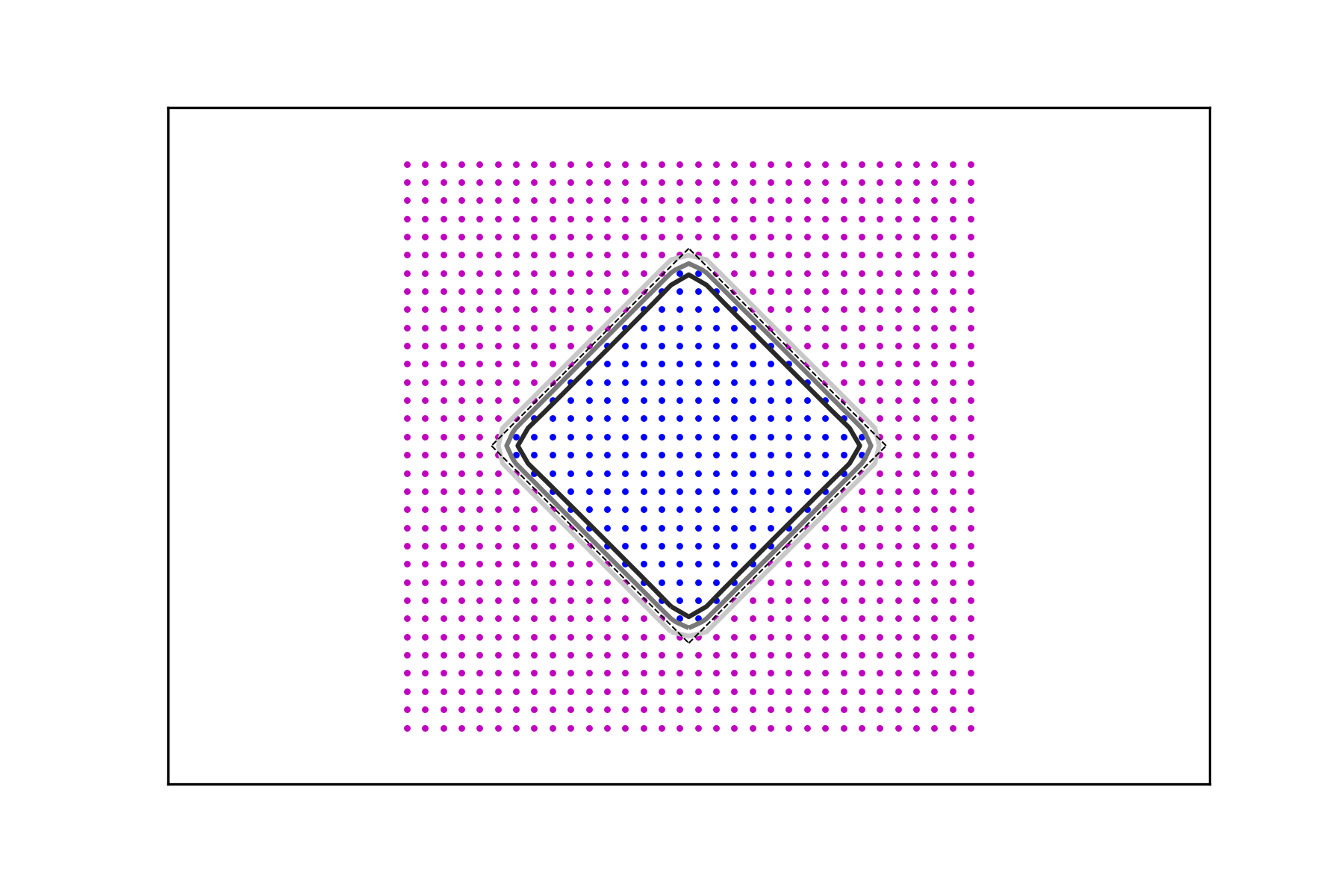}
  \includegraphics[scale=.35]{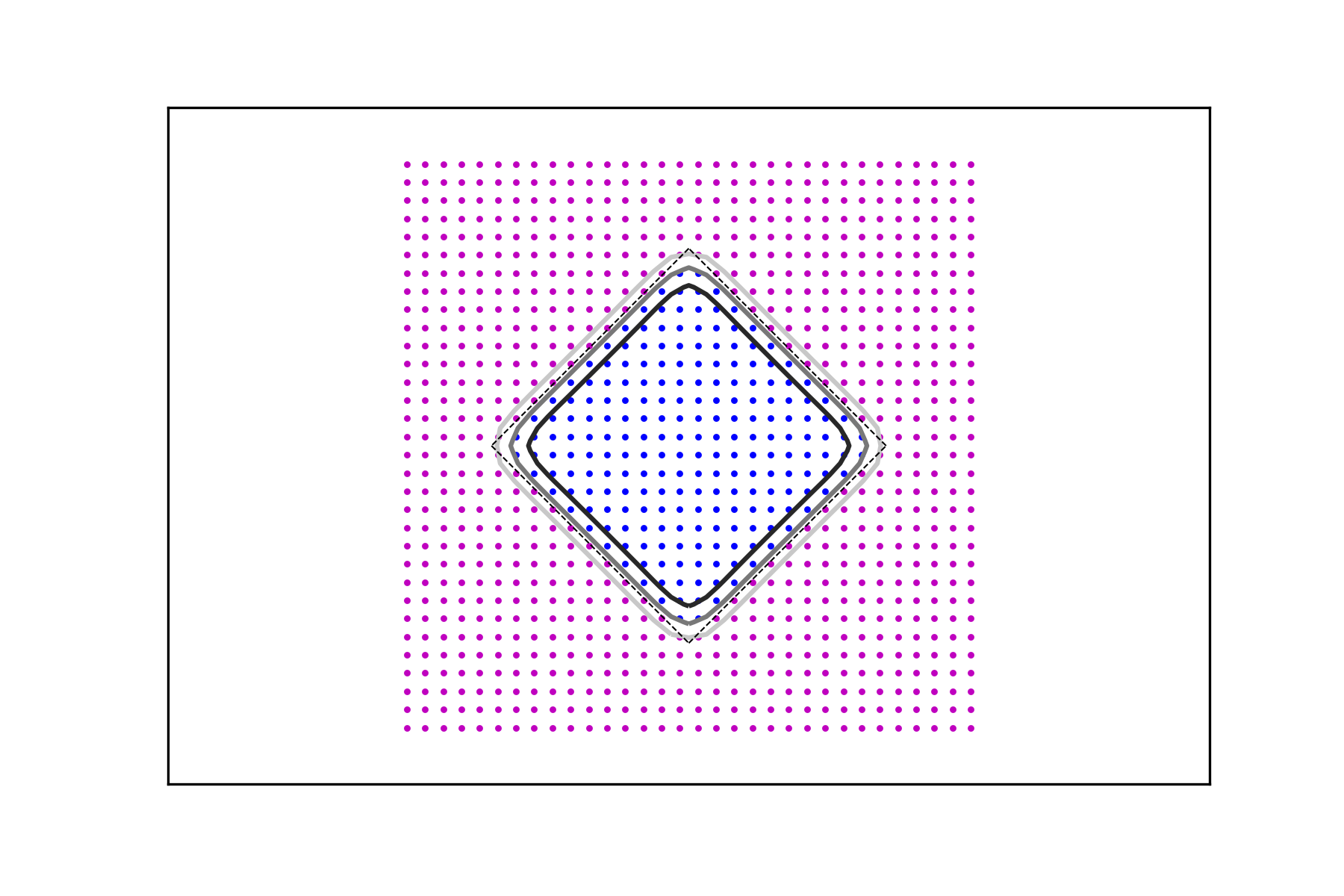}
  \includegraphics[scale=.35]{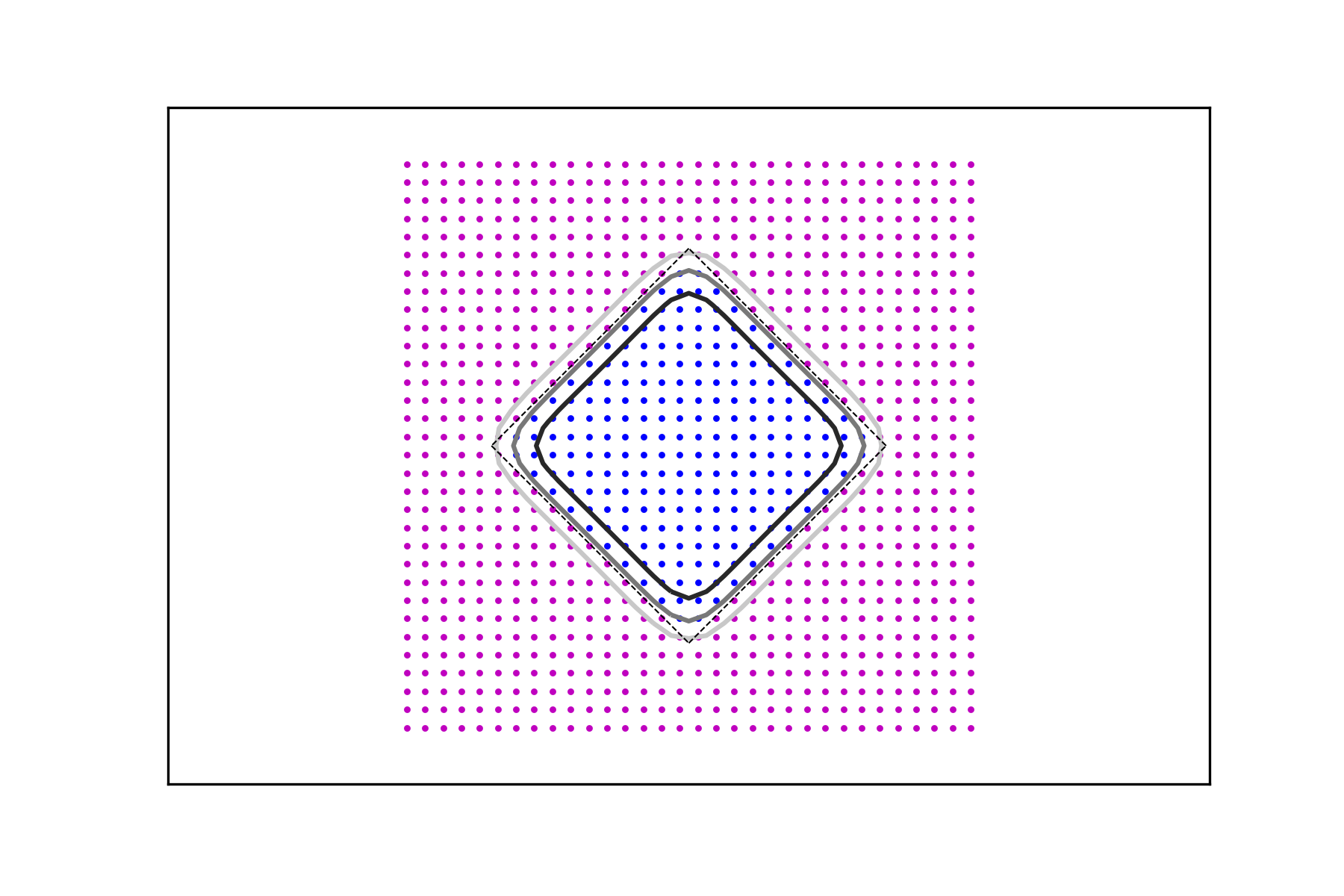}
    \includegraphics[scale=.35]{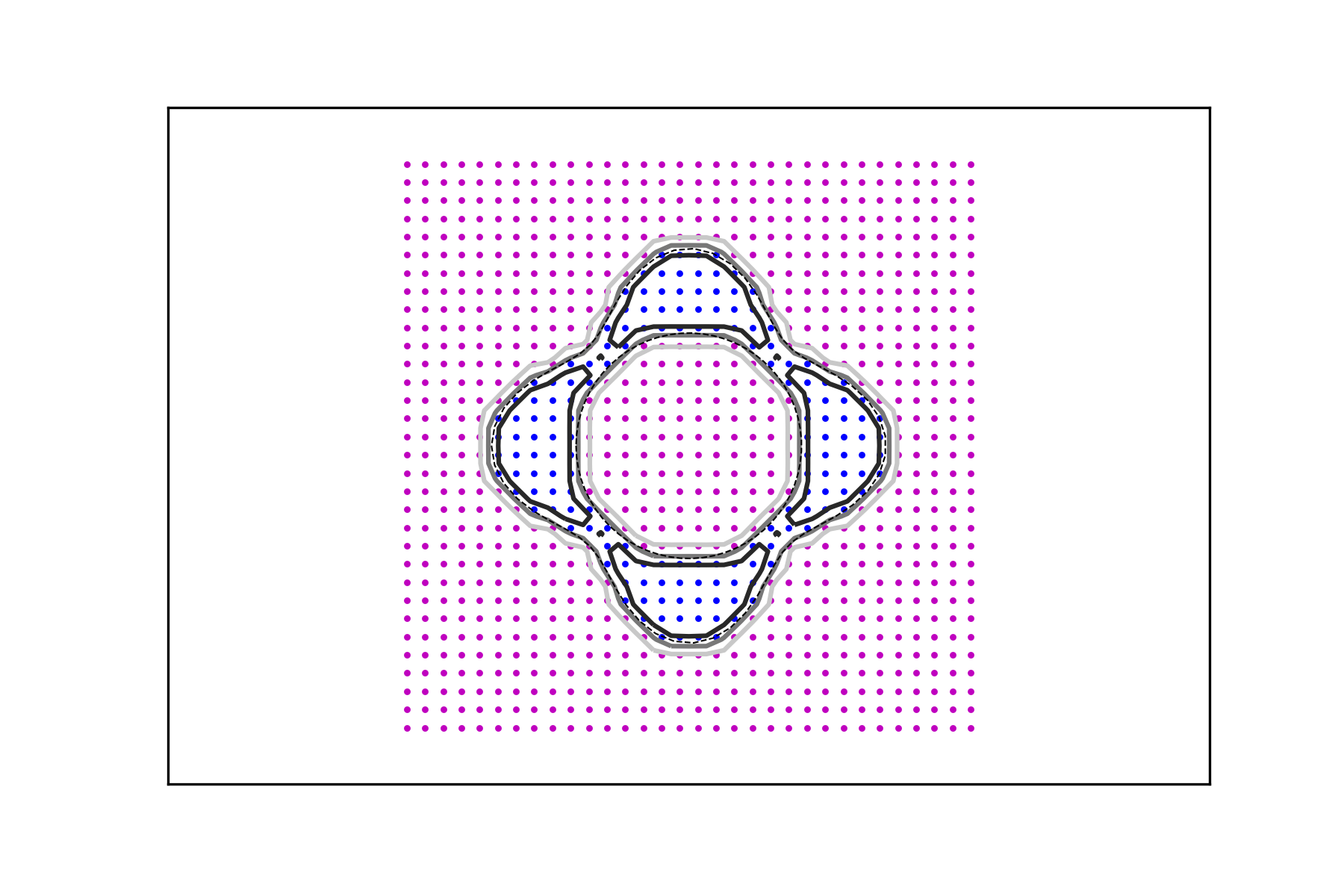}
  \includegraphics[scale=.35]{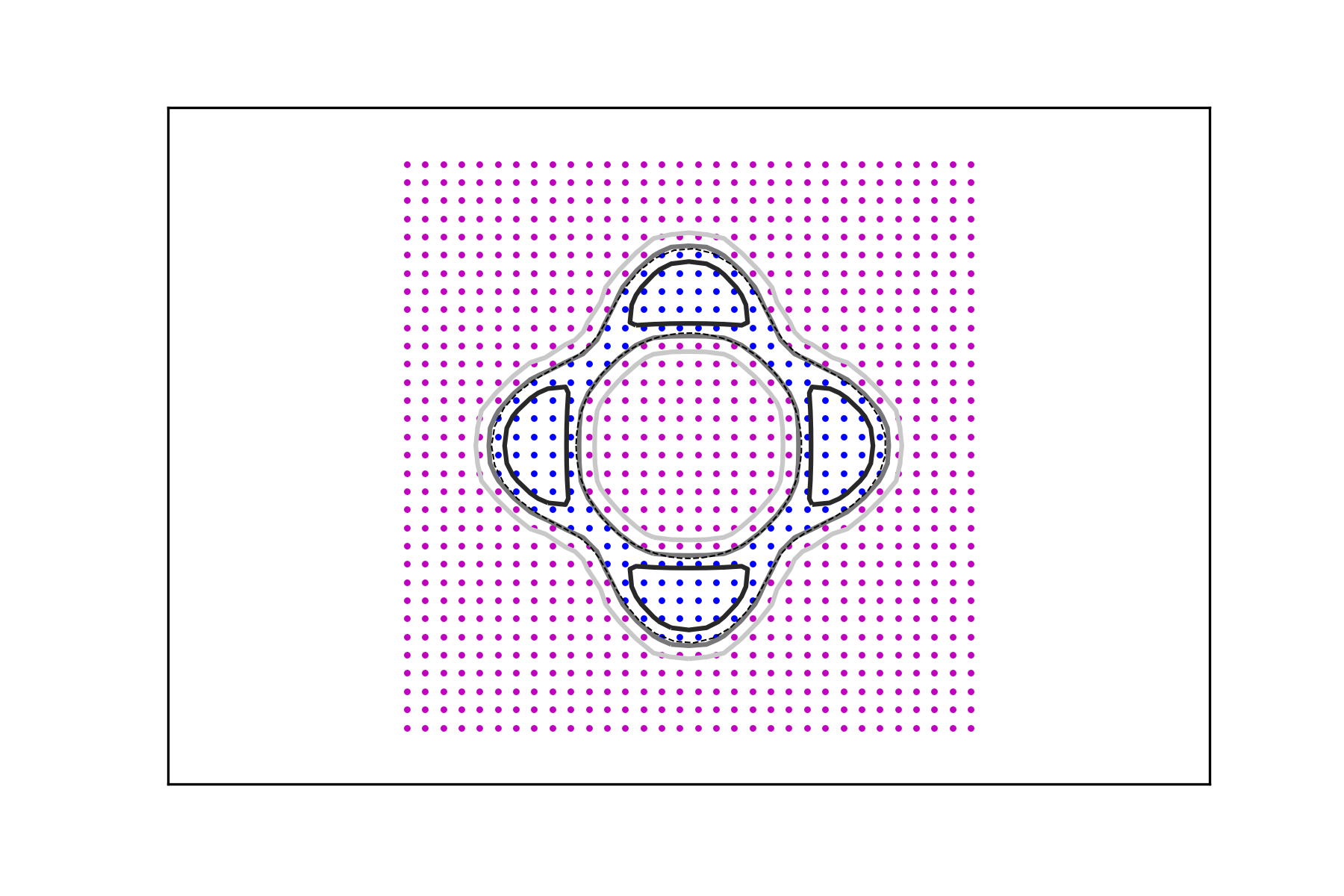}
  \includegraphics[scale=.35]{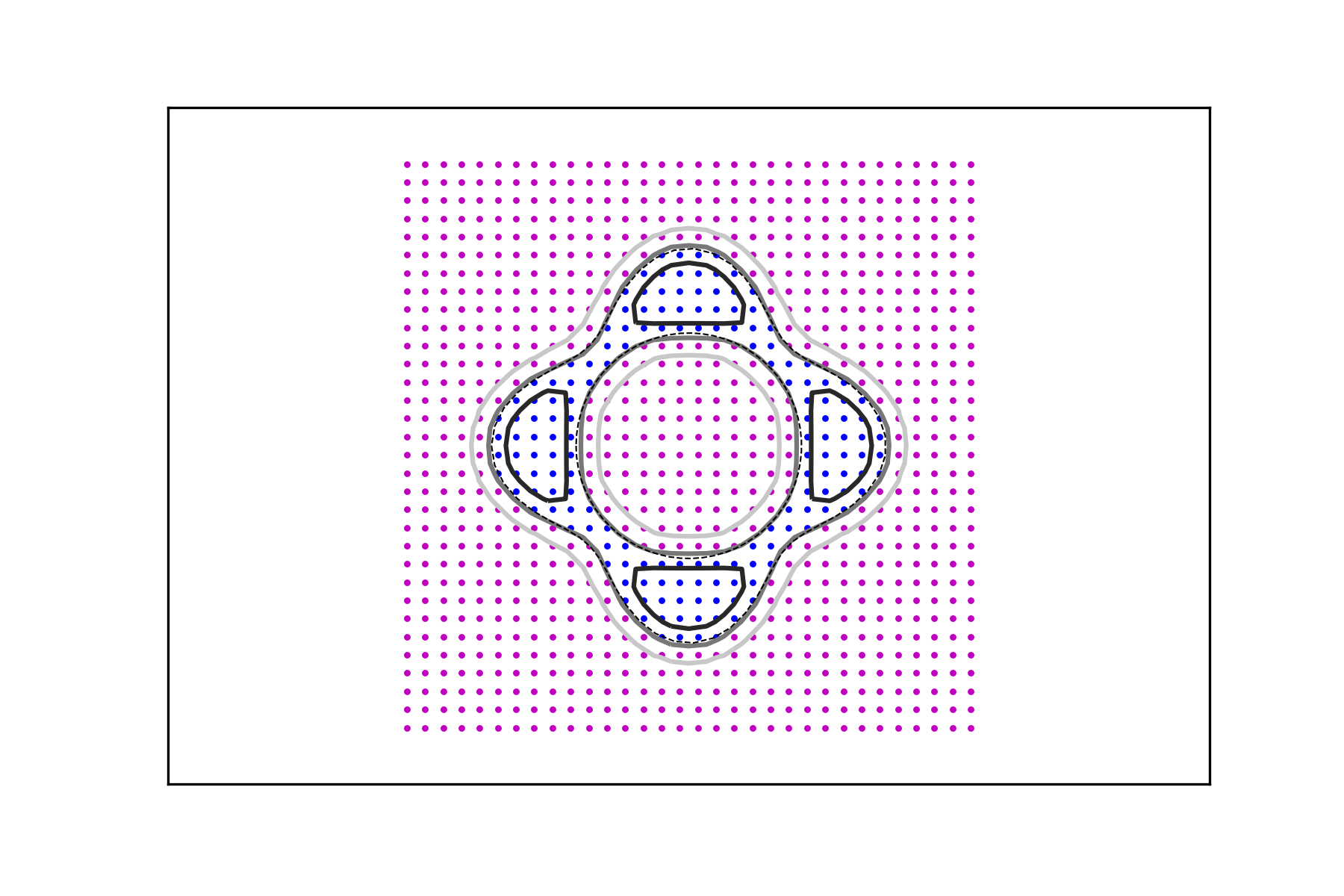}
  \caption{Level lines of the signal $u_{\mathbb{D}_{m,j}}$ for
    $m=32$, $j=1,2,3$, and $\alpha=0,1,1,2$. Depicted are the
    data set (blue dots correspond to the value 1 while magenta dots to
    the value 0) and three level lines corresponding to levels at
    20\%, 50\%, and 80\% of the signal's maximal value,
    respectively. Darker lines correspond to higher levels. The
    parameter $\alpha$ grows from left to right. The
    boundary of the set $S_j$ appears as a dashed black line.}
  \label{fig:noNoiseDenseGrid}
\end{figure}

\begin{figure}
  \includegraphics[scale=.35]{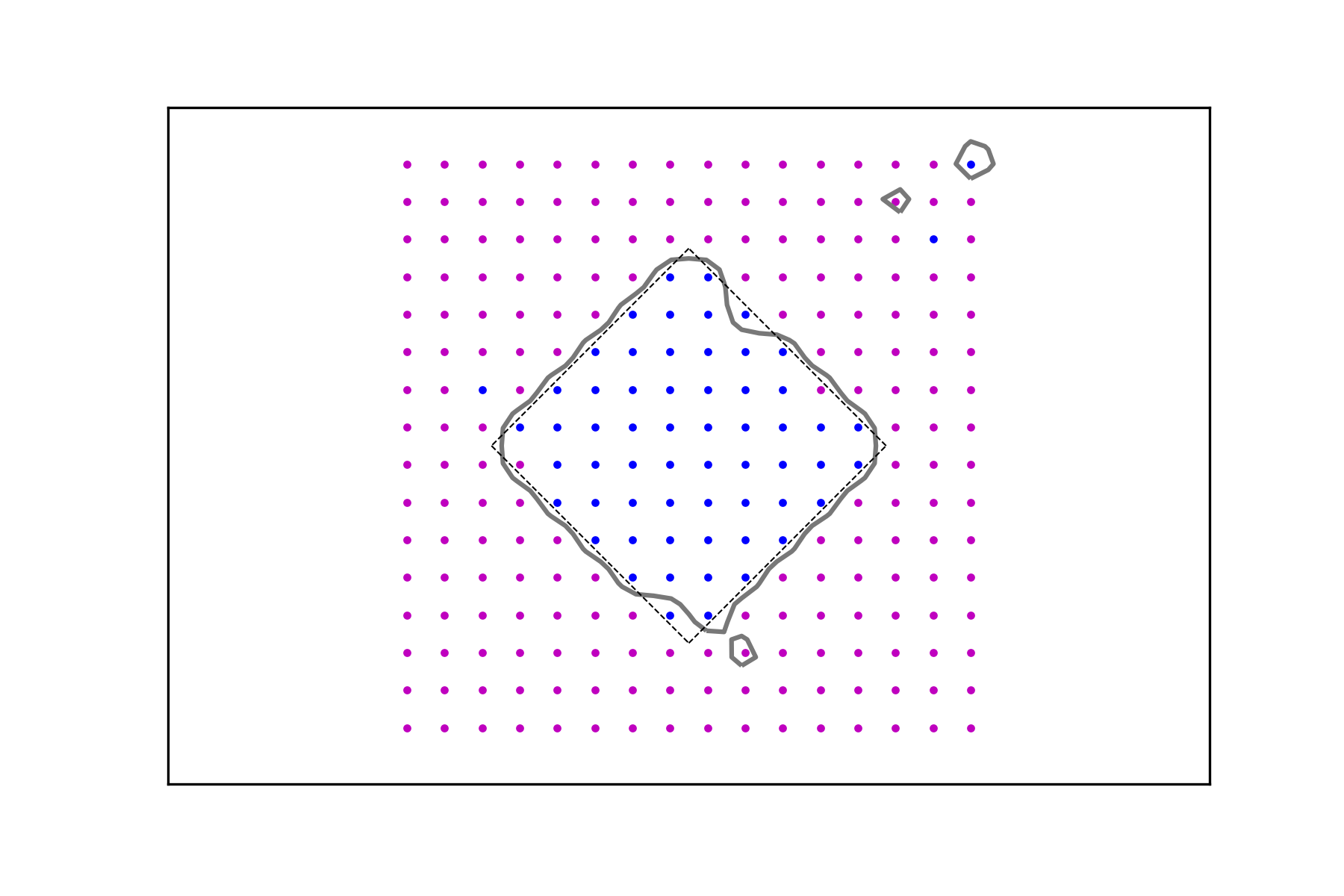}
  \includegraphics[scale=.35]{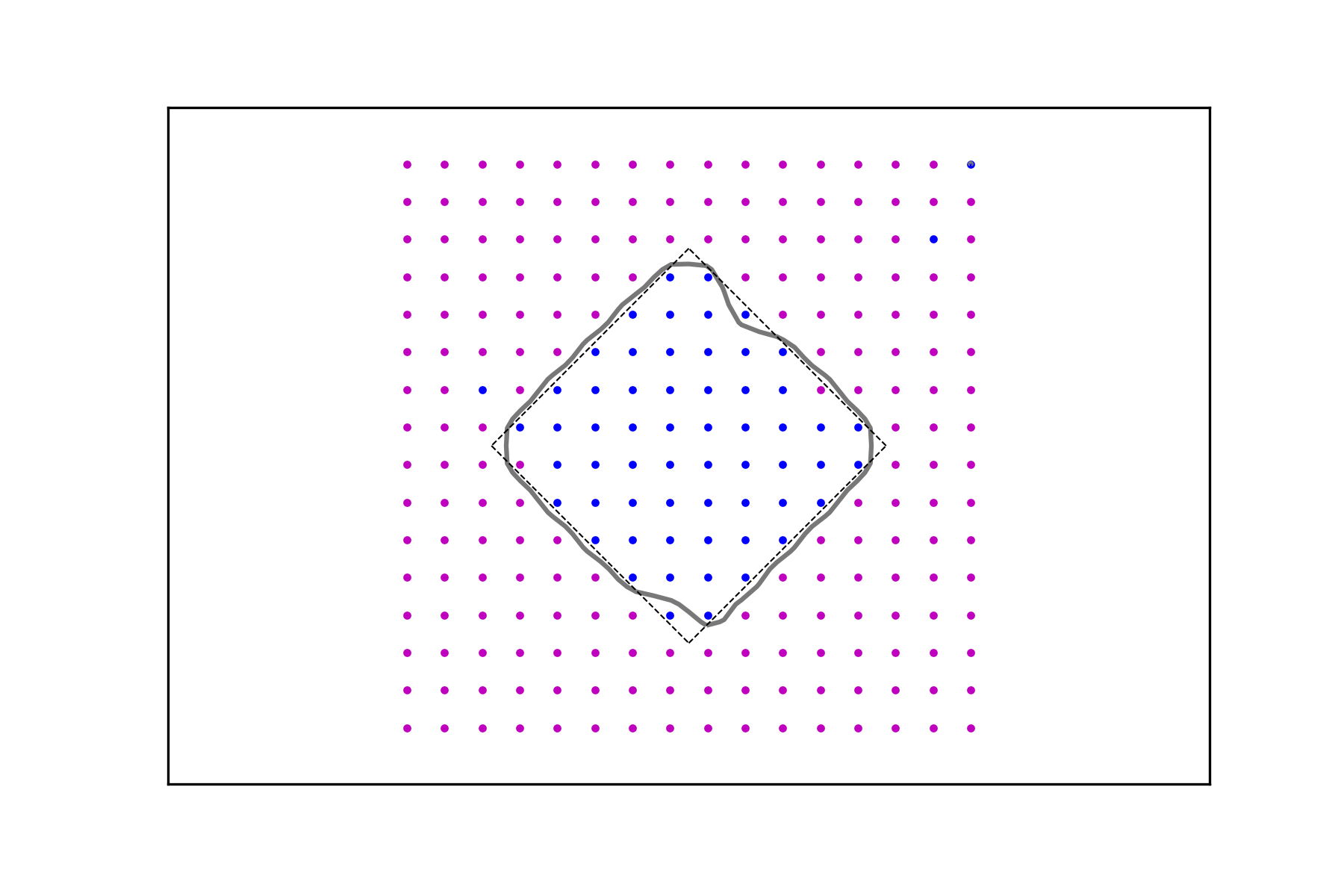}
  \includegraphics[scale=.35]{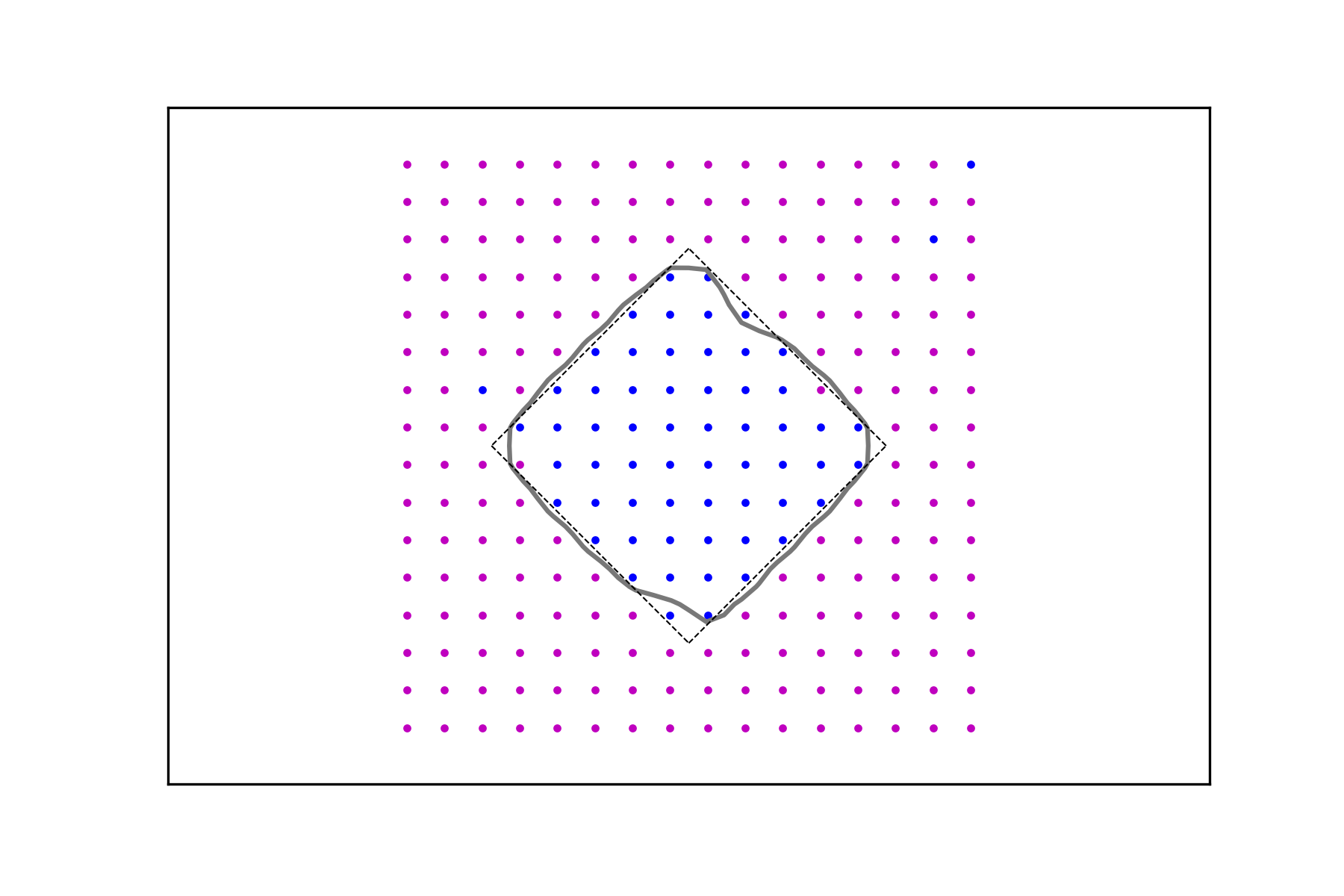}
    \includegraphics[scale=.35]{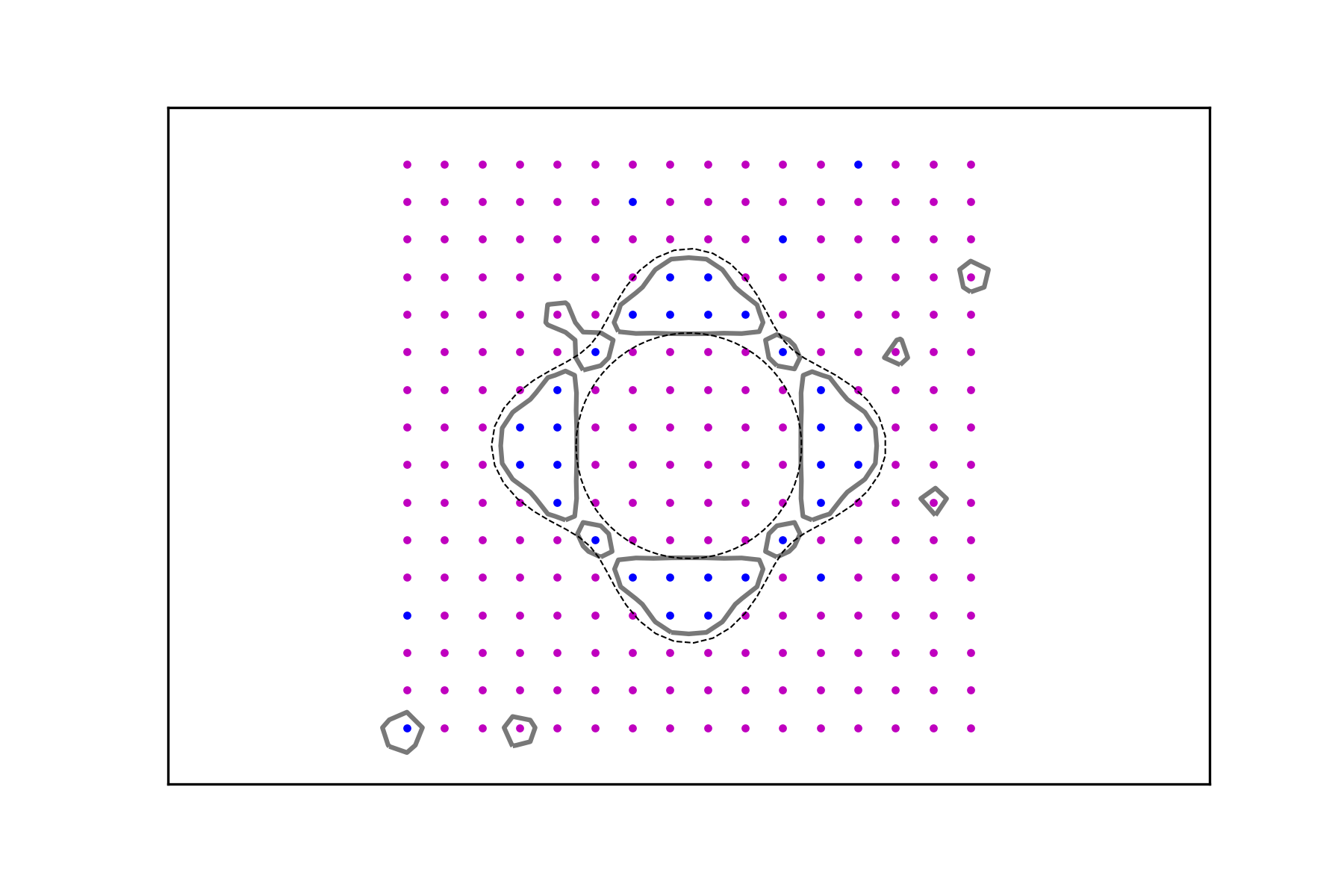}
  \includegraphics[scale=.35]{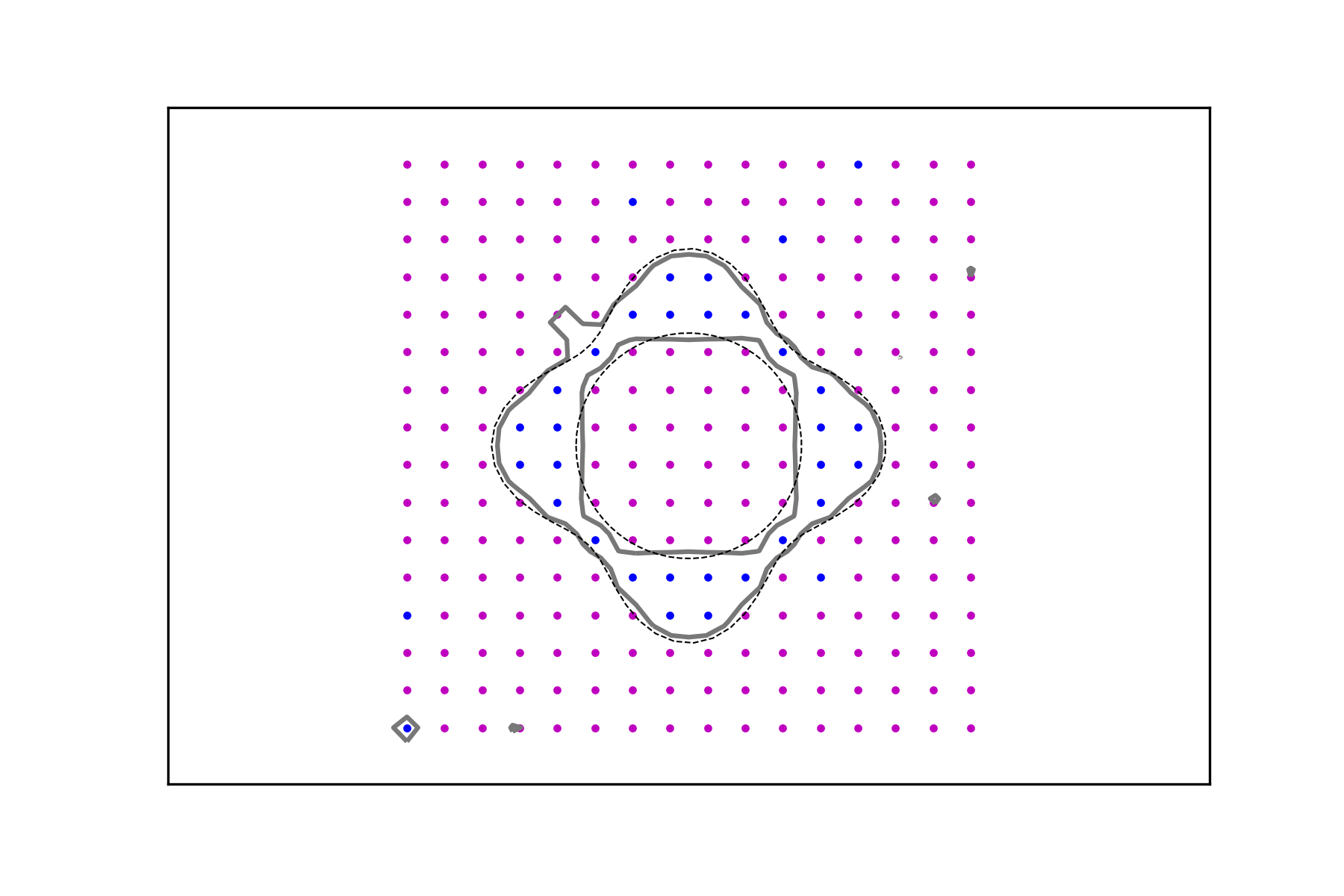}
  \includegraphics[scale=.35]{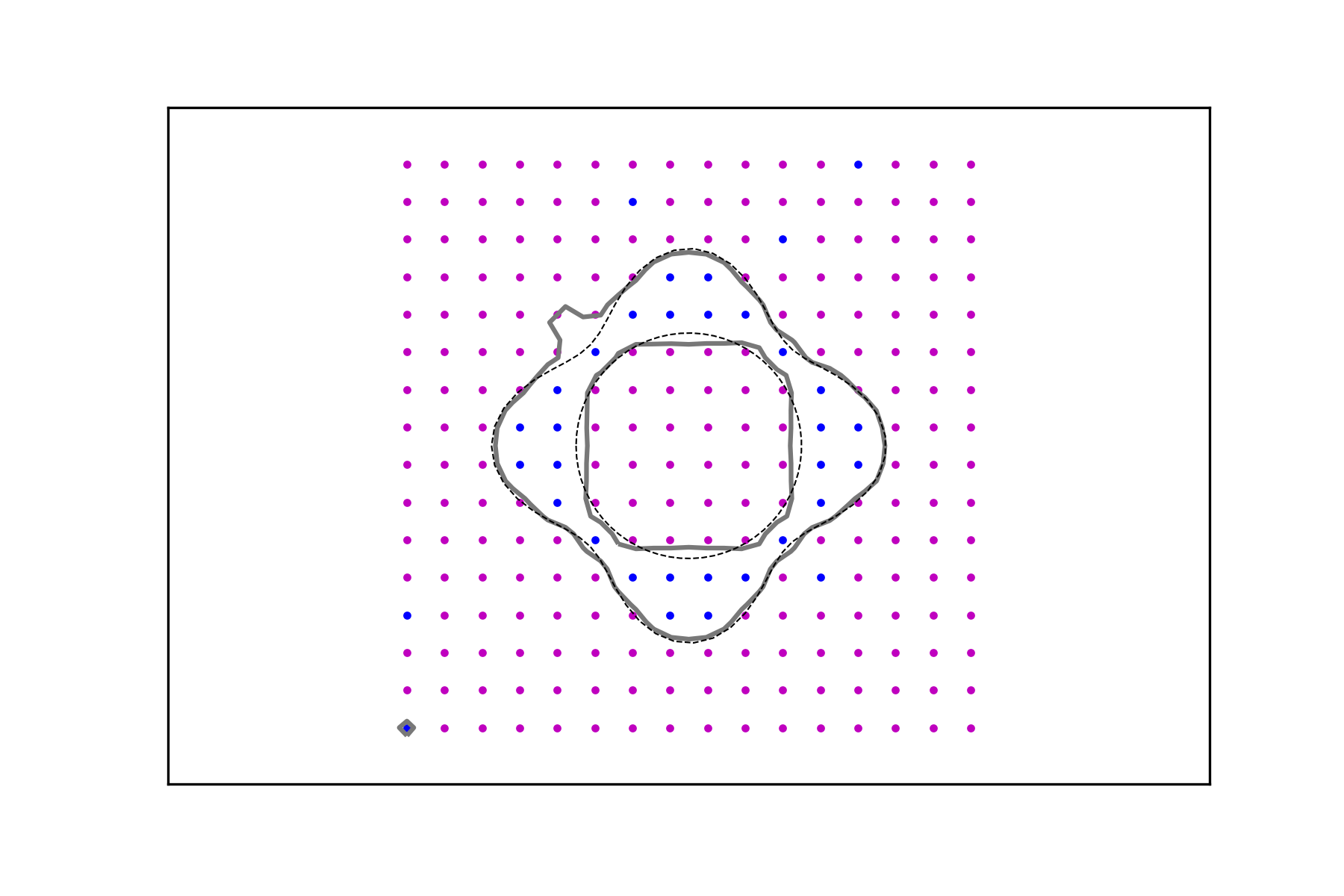}
  \caption{Decision boundary of the signal $u_{\mathbb{D}_{m,j}}$ for
    $m=16$, $j=2,3$, and $a=0,1,1,2$ with 2\% data corruption rate.
    Depicted are the data set (blue dots correspond to the value 1 while magenta dots to
    the value 0) and the level line corresponding to 50\% of the
    signal's maximal value. The parameter $\alpha$ grows from left to
    right. The boundary of the set $S_j$ appears as a dashed black
    line.}
  \label{fig:noise2}
\end{figure}

\begin{figure}
  \includegraphics[scale=.35]{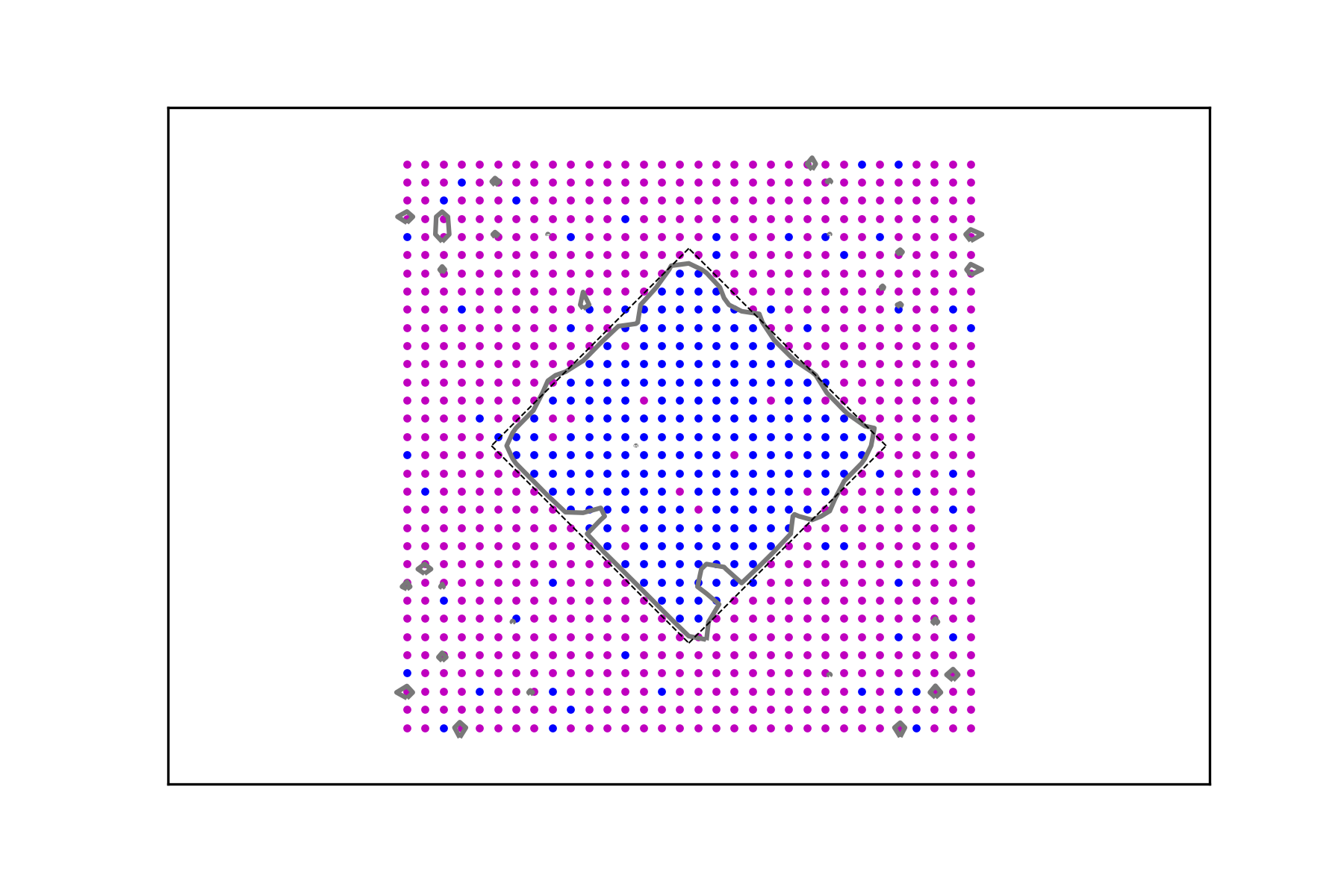}
  \includegraphics[scale=.35]{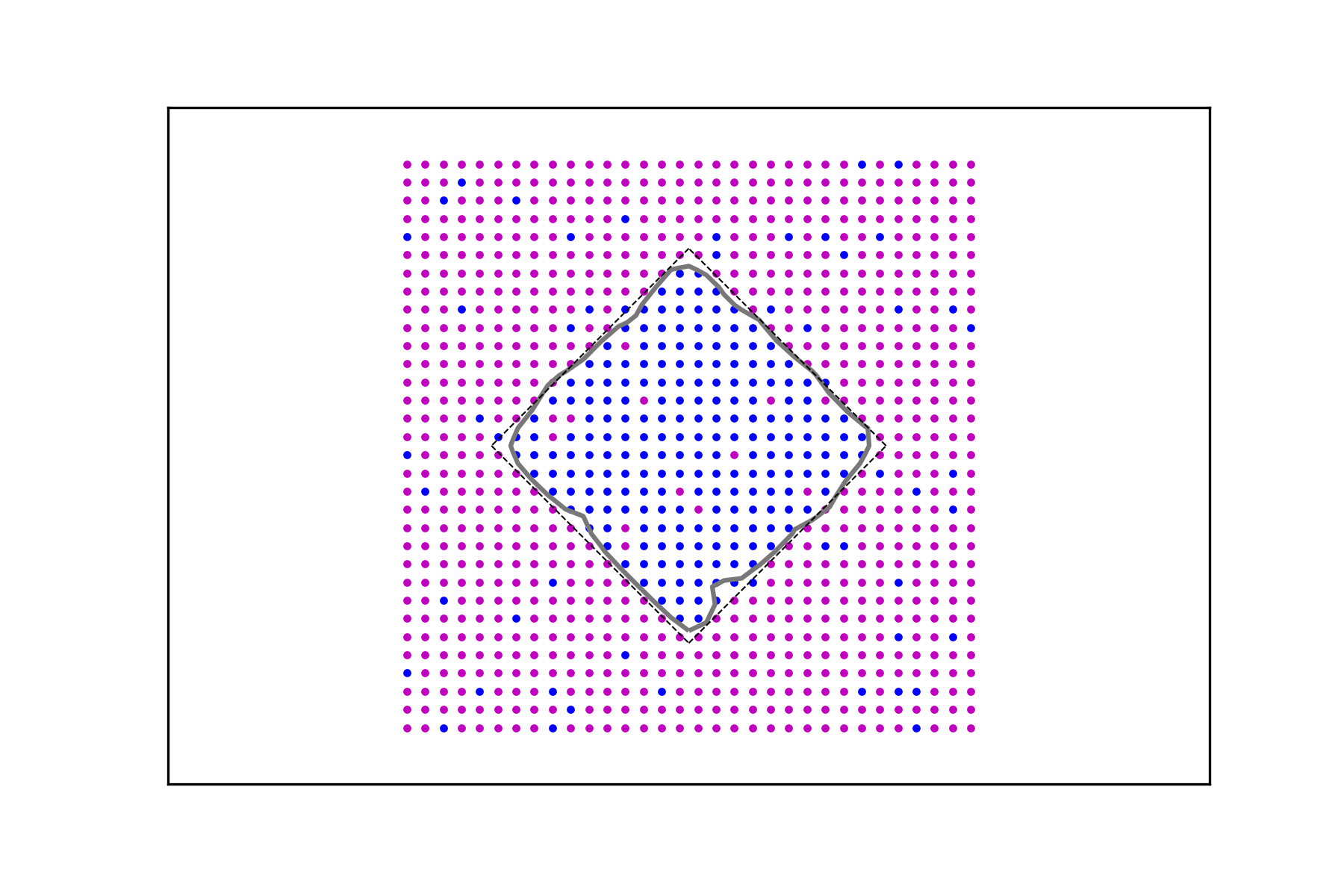}
  \includegraphics[scale=.35]{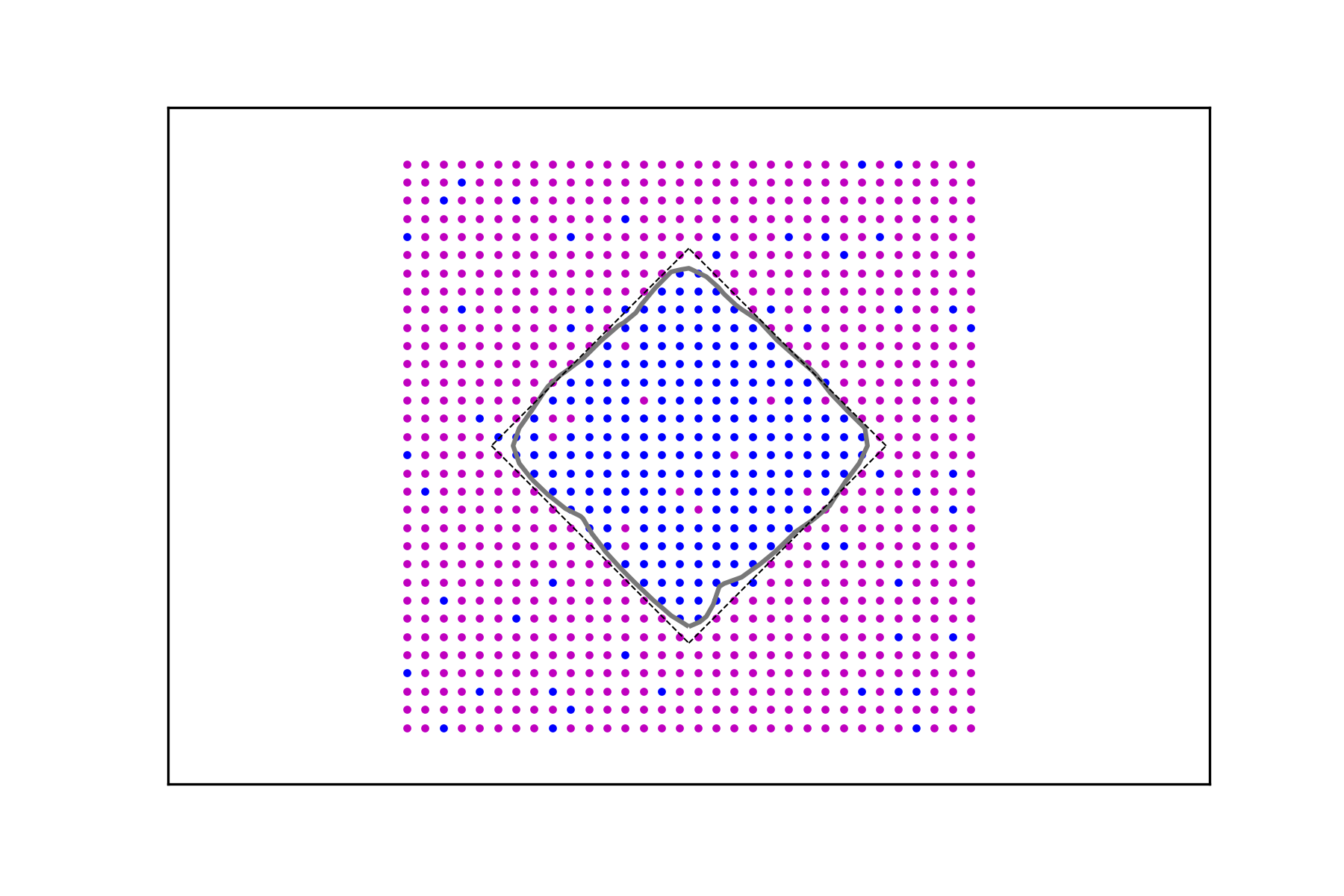}
    \includegraphics[scale=.35]{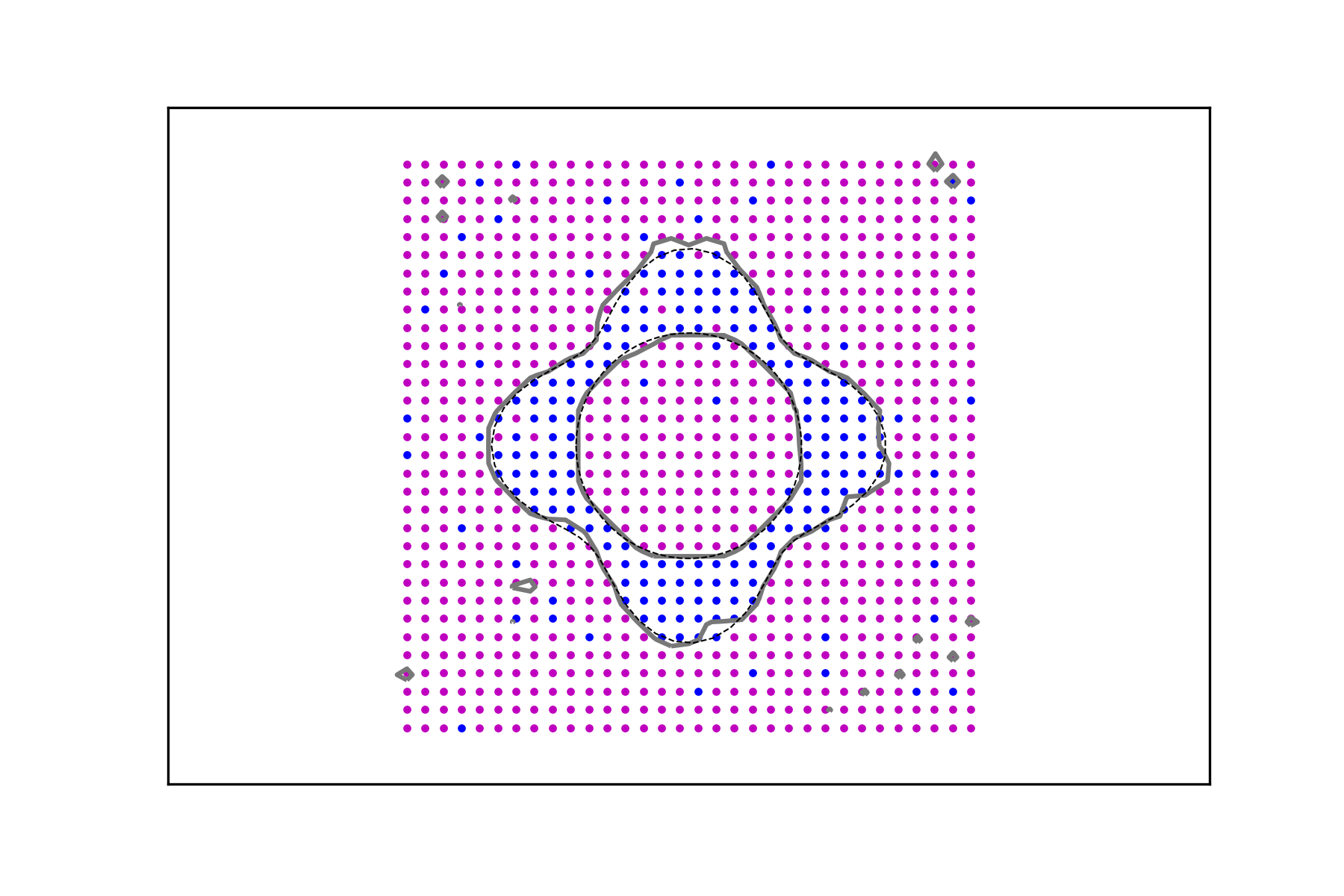}
  \includegraphics[scale=.35]{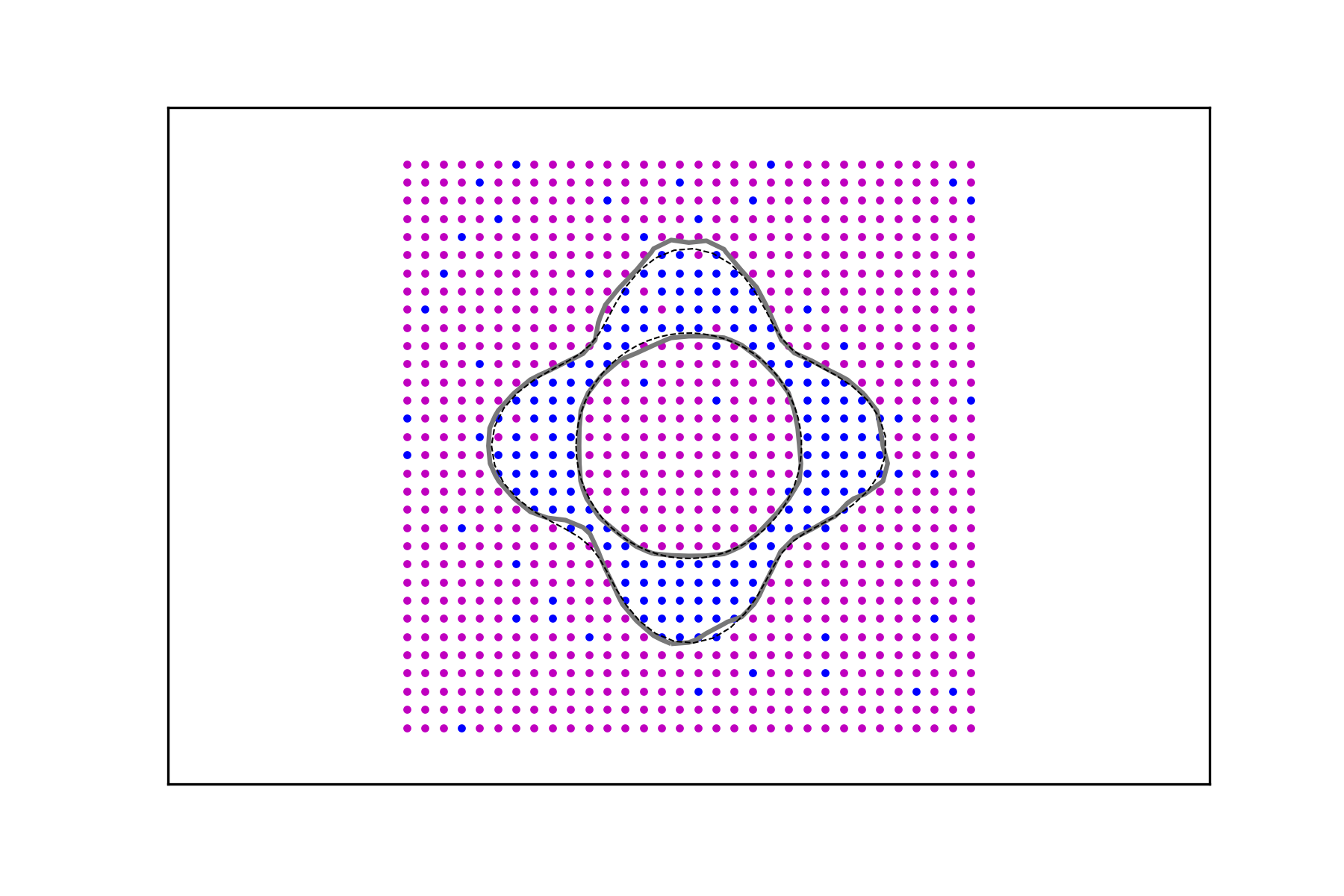}
  \includegraphics[scale=.35]{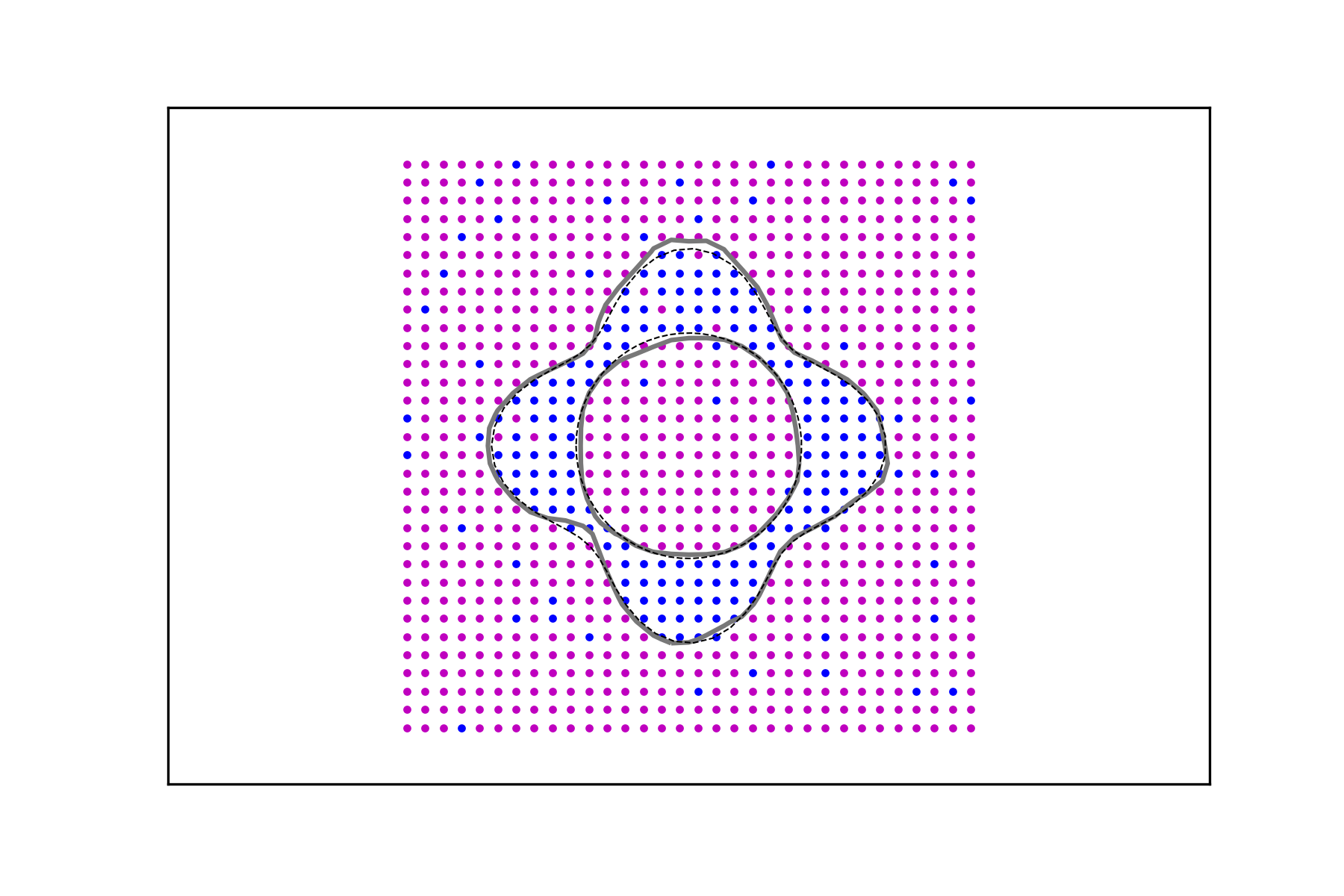}
  \caption{Decision boundary of the signal $u_{\mathbb{D}_{m,j}}$ for
    $m=32$, $j=2,3$, and $a=0,1,1,2$ with 5\% data corruption rate.
    Depicted are the data set (blue dots correspond to the value 1 while magenta dots to
    the value 0) and the level line corresponding to 50\% of the
    signal's maximal value. The parameter $\alpha$ grows from left to
    right. The boundary of the set $S_j$ appears as a dashed black
    line.}
  \label{fig:noise5}
\end{figure}

Take the three subsets of $\mathbb{R}^2$
\begin{align}\label{s1}
  S_1&=\big\{ x\in \mathbb{R}^2\,\big |\, |x|\leq .6\big\},\\
  S_2&=\big\{ x\in \mathbb{R}^2\,\big |\, |x_1|+|x_2|\leq .7\big\},\\
  S_3&=\big\{ r(\theta)\bigl( \cos(\theta),\sin(\theta)\bigr) \in
       \mathbb{R}^2\,\big |\, .4\leq r(\theta)\leq
       .6+.1\cos(4\theta),\: \theta\in[0,2\pi)\big\},
\end{align}
and consider the associated characteristic function $\chi_{S_j}$ for
$j=1,2,3$. The first data set consists of the values of these
functions on a regular grid, that is,
$$
\mathbb{D}_{m,j}=\big\{ \bigl(x^i,\chi_{S_j}(x^i)\bigr)\,\big |\,
i=1,\dots,m^2\big\}, \: j=1,2,3,
$$
for $\mathbb{X}_m=\big\{ (kh-1,lh-1)\,\big |\, 0\leq k,l\leq m-1\big\}$ and
$h=2/(m-1)$, which amounts to a uniform discretization of the box
$B=[-1,1]\times [-1,1]$. In Figure \ref{fig:noNoise} some level
lines of the interpolant $u_{\mathbb{D}_{m,j}}$ are shown for the three
functions $\chi_{S_j}$, $j=1,2,3$,  and for different values of the
regularizing parameter $\alpha>0$ and $m=16$. While the size of
the data set clearly correlates with the ``accuracy'' of the
interpolation, the approximating function does an excellent job at
generating meaningful and smooth level sets. Their smoothness is
affected mainly by the parameter $\alpha$ and the their proximity to
the level sets corresponding to the highest values.
It follows that if a characteristic function has to be recovered or
inferred from a data set, thresholding based on the interpolant
$u_{\mathbb{D}_{m,j}}$ is an effective strategy and the decision boundary
$\bigl[u_{\mathbb{D}_{m,j}}=.5\max(u_{\mathbb{D}_{m,j}})\bigr]$ is a
solid choice across a range of values of the regularization parameter.
Figure \ref{fig:noNoiseDenseGrid} depicts the same experiments using the
denser data sets $\mathbb{D}_{32,j}$ for $j=1,2,3$.
In Figures \ref{fig:noise2} and \ref{fig:noise5}, it is shown how the
method performs in the presence of data corruption. In Figure \ref{fig:noise2}
2\%  of the data is misclassified, whereas the misclassification rate
in Figure \ref{fig:noise5} is 5\%. By this we mean that a mistake is
made, with the given probability, when a value is assigned to an
argument by evaluating the corresponding characteristic
function. These examples clearly demonstrate the usefulness of the
regularizing parameter which leads to data signals whose decision
level sets are more stable in the presence of classification errors.
\subsection{Sample Data}
\begin{figure}
  \includegraphics[scale=.35]{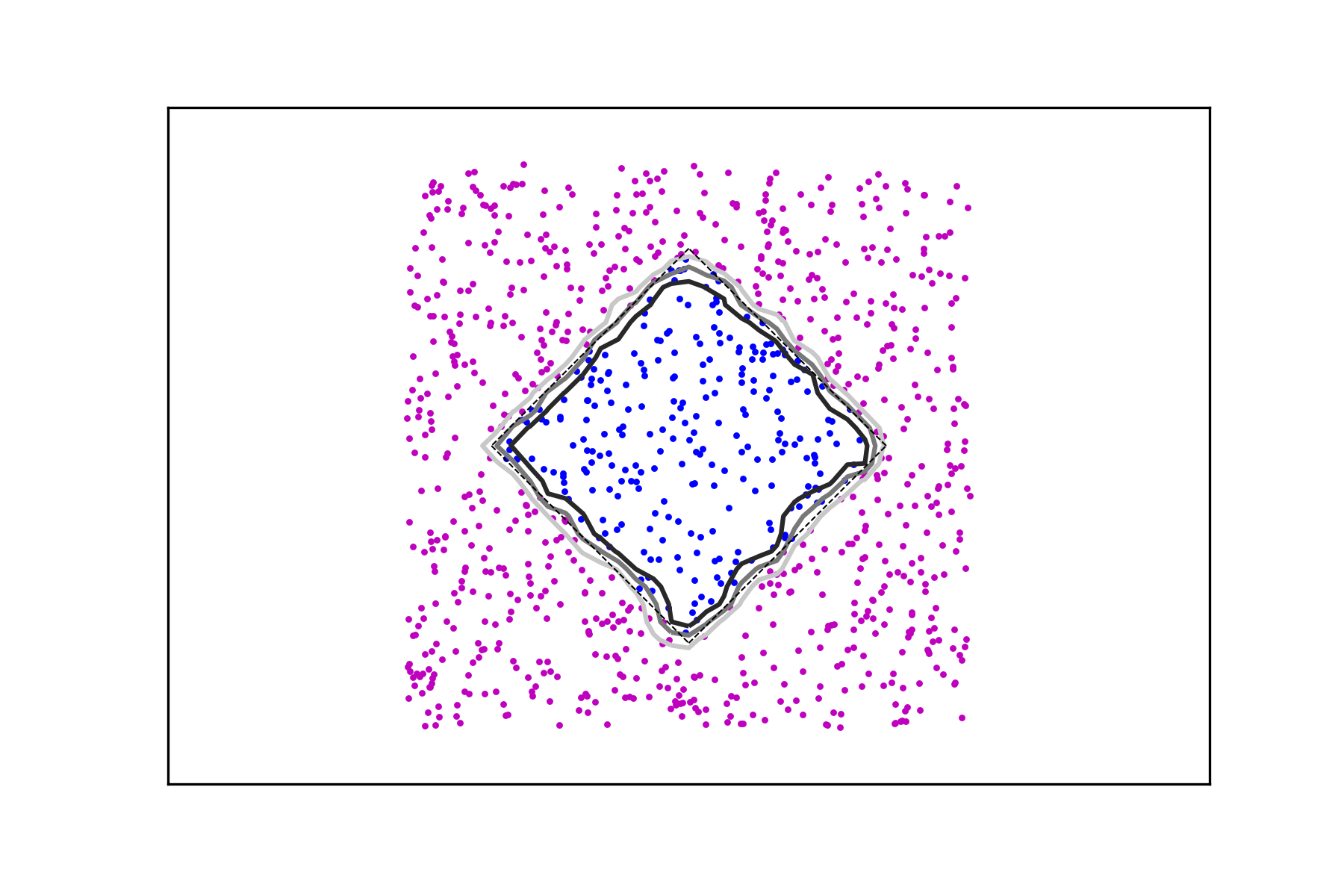}
  \includegraphics[scale=.35]{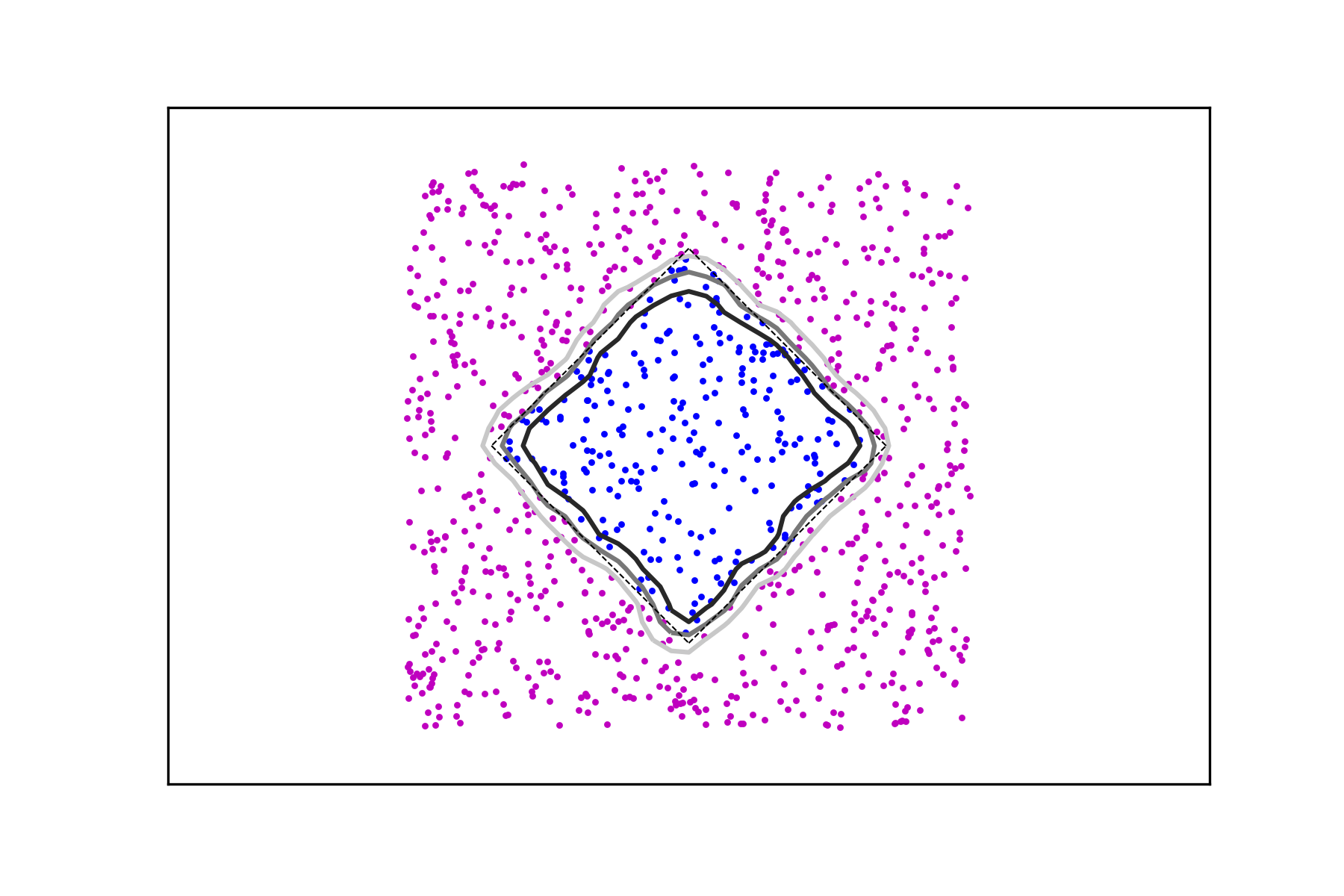}
  \includegraphics[scale=.35]{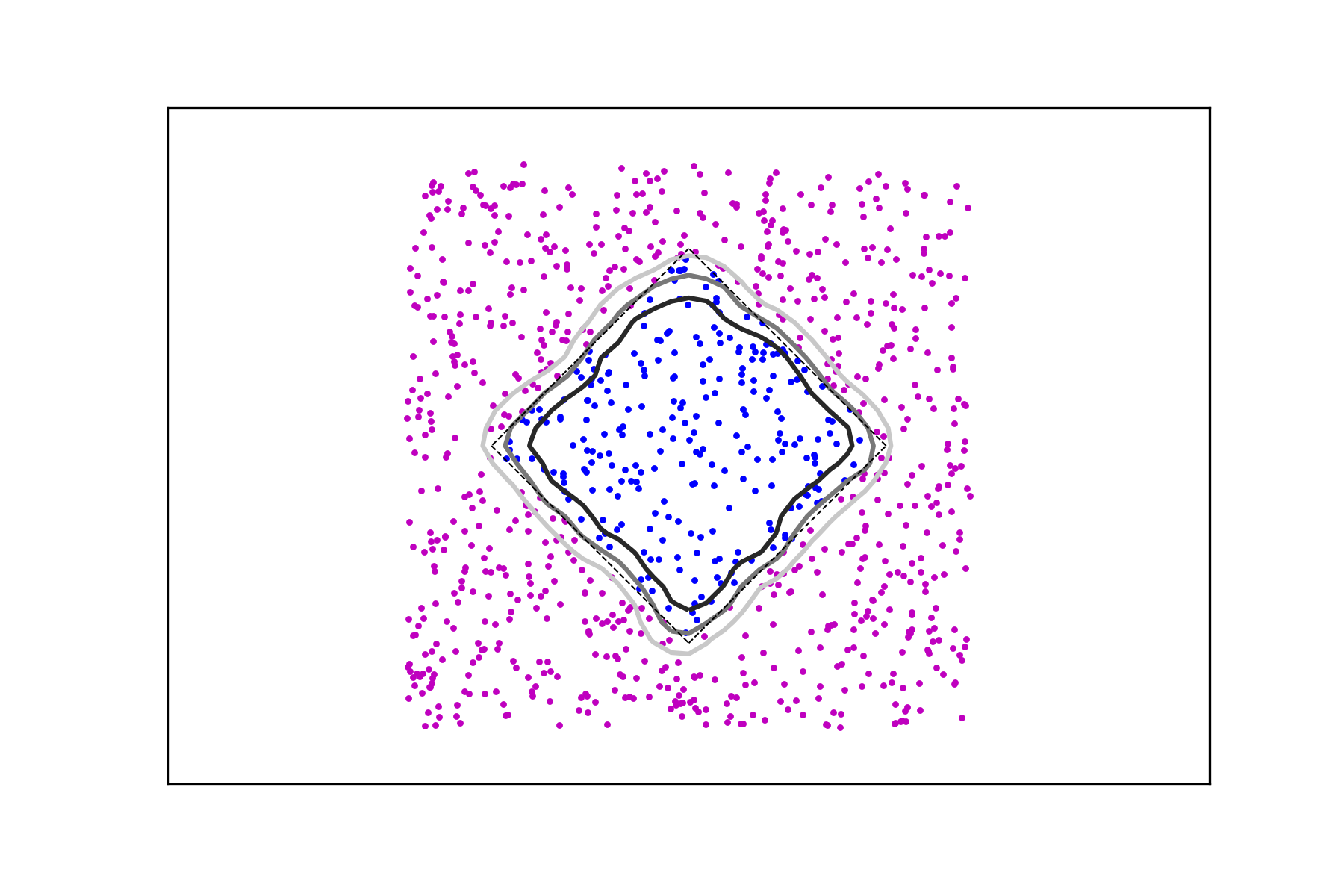}
    \includegraphics[scale=.35]{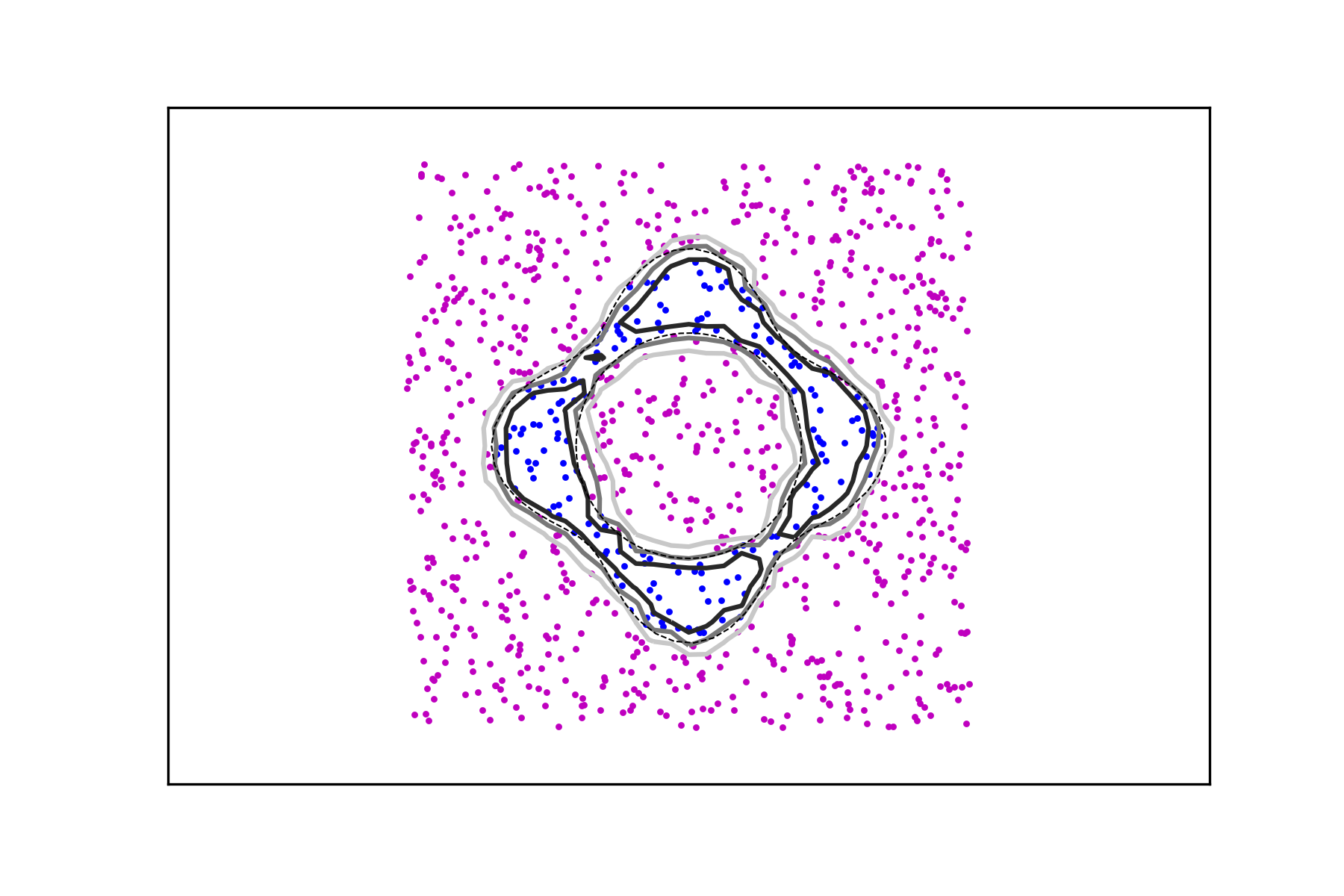}
  \includegraphics[scale=.35]{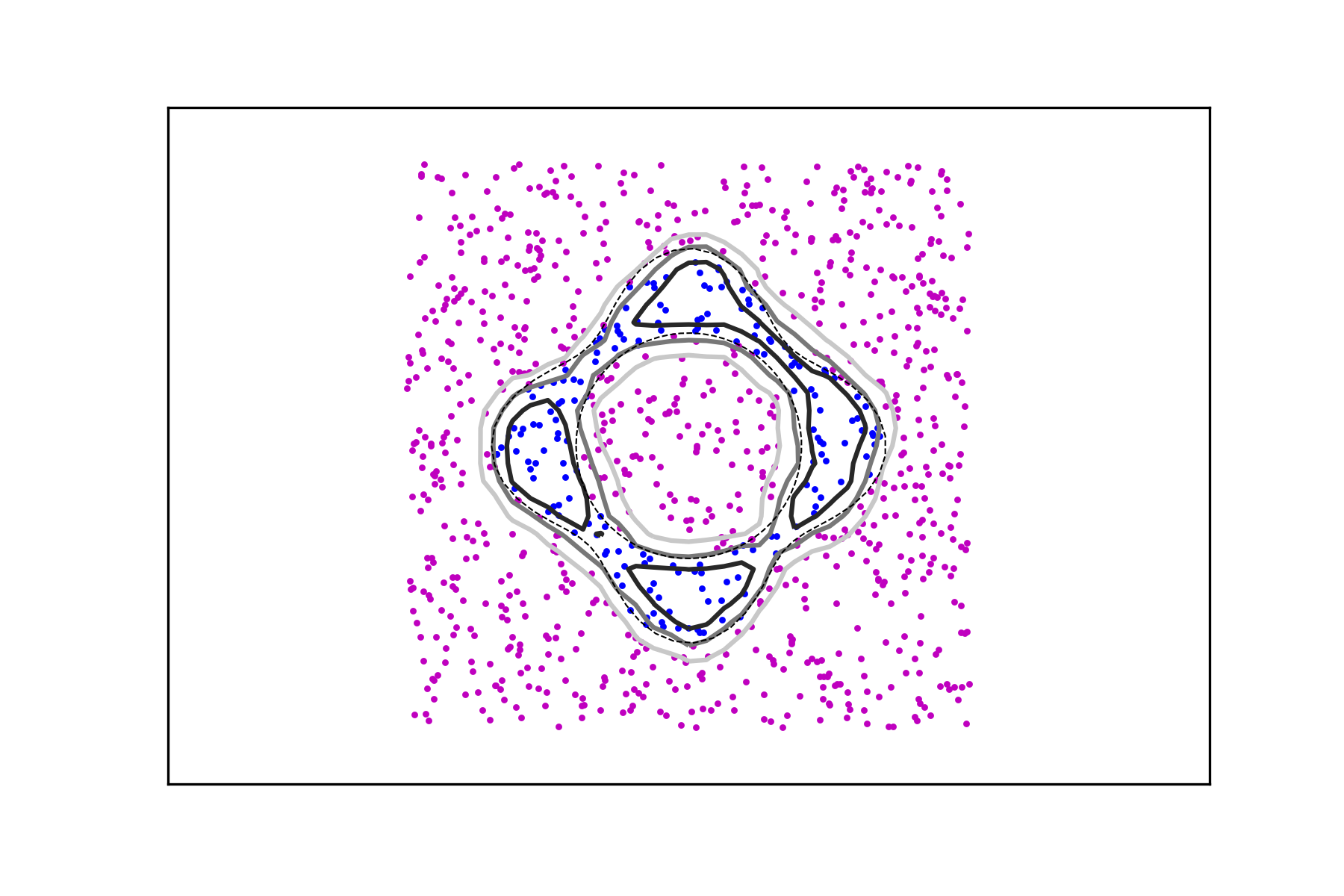}
  \includegraphics[scale=.35]{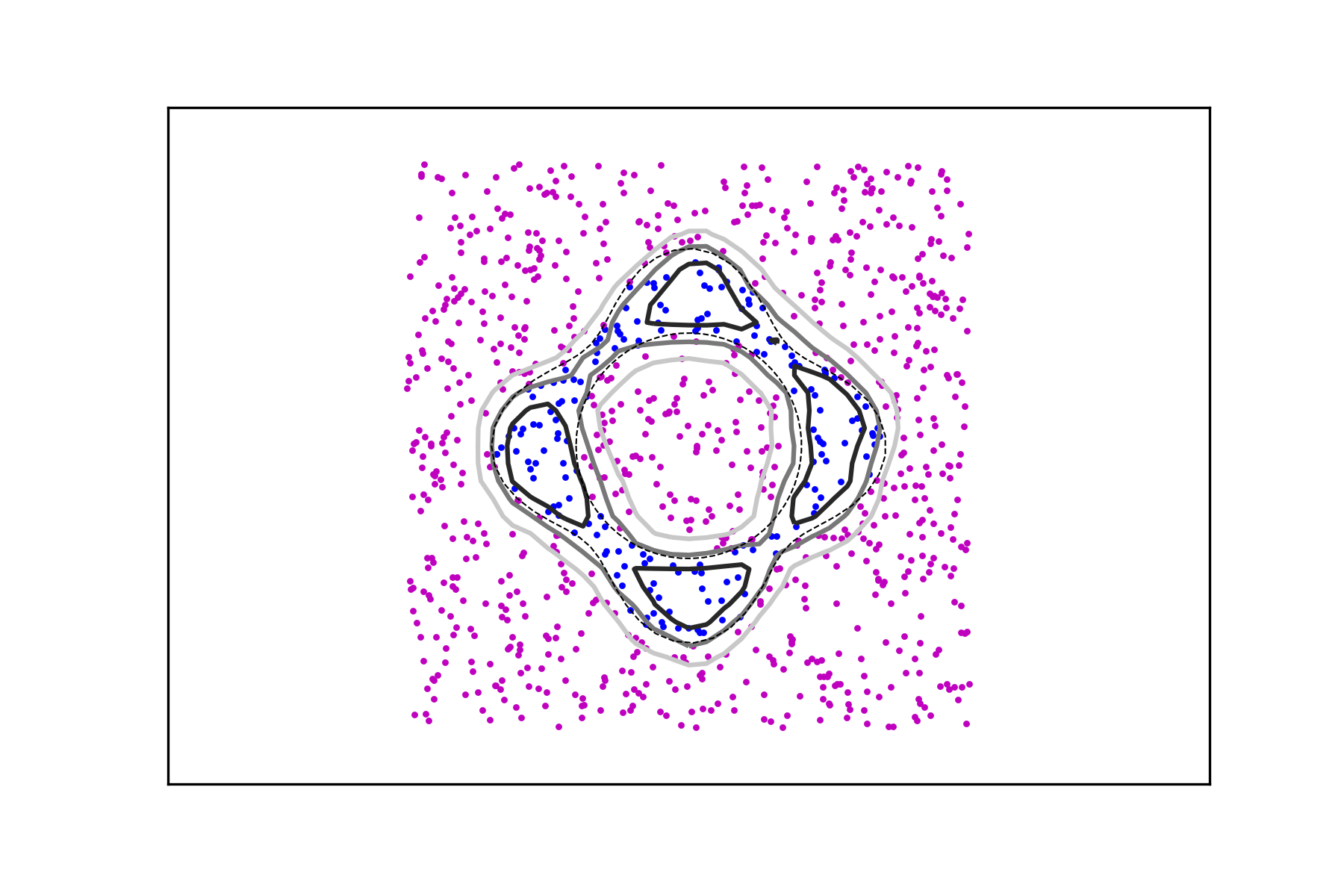}
  \caption{Level lines of 20\%, 50\%, and 80\%, increasingly dark, of
    the signal's maximal 
    value for regularizations parameter $\alpha=.1,1,2.$, from left to
  right. The number of randomly sampled points, also depicted with the
  associated value color-coded (with blue representing the value 1, while
  magenta the value 0), is $1024$ as in the denser regular
  grids of previous examples.}
  \label{fig:sampled}
\end{figure}

Finally we demonstrate that the method offers a degree robustness 
when the data arguments are randomly sampled from a uniform
distribution supported on the box $B$. The resulting decision
boundaries of half maximal value are depicted in Figure
\ref{fig:sampled} along with the sampled argument data sets
$\mathbb{X}_m$. The sampling rate clearly affects the smoothness of
the level sets, a deterioration that is to some extent counteracted by
the regularization.
\subsection{Classification}
Continuity of the interpolant and its level sets (almost all of them
actually smooth) are obtained at the 
cost of approximate interpolation. Such an approximation can still be
accurate when the argument data set covers the function's domain of
definition uniformly and the value set is accurate, but the real
advantage of this method is its applicability to incomplete data and/or
noisy data sets. This point is further reinforced with the next series
of experiments, where the data build a lower dimensional manifold of
the ambient space and its signal is weak, that is, information about
the underlying function is limited to sets of zero measure, or the
data is not deterministic (in its argument set) but only has a
probability distribution for its location.

\begin{figure}
  \includegraphics[scale=.4]{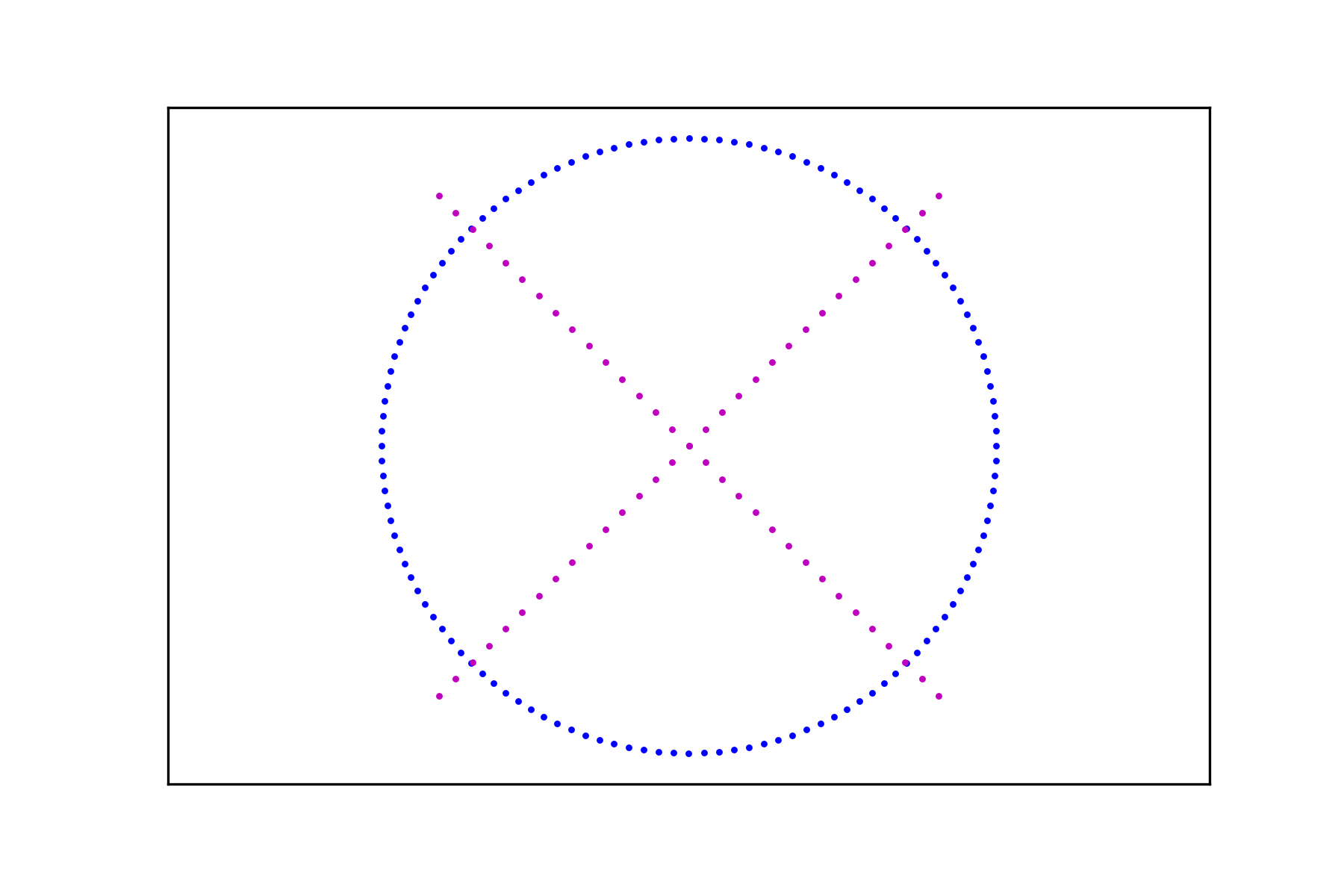}
  \includegraphics[scale=.4]{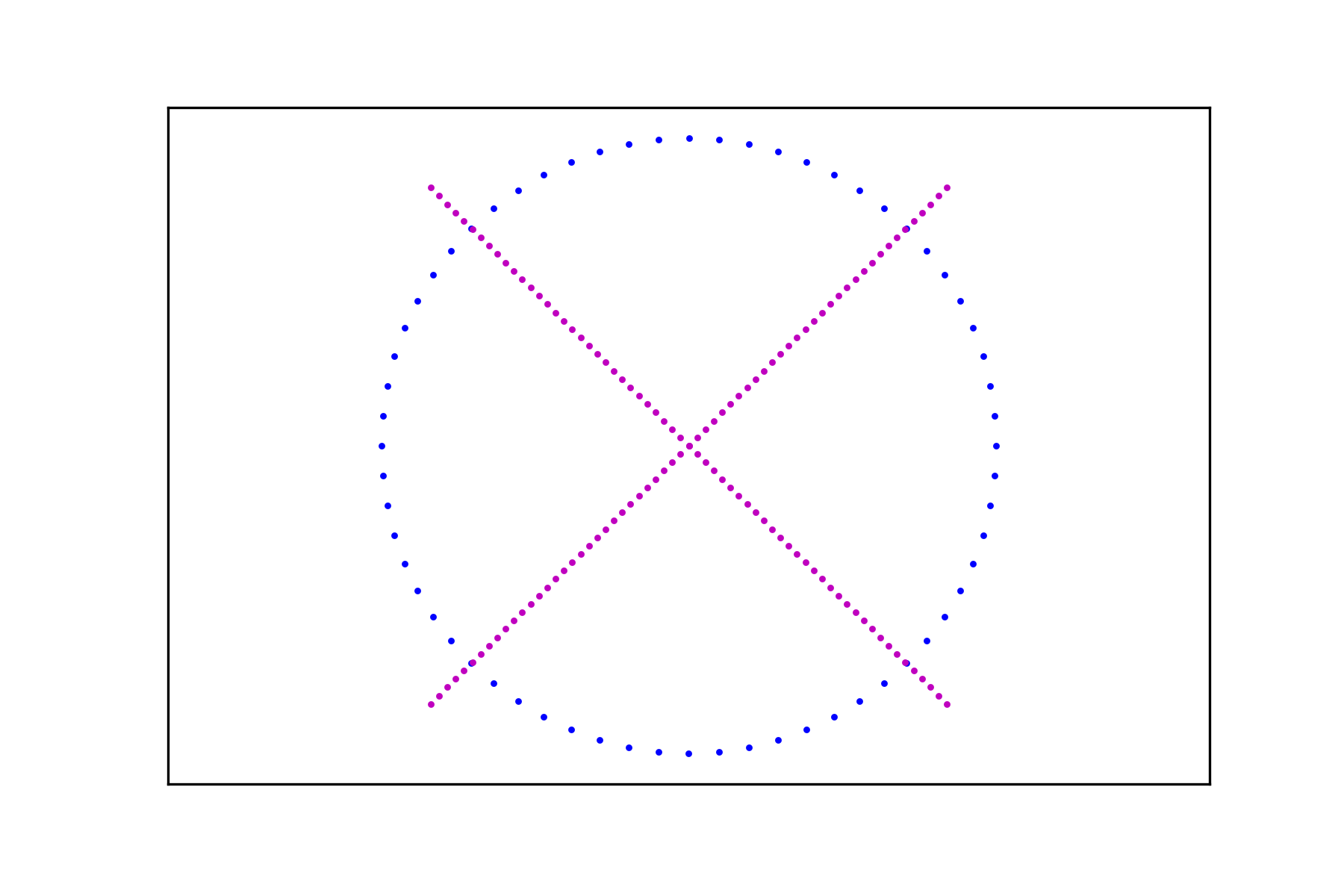}
  \caption{From left to right, the data set pairs $(C_k,S_k)$,
    $k=1,2$. They can be thought as different samplings of the same
    pair of ``continuous'' sets.}
  \label{fig:cCdata}
\end{figure}

\begin{figure}
  \includegraphics[scale=.35]{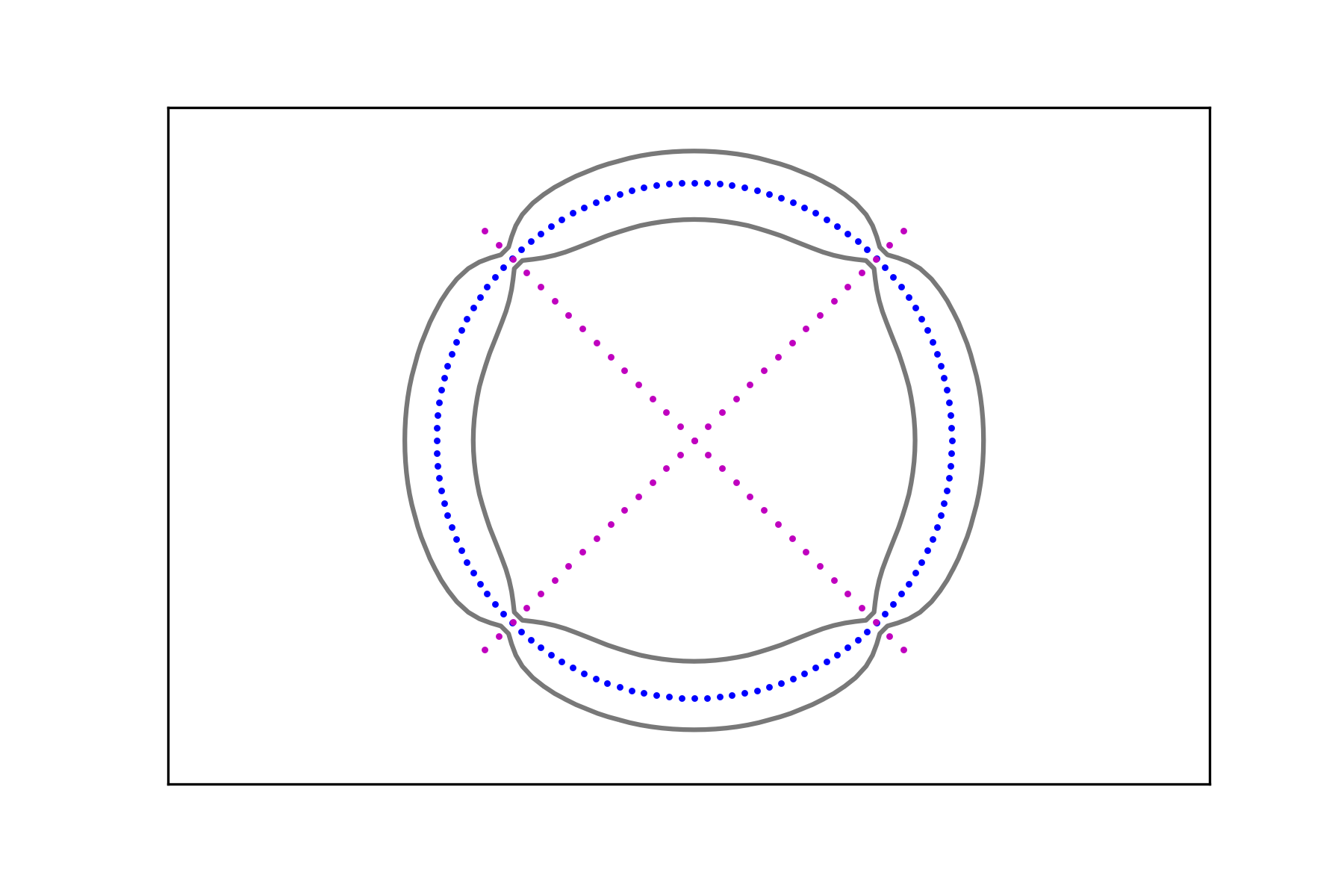}
  \includegraphics[scale=.35]{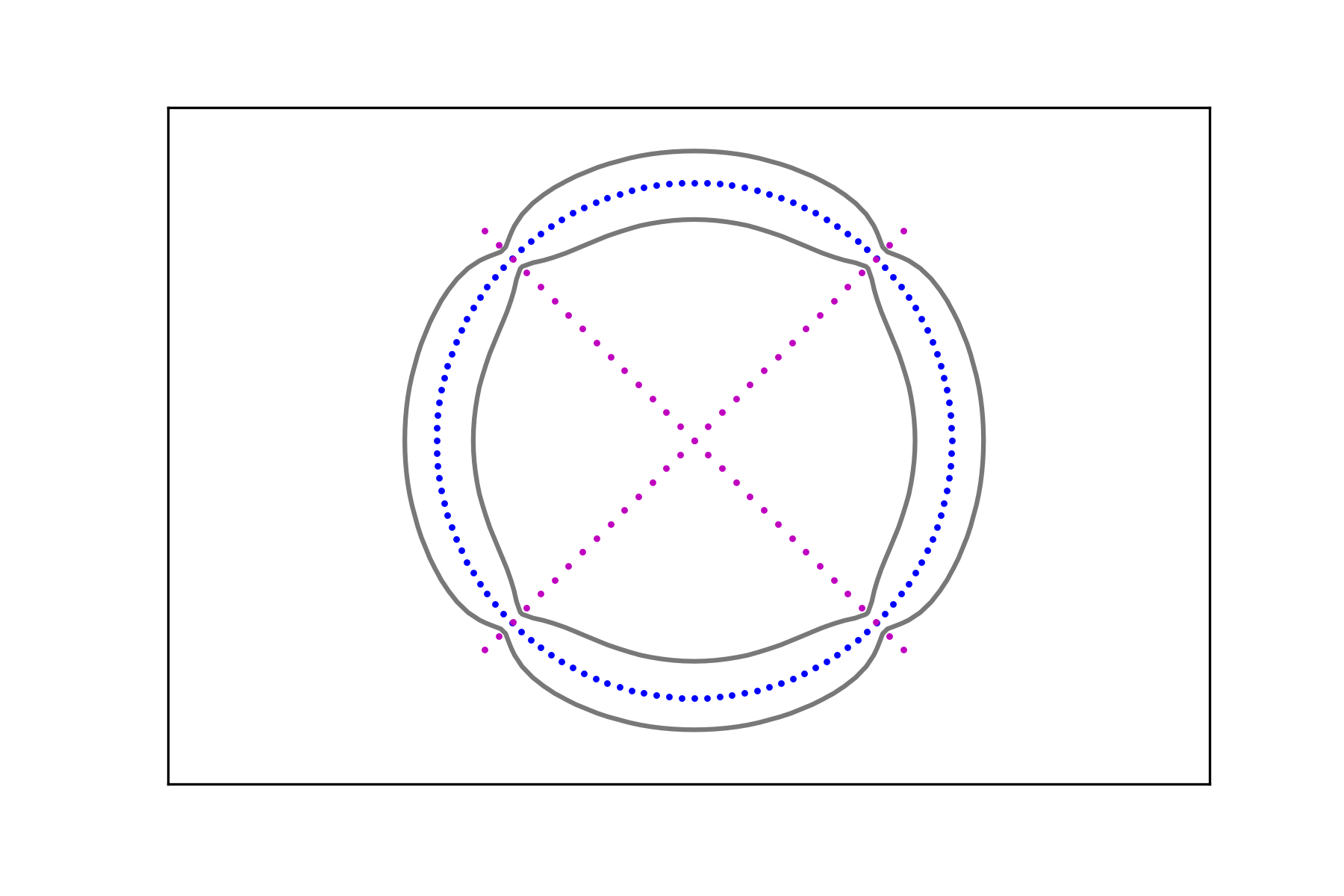}
  \includegraphics[scale=.35]{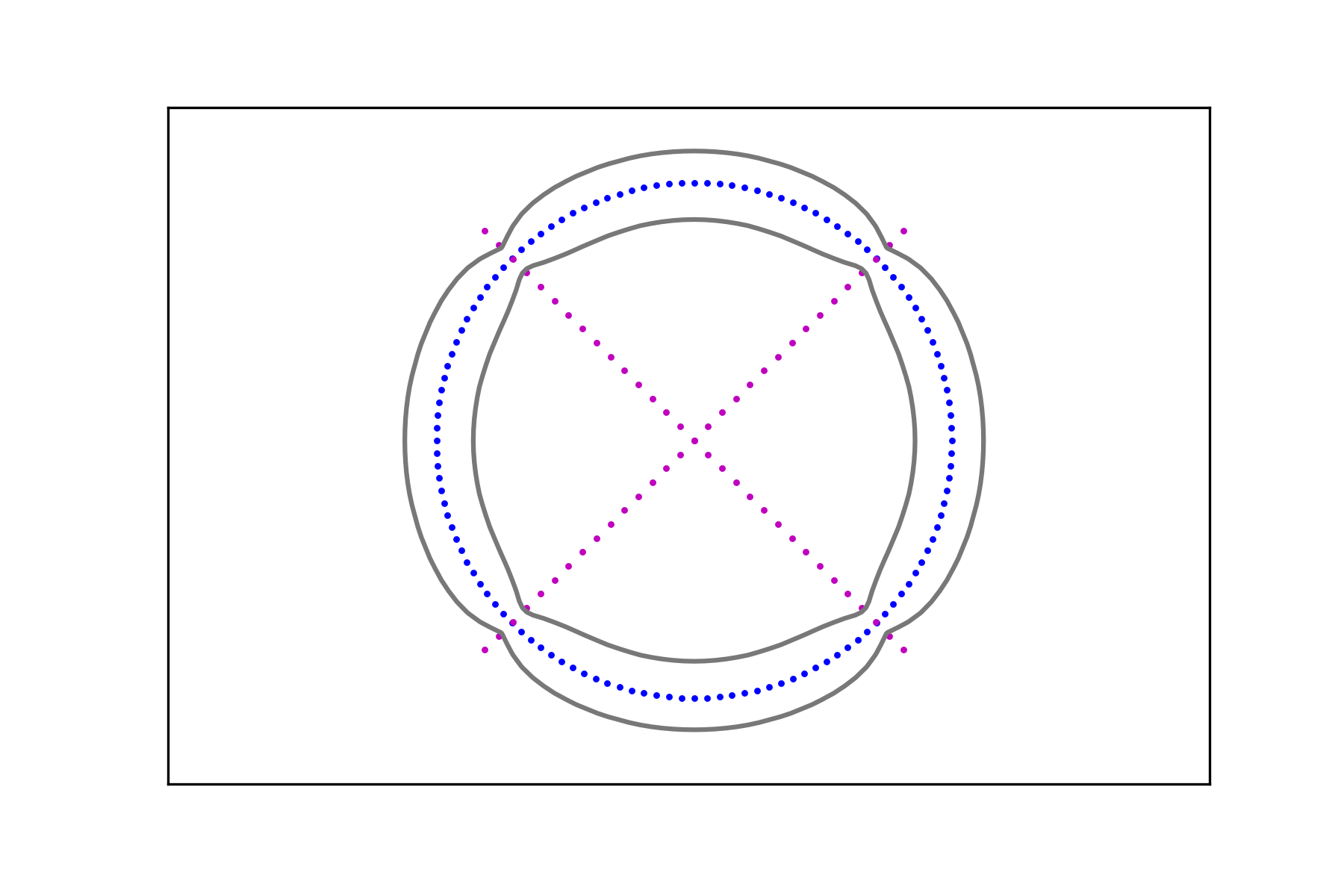}
  \includegraphics[scale=.35]{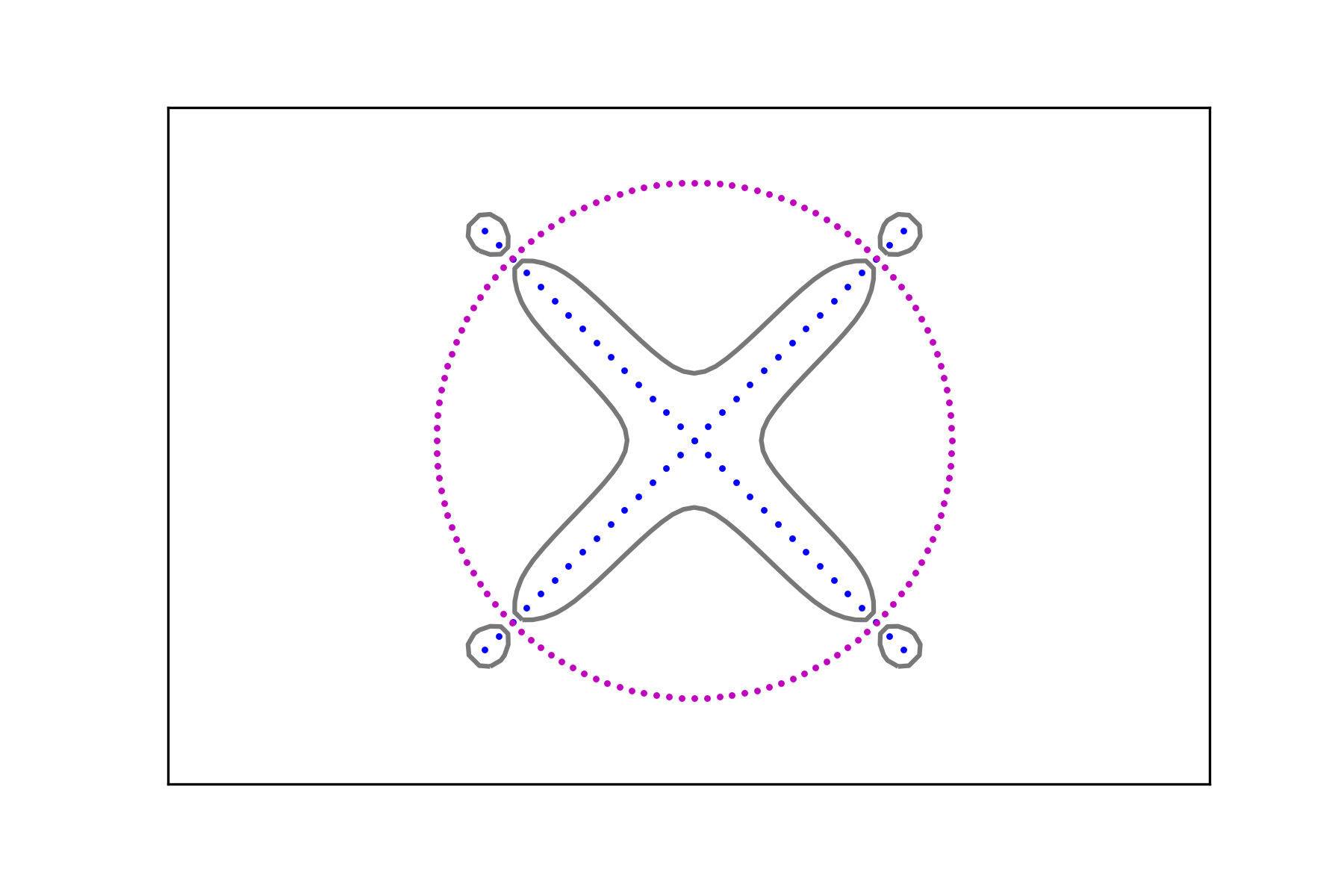}
  \includegraphics[scale=.35]{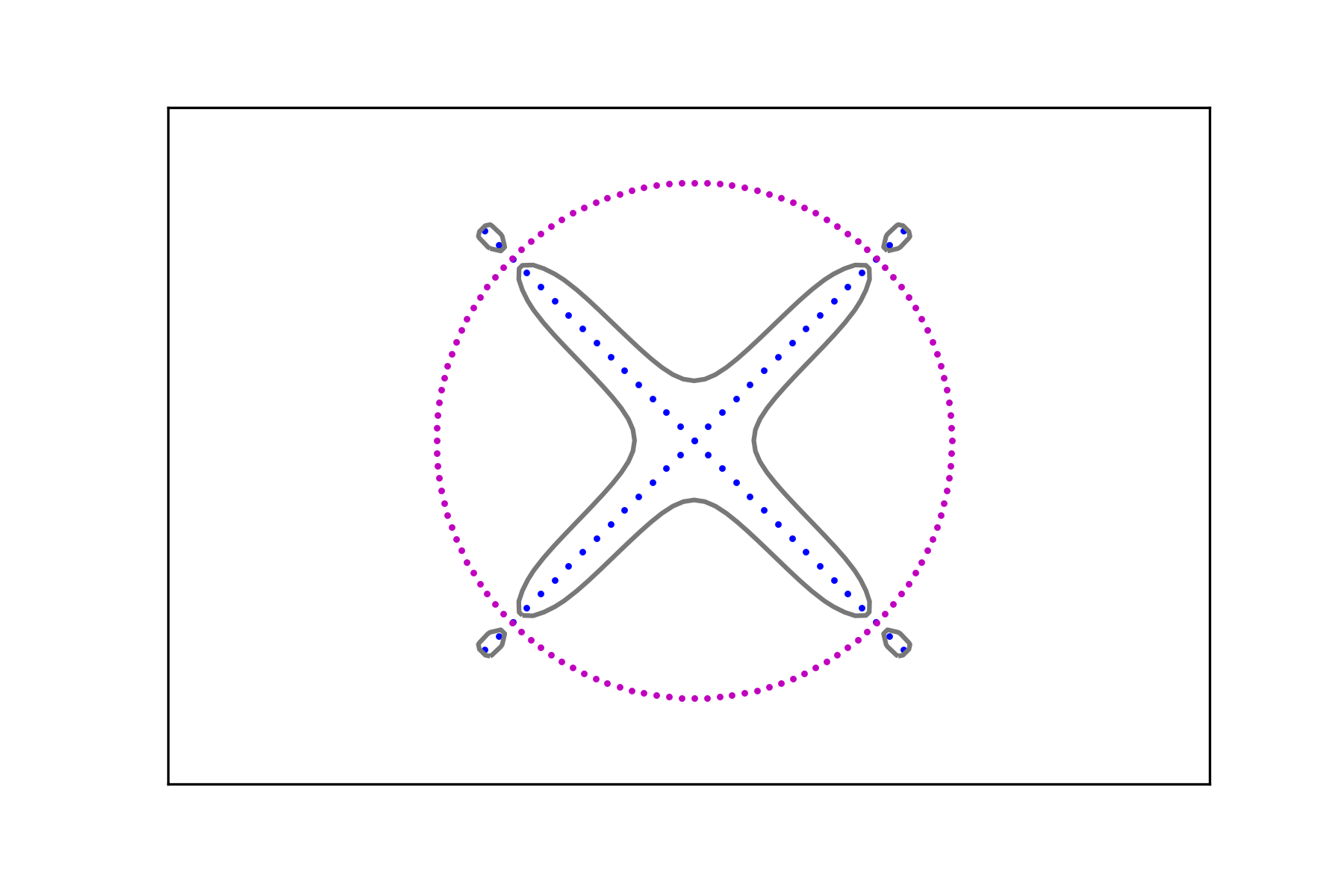}
  \includegraphics[scale=.35]{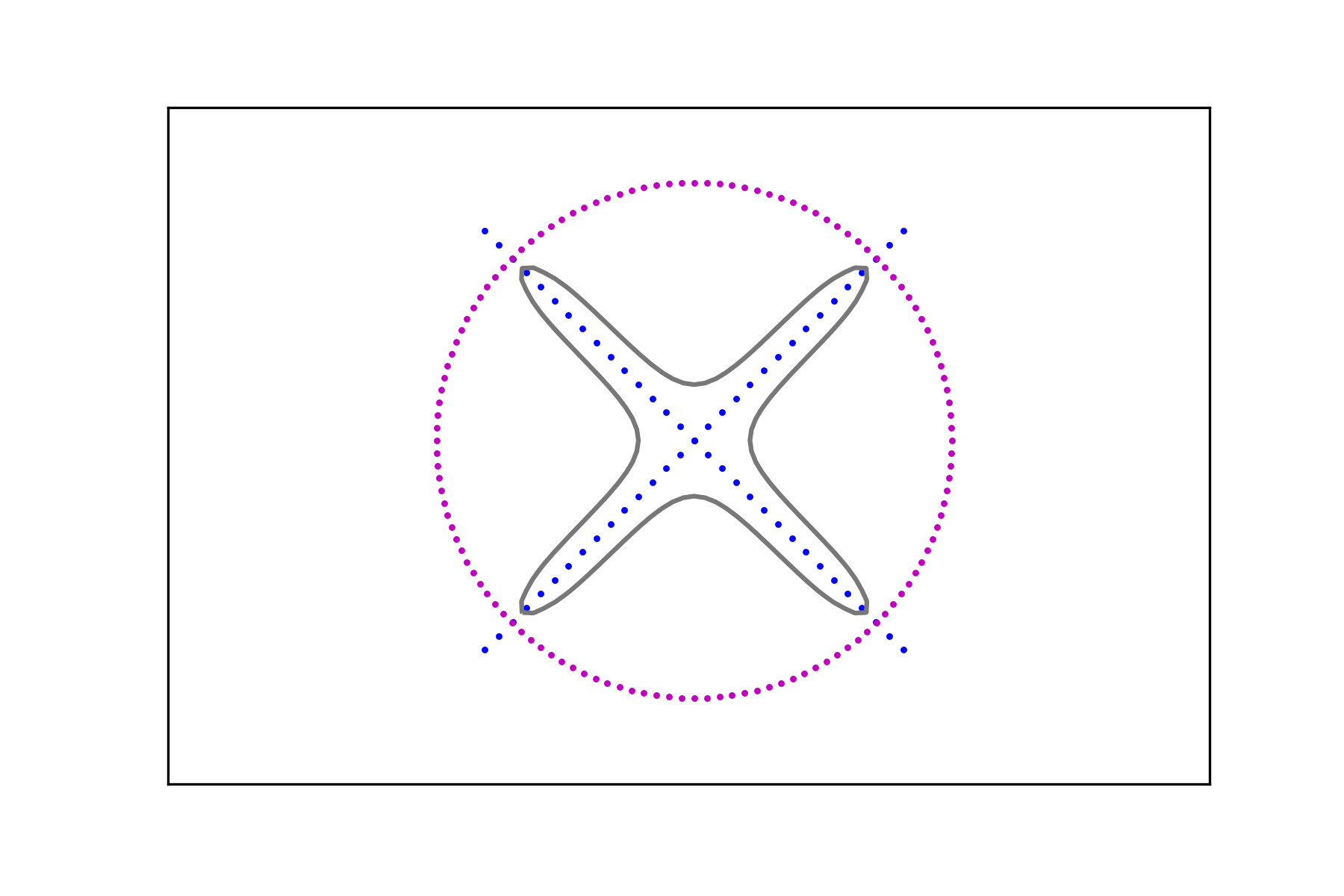}
   \includegraphics[scale=.35]{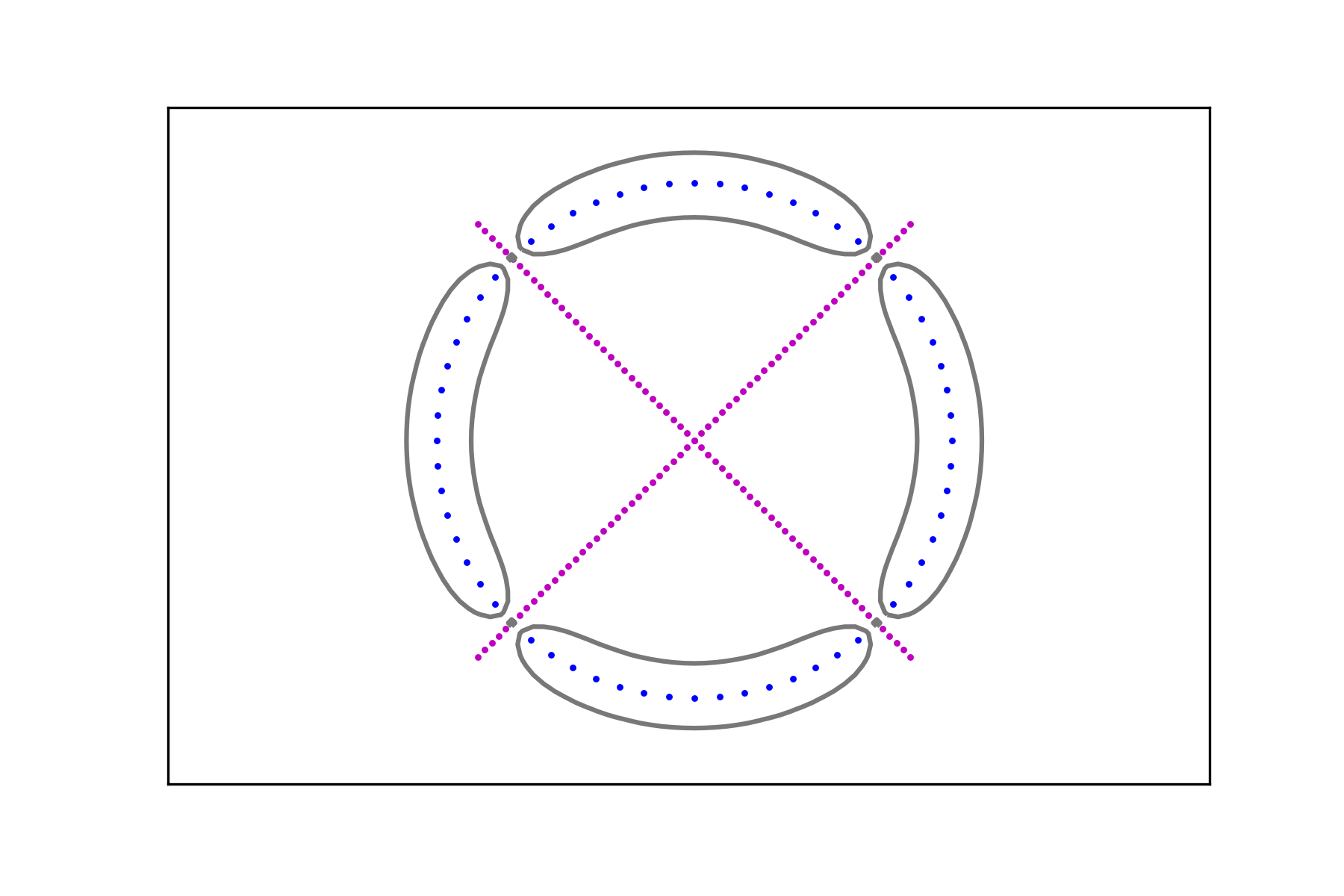}
  \includegraphics[scale=.35]{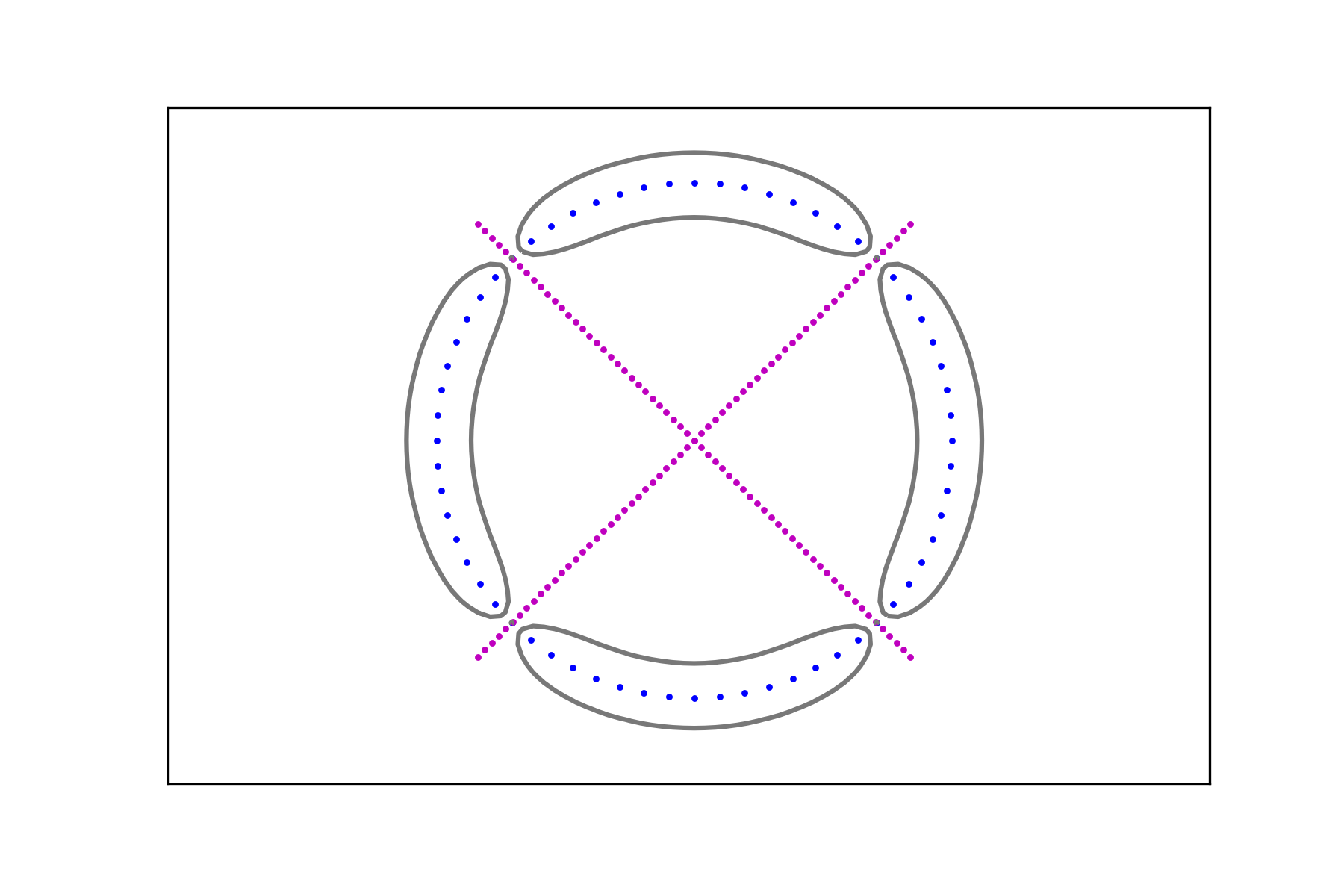}
  \includegraphics[scale=.35]{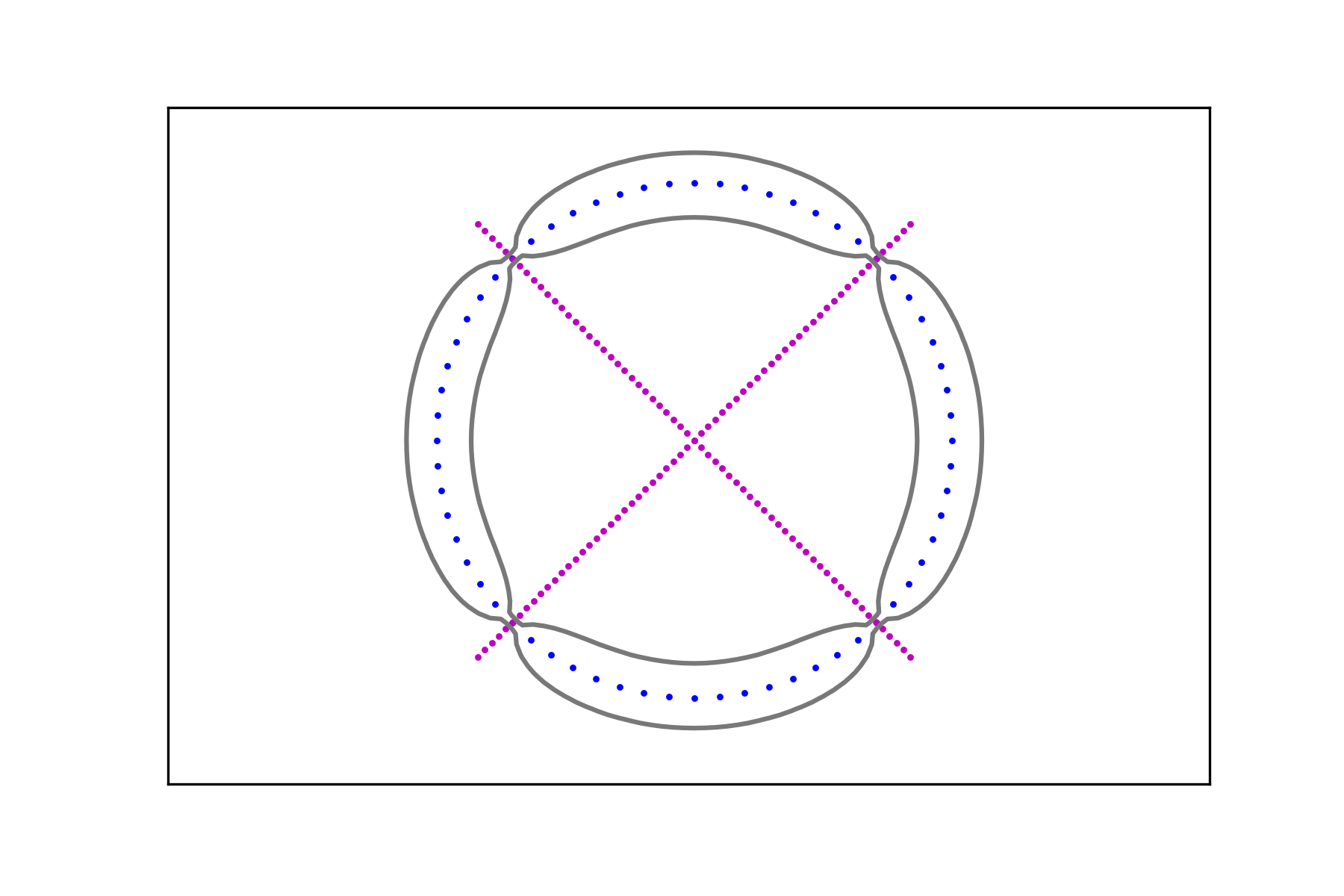}
  \includegraphics[scale=.35]{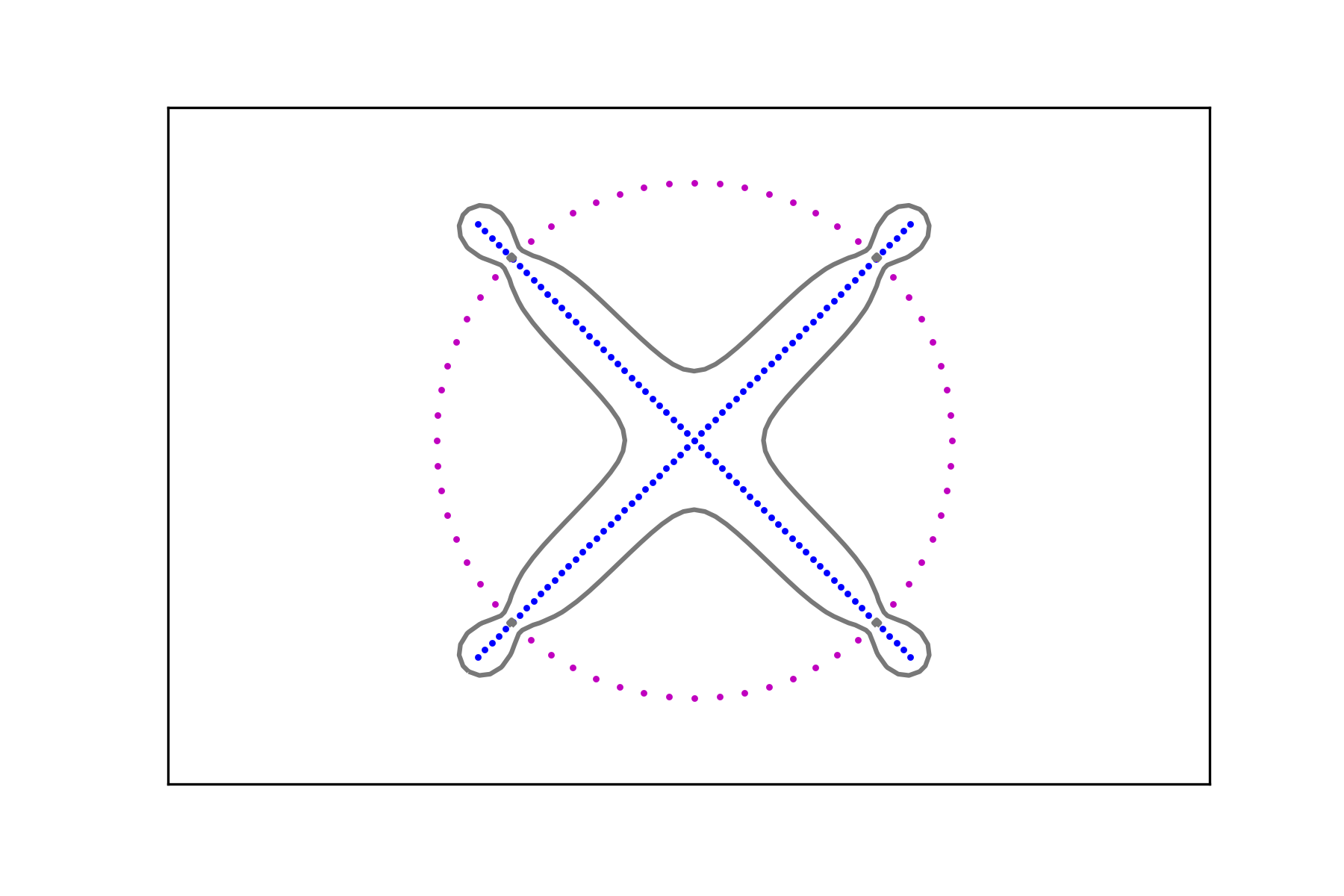}
  \includegraphics[scale=.35]{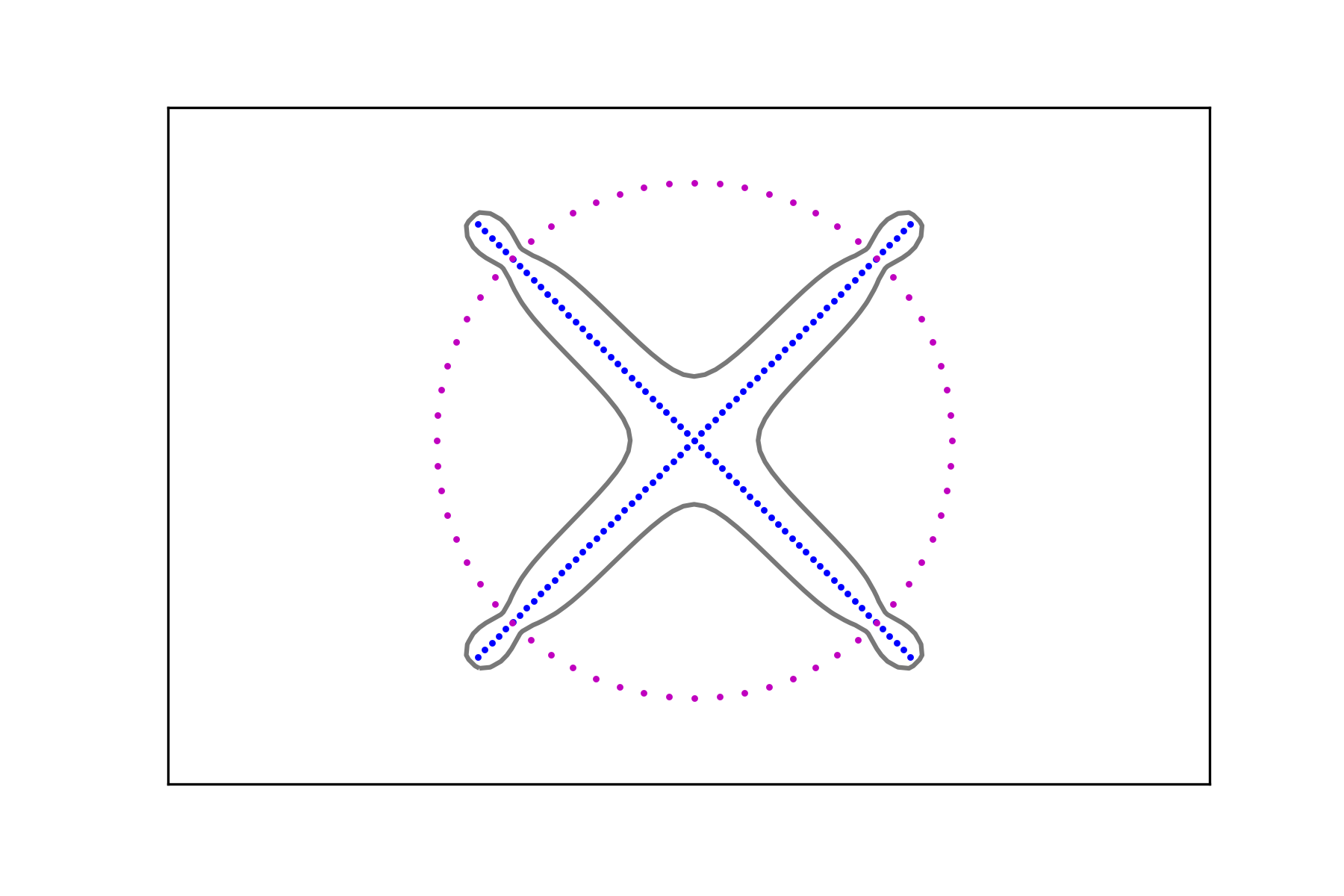}
  \includegraphics[scale=.35]{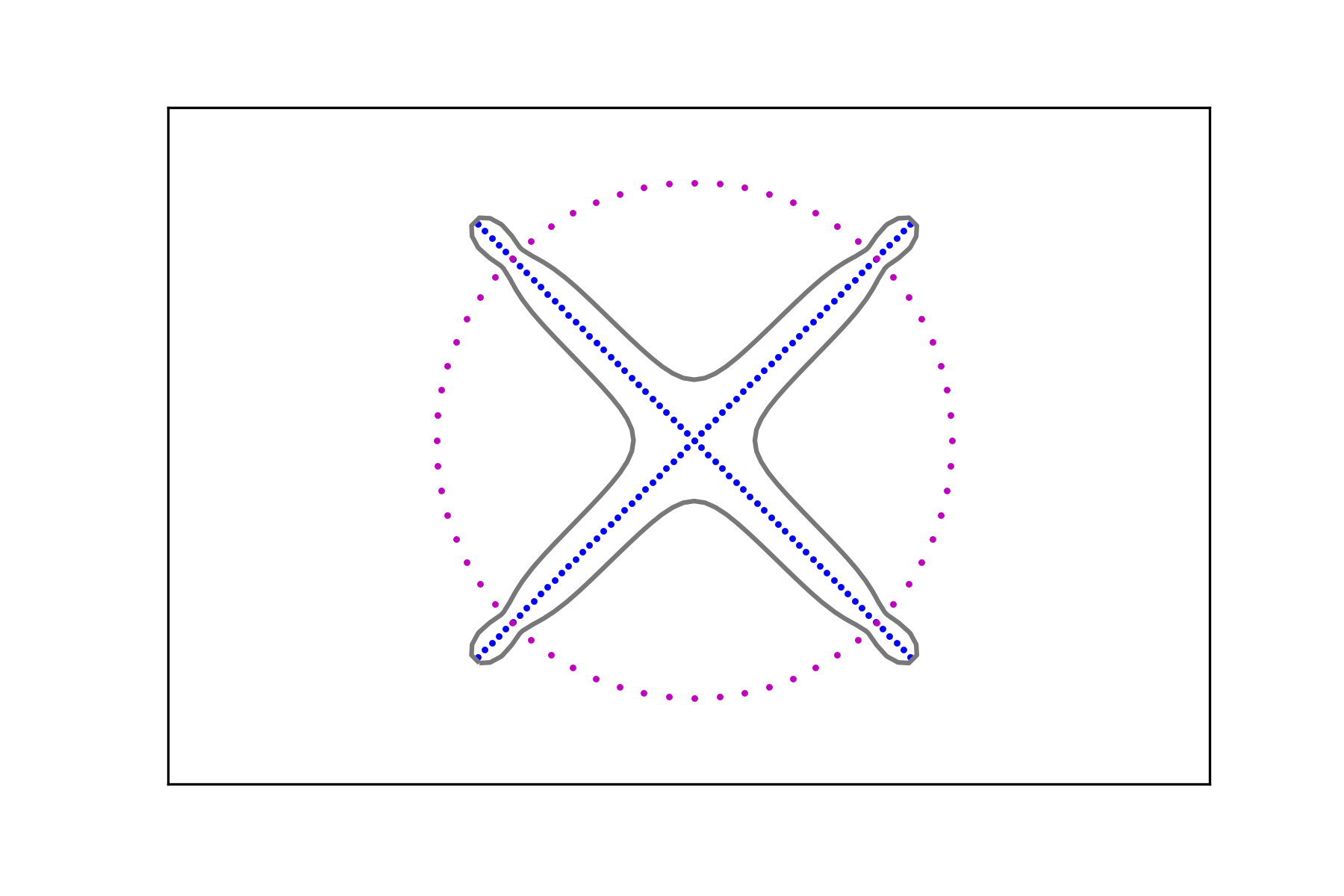}
  \caption{The 50\% of maximum level lines for the signals
    $u_{\mathbb{D}_C}$ and $u_{\mathbb{D}_S}$ of the data sets based
    on the two data classes $C_k$ and $S_k$ for $k=1,2$. The first two
    rows correspond to $k=1$ and the second two to $k=2$. Withing each
    row, from left to right, the regularization parameter is $\alpha=.1,1,2$.}
  \label{fig:cClevels}
\end{figure}

Consider the data sets $\mathbb{D}_k$, $k=1,2$, consisting of points
belonging to two distinct classes: points along a circle and points on
the union of two segments with two different densities as depicted in
Figure \ref{fig:cCdata}. For $k=1,2$, denote the circular data sets by
$$
C_k=\big\{ c^i_k\,\big |\, i=1,\dots m_{C_k}\big\}
$$
and the union of segments data sets by
$$
S_k=\big\{ s^i_k\,\big |\, i=1,\dots m_{S_k}\big\}
$$
In the spirit of the previous examples, we create two pairs of data sets
\begin{gather*}
  \mathbb{D}_{C_k}=\big\{ (c^i_k,1)\,\big |\, i=1,\dots m_{C_k}\big\}\cup
  \big\{ (s^i_k,0)\,\big |\, i=1,\dots m_{S_k}\big\}\text{ and
  }\\\mathbb{D}_{S_k}=\big\{ (s^i_k,1)\,\big |\, i=1,\dots m_{S_k}\big\}\cup
  \big\{ (c^i_k,0)\,\big |\, i=1,\dots m_{C_k}\big\},\: k=1,2
\end{gather*}
and compute the associated signals $u_{\mathbb{D}_{C_k}}$ and
$u_{\mathbb{D}_{S_k}}$, $k=1,2$. Figure \ref{fig:cClevels} shows the level sets
$$
[u_{\mathbb{D}_{C_k}}= .5\max(u_{\mathbb{D}_{C_k}})\bigr]\text{ and }
[u_{\mathbb{D}_{S_k}}=.5 \max(u_{\mathbb{D}_{S_k}})\bigr],\: k=1,2.
$$
for different values of the regularization parameter.
The region in their interior (i.e. the one containing the
corresponding data set) can be 
considered as a smooth fattening of the data set to a set of positive
measure. It can be obtained for any data set regardless of the
intensity of its signal.

\begin{figure}
  \includegraphics[scale=.35]{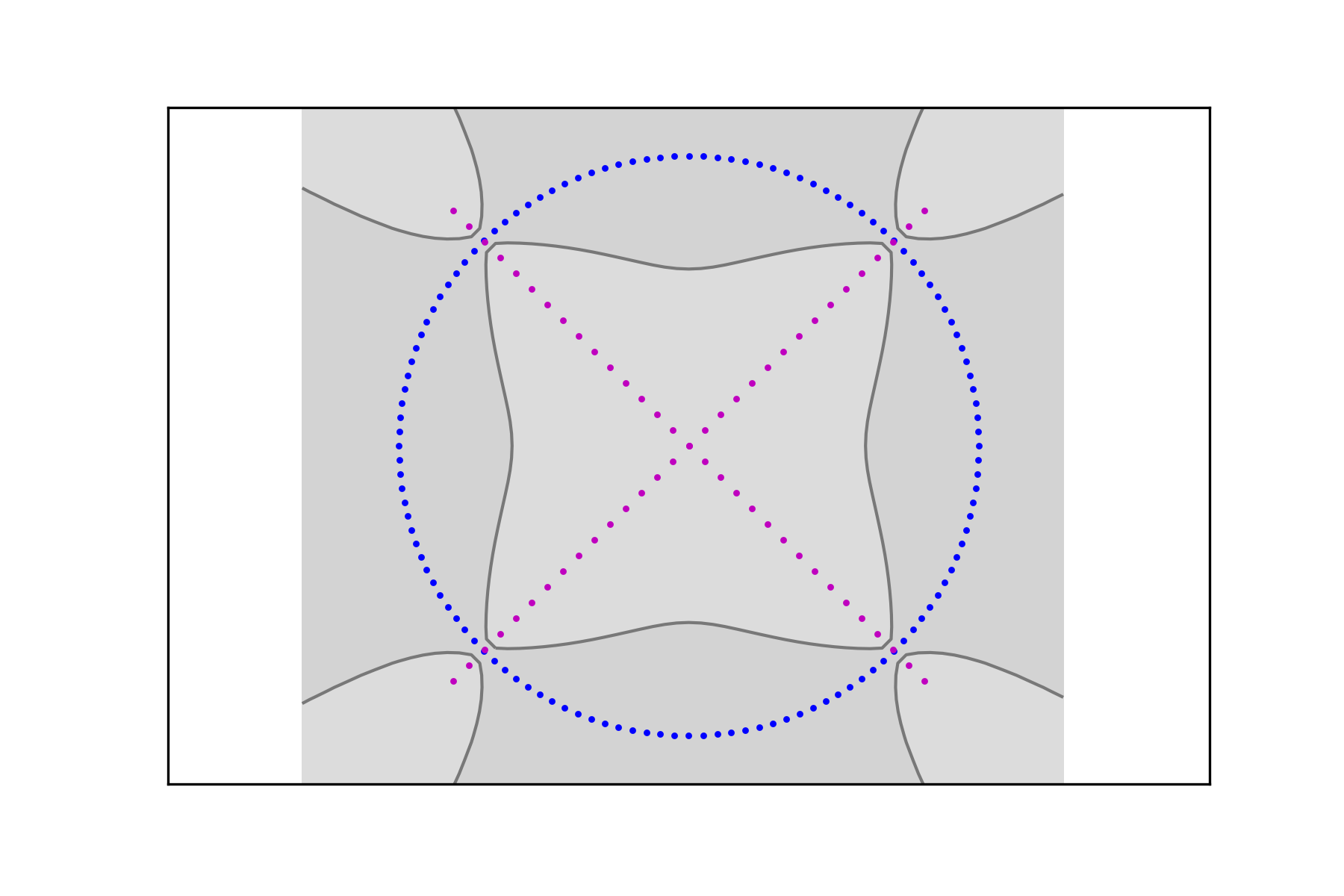}
  \includegraphics[scale=.35]{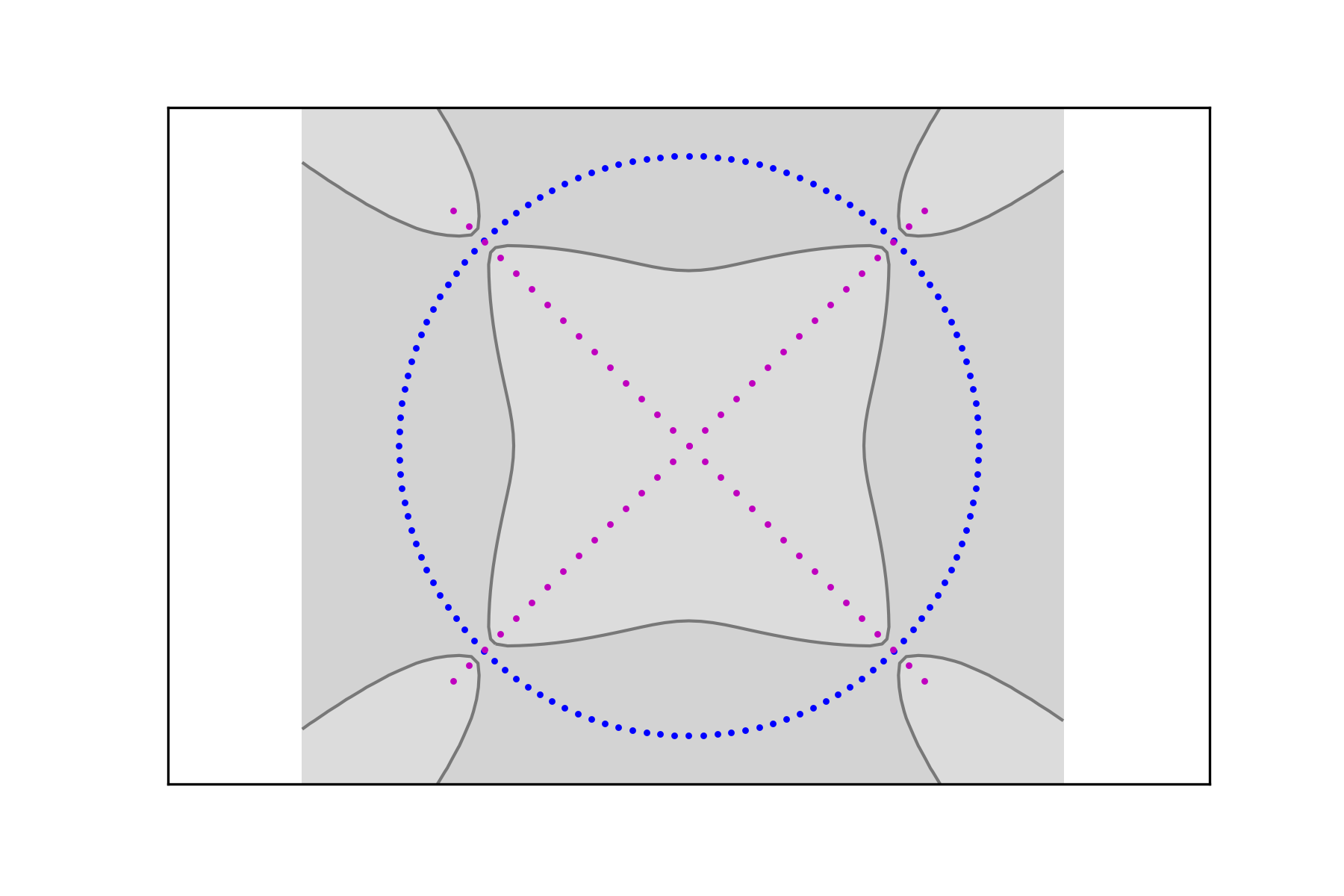}
  \includegraphics[scale=.35]{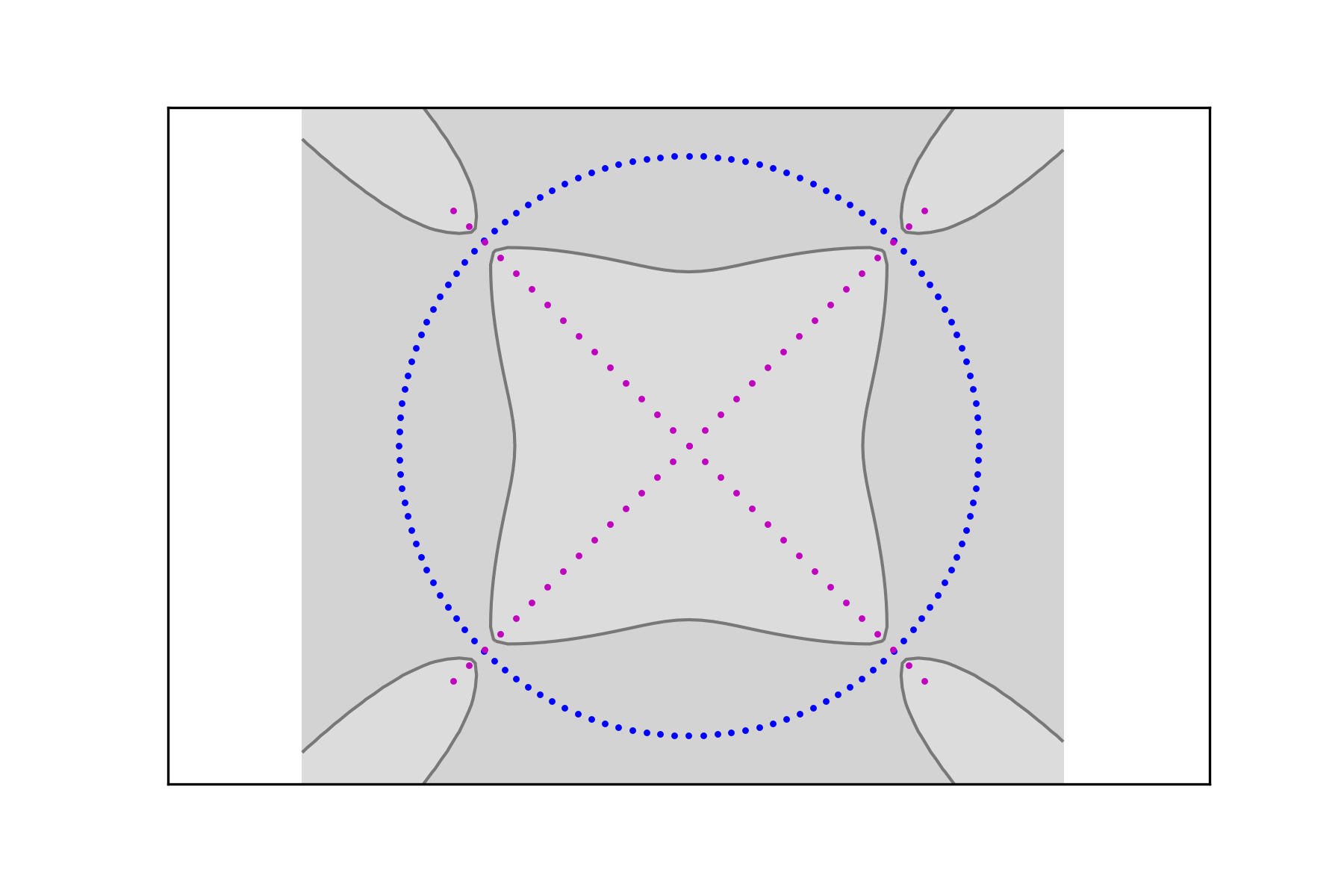}
  \includegraphics[scale=.35]{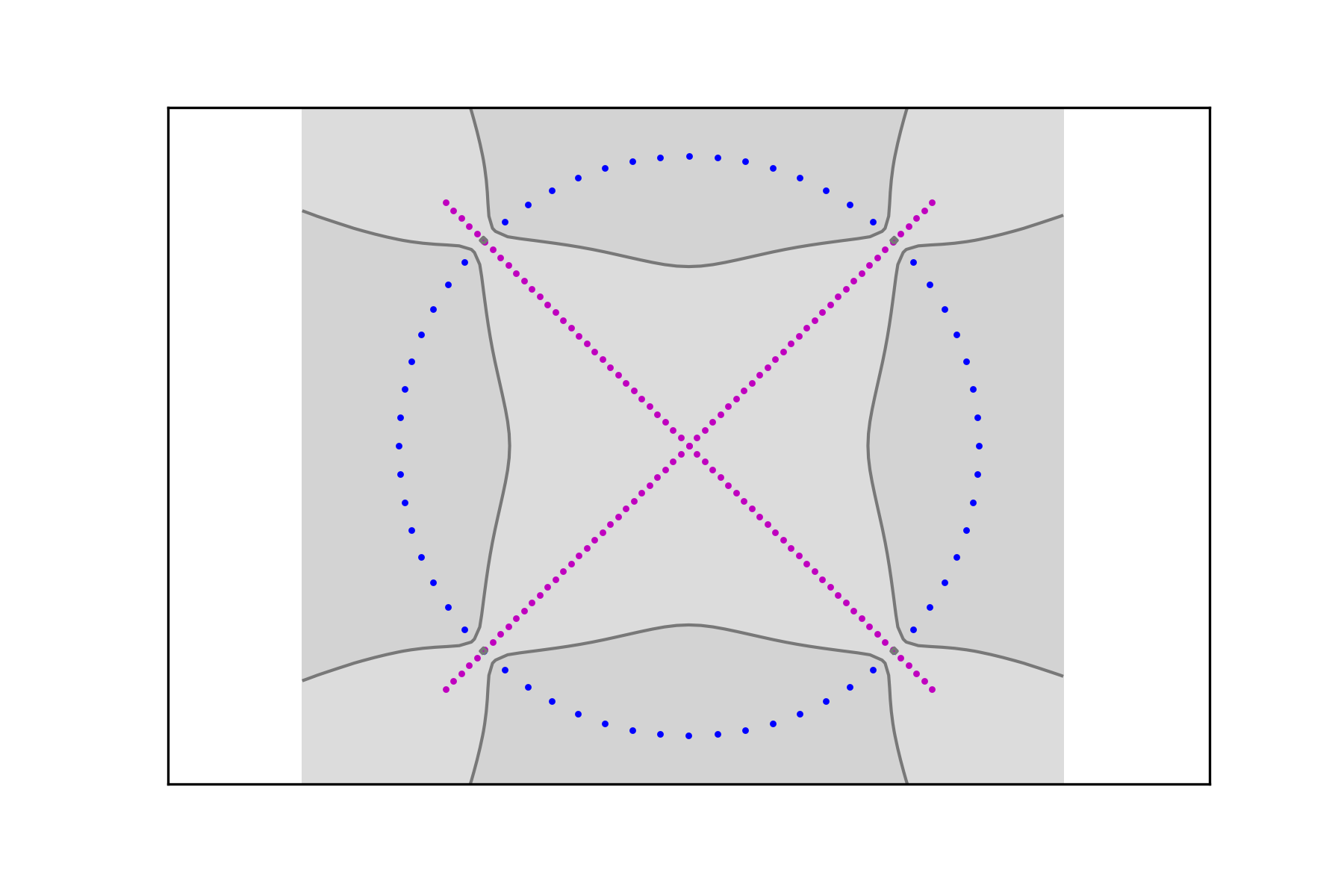}
  \includegraphics[scale=.35]{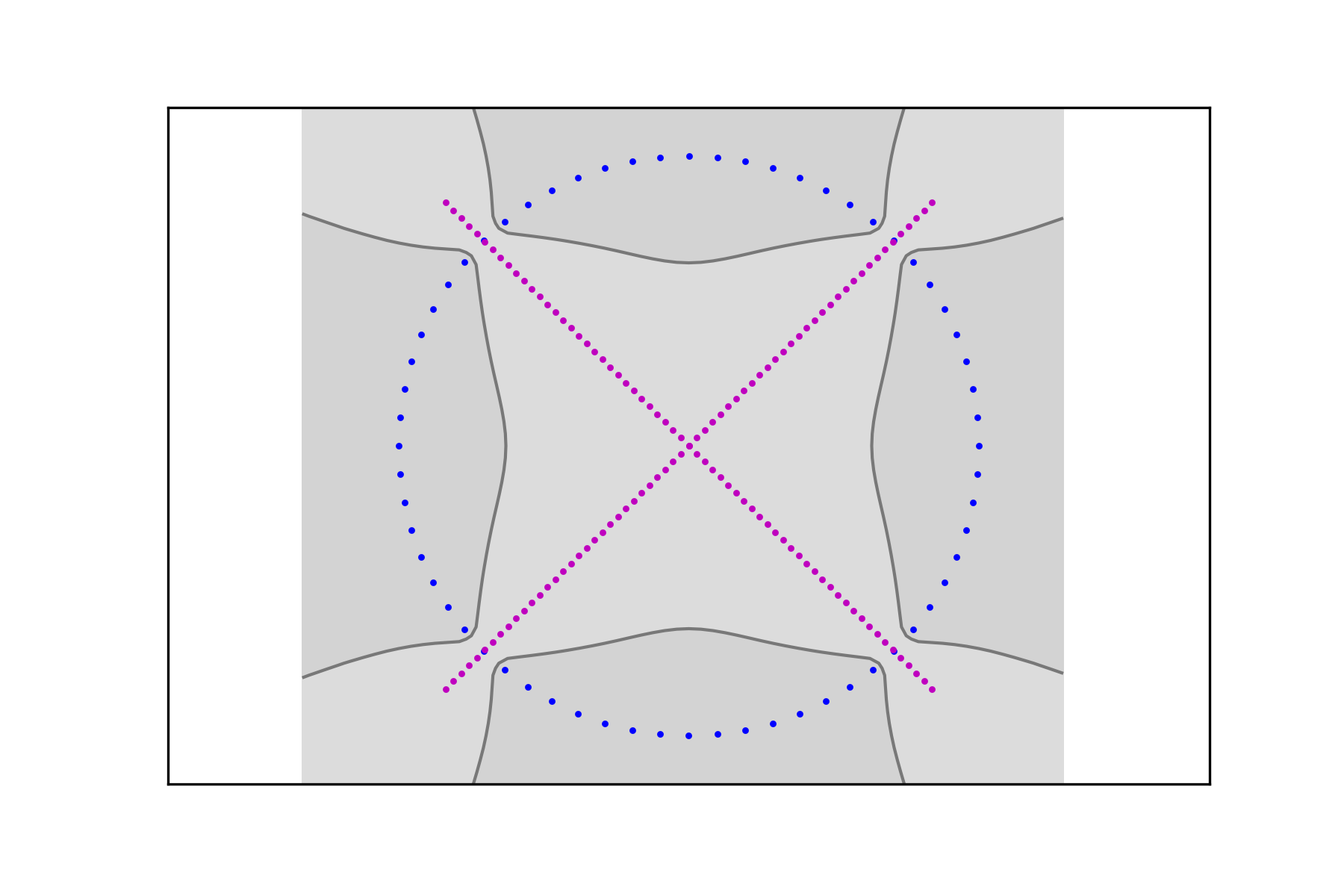}
  \includegraphics[scale=.35]{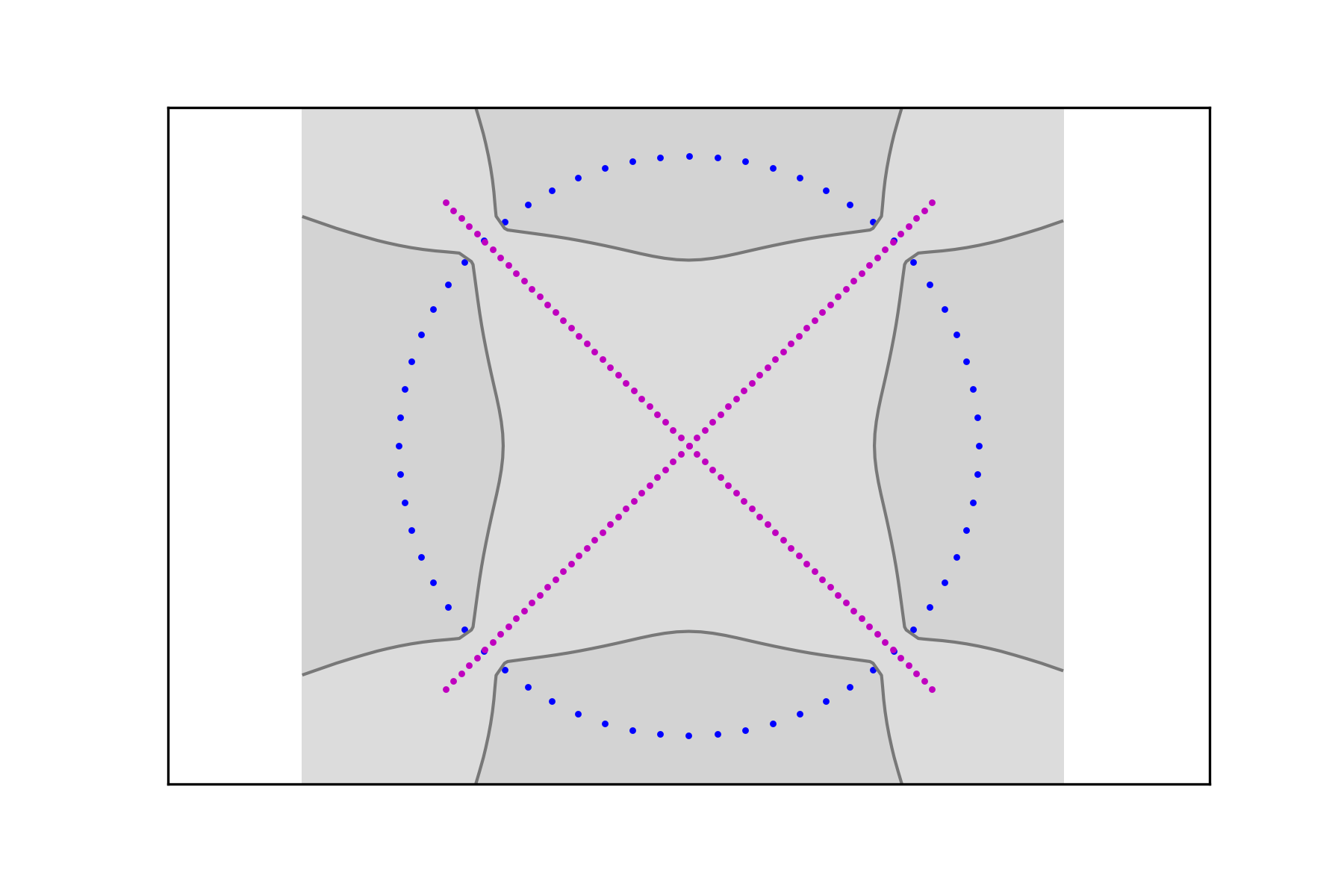}
  \caption{The decision boundary computed according to \eqref{dBdry}
    for two classes depicted in the same image. The $k$th row
    corresponds to the class pair $(C_k,S_k)$, $k=1,2$. Notice how the
    decision boundary is affected by the ``density'' of the data
    sets and not very strongly affected by the regularization
    parameter. The latter is, from left to right, 
    $\alpha=.1,1,2$.}
  \label{fig:dBdry}
\end{figure}

Next we turn our interest to the question of classification: given a
point $z\in \mathbb{R}^2$ that needs to be classified, we use the
decision algorithm defined by \eqref{algo}. This gives
$$
 L(z)=\operatorname{argmax}_{l=1,2}u_{\mathbb{D}_l}(z),
$$
which, in this case, yields the level lines (hypersurfaces in
higher dimension)
\begin{equation}\label{dBdry}
\bigl[ u_{\mathbb{D}_1}=u_{\mathbb{D}_2}\bigr]
\end{equation}
as the decision boundary. This is illustrated in Figure
\ref{fig:dBdry} for the classification problem of the two class pairs
$(C_k,S_k)$ for $k=1,2$ introduced above.
If the data pairs $(C_k,S_k)$, $k=1,2$, are considered the ground truth, then the
above decision boundary is arguably optimal. If, on the other hand, it
is known that the actual sets are the continuous circle and the union
of two segments, then the data are only a sampling of these sets. In
this case, the decision boundary may be biased by the relative
oversampling of the one set compared to the other. This is evident when one
compares the decision boundaries of Figure \ref{fig:dBdry}. In
concrete situations, if information about the dimensionality of the
ground truth is known, this effect can be mitigated by using
comparable sampling rates for the different classes (see next section
for an example of this procedure).

\begin{figure}
  \includegraphics[scale=.35]{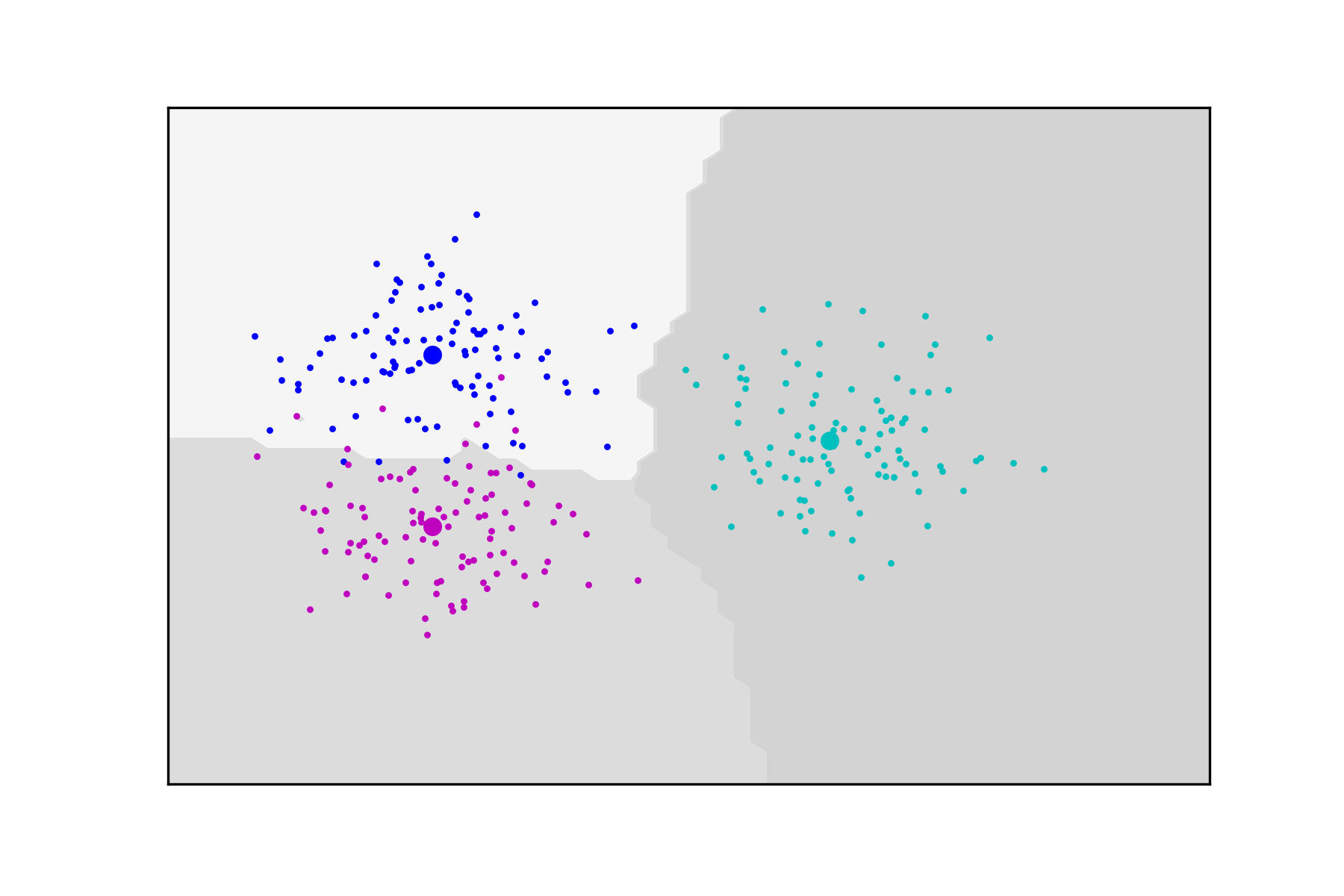}
  \includegraphics[scale=.35]{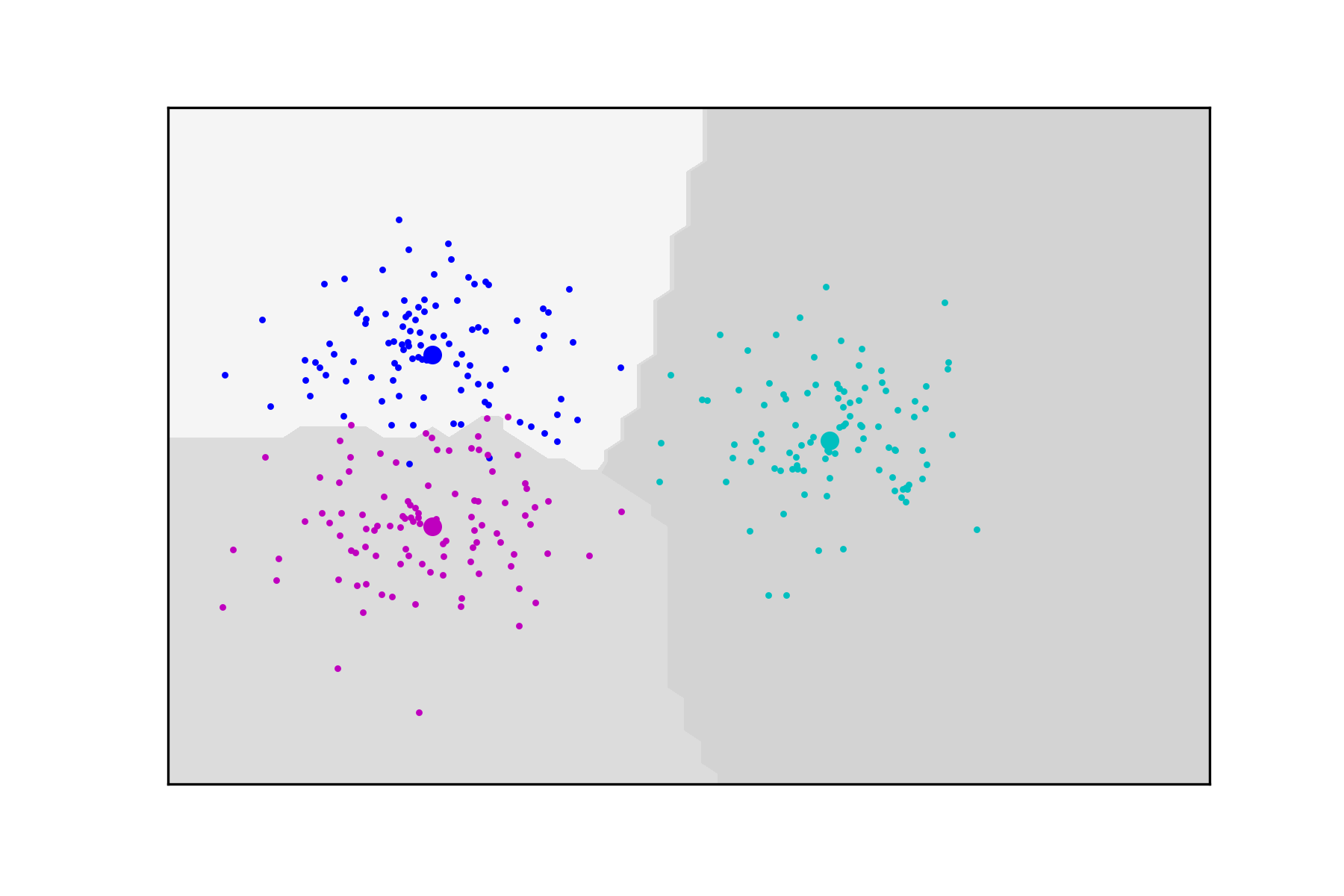}
  \includegraphics[scale=.35]{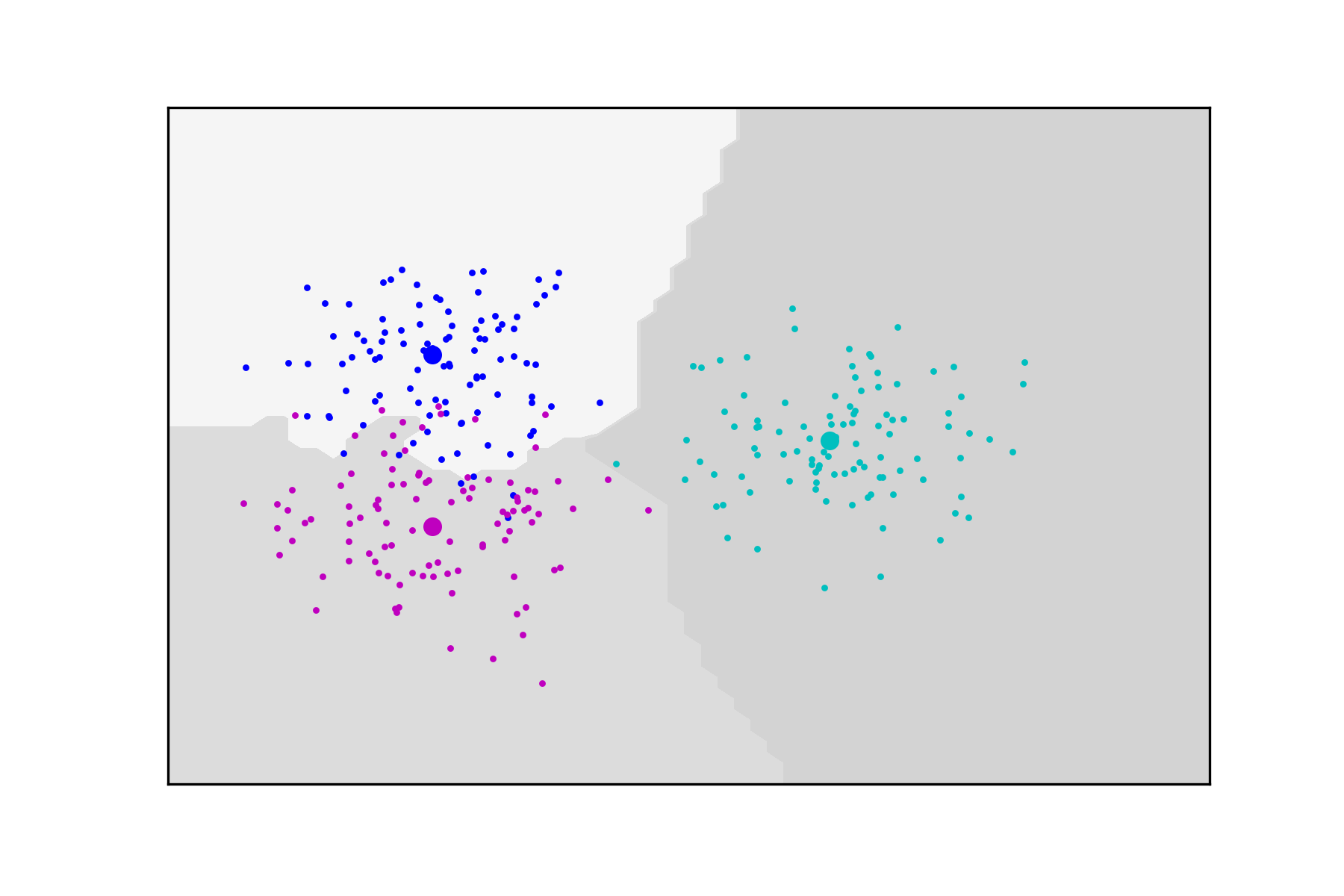}
  \includegraphics[scale=.35]{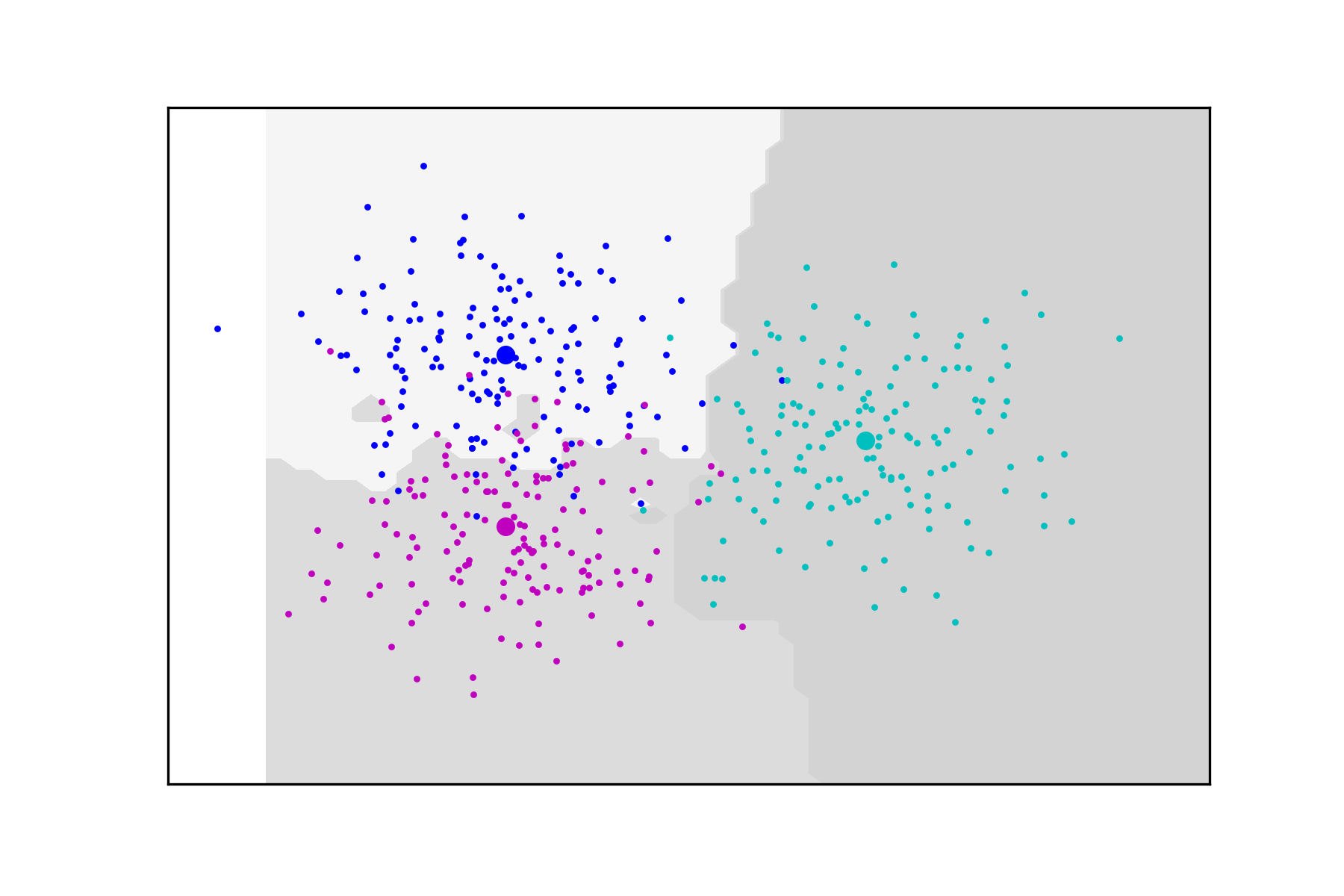}
  \includegraphics[scale=.35]{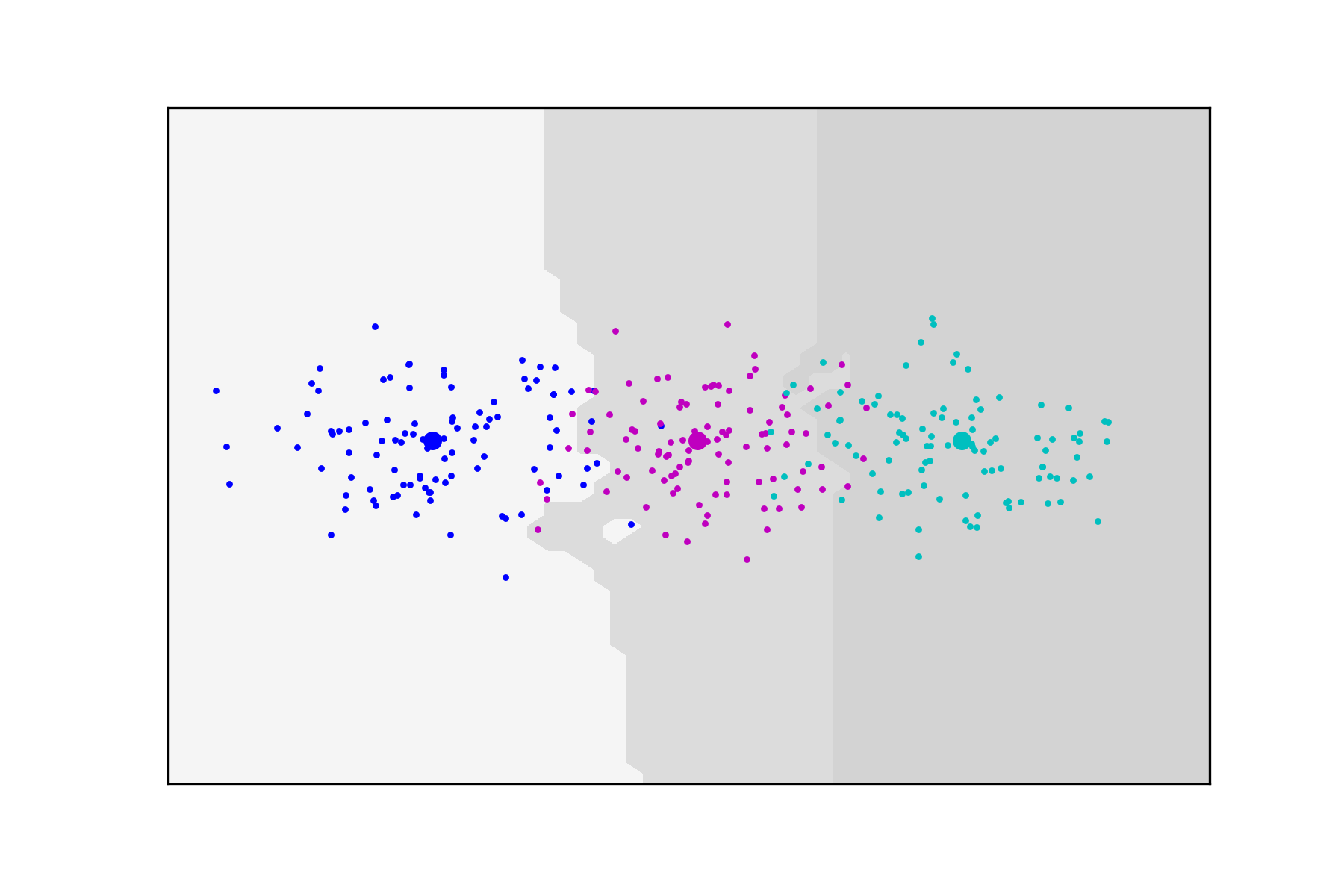}
  \includegraphics[scale=.35]{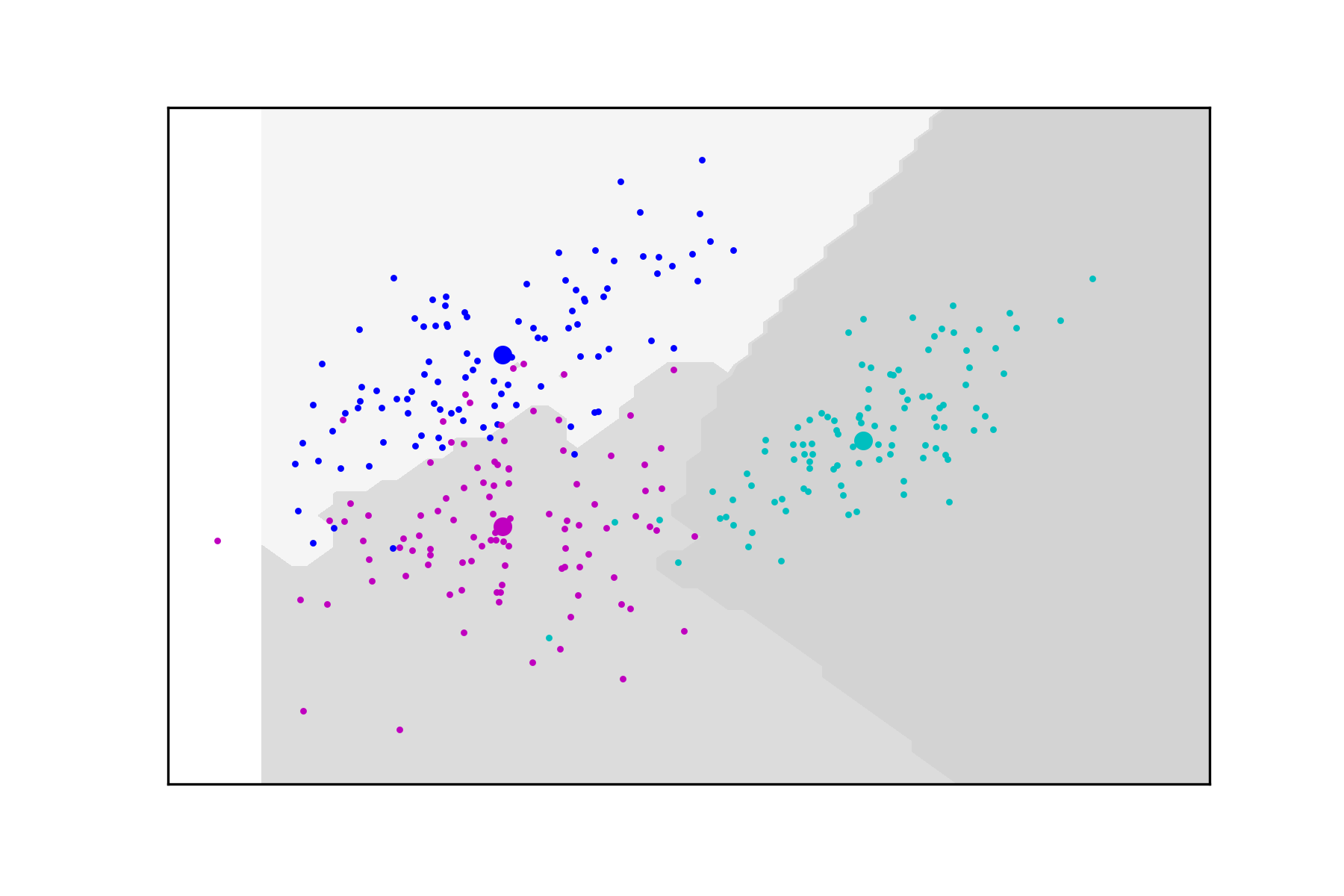}
  \caption{The decision regions computed according to \eqref{algo}
    for three classes of normally distributed points depicted using
    different colors in the same image. The average of each class is
    plotted as a large disk. The first row shows three different
    realization of the same three normal distributions and the
    computed decision regions. The first image in the second row is
    based on the same distributions as the first row but the sample
    size is increased, the second and third show the decision regions
    for different choices of means and covariance matrices. In all but the last
    example the covariance is taken to be diagonal. In the third and
    the last image, one of the sample points is outside the
    computational box where the data signals are generated and hence
    generated an unshaded region.}
  \label{fig:gaussians}
\end{figure}

We conclude this section with a classification problem for data $\mathbb{X}_0$
split into three classes $\mathbb{X}_l$, $l=1,2,3$, each consisting of a set of
points which are normally distributed with mean $p_l\in \mathbb{R}^2$
and different covariance matrices. The data set and the corresponding
decision regions computed based on \eqref{algo} are depicted in Figure
\ref{fig:gaussians}. In these experiments $\alpha=1$.
\subsection{The MNIST data set}
The final application is to the standard machine learning example and
toy problem of digit classification for the MNIST data set. The data
set consists of $28\times 28$ grayscale images of hand-written digits
stored as vectors in $[0,255]^{784}$. Here it is considered that
the ambient space is simply $\mathbb{R}^{784}$. The argument data is
normalized to have unit Euclidean norm, that is, each original vector
$x$ is replaced by $x/|x|$. In this way the maximal Euclidean distance
between any two data points is $2\sqrt{2}$. The data set is split into a
training set containing 60,000 data points $x$ and their corresponding
label $d(x)$ indicating which digit is represented, and a testing
data set of size 10,000. The labels of the testing data are known but
need to be inferred from any knowledge that can be gleaned from the
training set. This an example of when, due to the so-called curse of
dimensionality, the data does not have any hope to fill the ambient
space uniformly and thus, even if one assumed the existence of an
underlying function $d:\mathbb{R}^{784}\to\{ 0,1,\dots,9\}$,
the data would never be sufficient to accurately approximate it. It
has to be said of course, that the testing data mostly does not stray
away significantly from the training data and the different digits in
the latter build thin subsets of the ambient space. This fact is typically
captured by saying that the data lives in some lower dimensional
manifold(s). We know from the previous section and from the two
dimensional experiments, however, that the (training) data still
generates a significant, if not strong, signal. The classification
method described in the sequel exploits this signal and does not
require any kind of training based on the minimization of non-convex
functionals, as is often the case for machine learning algorithms
based on neural nets. It
is, in fact, based on the solution of low dimensional, well-posed
linear systems as is about to be explained. First, in order to
strengthen the signal somewhat, the training set is expanded to
include rotations by $\pm 10^\circ$ and horizontal/vertical
translations by $\pm2$ pixels of each image. Then, given a test image
$z$, the closest $5$ training images of each digit class are
determined 
$$
\mathbb{X}=\big\{ x^{i_j}\,\big |\, j=1,\dots,50\big\},
$$
where $d(x^{i_j})=\lfloor j/5\rfloor$. The idea is now to use system
\eqref{theSys} in order to produce
approximate interpolants $u_d:=u_{\mathbb{D}_d}$ of the characteristic
functions of each digit class given by the data set
$$
\mathbb{D}_d=\big\{ \bigl( x^{i_j}, \delta_{d,d(x^{i_j})}\bigr)\,\big |\,
j=1,\dots,50\big\},
$$
where $d=0,\dots,9$, and
$$
\delta_{d,\bar d}=\begin{cases} 1,& d=\bar d,\\ 0,& d\neq \bar d.
\end{cases}
$$
Finally the approximative characteristic functions $u_d$ will compete
to determine the digit $d(z)$ to be associated with the test image $z$
via \eqref{algo}, in this case
$$
 d(z)=\operatorname{argmax}_d u_d(z).
$$
This approach yields an accuracy of $98.56\%$\footnote{For comparison,
a classification based on a direct nearest neighbor approach using the
extended training set has an
accuracy of $97.86\%$} on the test set. In Table \ref{table:cTable} we
record the detailed outcome of the classification, performed with
$\alpha=1.5$. 

Recall that this method is stable and depends continuously on the data set
and hence delivers a robustness that methods with higher
classification rates typically do not. Unlike neural networks it, moreover,
does not require any training but uses the training set directly in a
fully transparent way.
\begin{table}[h!]
  \centering
  \begin{tabular}{ |c||c|c|c|c|c|c|c|c|c|c| }\hline
    \diagbox{Digit}{Label}&0&1&2&3&4&5&6&7&8&9\\\hline\hline
    0&973&0&1&0&0&2&3&1&0&0\\\hline
    1&0&1131&2&1&0&0&0&1&0&0\\\hline
    2&4&1&1015&0&1&0&0&10&0&1\\\hline
    3&0&0&1&995&0&8&0&3&3&0\\\hline
    4&0&0&0&0&971&0&4&1&0&6\\\hline
    5&1&0&0&8&0&877&4&1&0&1\\\hline
    6&0&2&0&0&0&1&955&0&0&0\\\hline
    7&0&4&4&0&0&0&0&1020&0&0\\\hline
    8&2&0&3&9&3&2&3&4&946&2\\\hline
    9&1&2&1&4&9&7&0&10&2&973\\\hline
  \end{tabular}
  \vspace{2pt}
  \caption{MNIST classification results: the $k$th row of
the table indicates in column $j$ how many times the digit $k$ is
assigned the label $j$ by the algorithm.}
  \label{table:cTable}
\end{table}
\begin{rem}
It should be pointed out that the proposed classification method
performs well when the data classes effectively lie on submanifolds
the shape of which their relative signals are able to capture. If the
class similarity is not geometric in this sense, this
method will likely not produce satisfactory results if applied to the
original data. It is indeed possible for general data sets to exhbit
classes that share some common feature but are far apart as points in
space. In this case, the dots cannot be easily connected.
\end{rem}
%\bibliography{../../lite.bib}

\end{document}